\documentclass[journal]{IEEEtran}
\usepackage[table]{xcolor}
\usepackage{url}
\usepackage{amsmath}

\usepackage{mathtools}
\usepackage{booktabs}
\usepackage{rotating}
\usepackage{pifont}
\usepackage{bbold}
 \usepackage[textsize=footnotesize]{todonotes}
\ifCLASSOPTIONcompsoc
  \usepackage[caption=false,font=normalsize,labelfont=sf,textfont=sf]{subfig}
\else
  \usepackage[caption=false,font=footnotesize]{subfig}
\fi
\usepackage{lscape}
\usepackage{pdflscape}
\usepackage{array}
\usepackage{tabulary}
\usepackage{longtable}
\usepackage{booktabs}
\usepackage{longtable}
\usepackage{graphicx}
\usepackage{amssymb}
\usepackage[numbers]{natbib}
\usepackage{pgfplots}
\usepackage{multirow}
\usepackage{longtable}
\usepackage{mathtools}
\usepackage{makecell}
\usepackage{graphicx}
\usepackage{MnSymbol}
\usepackage{stackengine}
\usepackage{amsmath}
\usepackage{nccmath}
\usepackage{xspace}
\usepackage{tikz}
\usepackage{scalerel}
\usepackage{tabularx}
\usepackage[edges]{forest}

\usepackage{algorithmic}
\usetikzlibrary{svg.path}

\DeclareMathOperator*{\argmax}{arg\,max} 
\DeclareMathOperator*{\argmin}{arg\,min} 
\DeclareMathOperator*{\E}{\mathbb{E}}

\usetikzlibrary{fit, positioning, arrows, shapes.geometric, bayesnet}
\tikzstyle{startstop} = [rectangle, rounded corners, minimum width=3cm, minimum height=1cm,text centered, draw=black, fill=red!30]
\tikzstyle{io} = [trapezium, trapezium left angle=70, trapezium right angle=110, minimum width=3cm, minimum height=1cm, text centered, draw=black, fill=blue!30]
\tikzstyle{process} = [rectangle, minimum width=3cm, minimum height=1cm, text centered, draw=black, fill=orange!30]
\tikzstyle{decision} = [diamond, minimum width=3cm, minimum height=1cm, text centered, draw=black, fill=green!30]
\tikzstyle{arrow} = [thick,->,>=stealth]
\forestset{
  every leaf node/.style={
    if n children=0{#1}{}
  },
  every tree node/.style={
    if n children=0{}{#1}
  },
}

\newcommand{\define}{=}

\definecolor{mycolor}{RGB}{0, 0, 0}
\definecolor{one}{HTML}{7400B8}
\definecolor{two}{HTML}{5E60CE}
\definecolor{three}{HTML}{5390D9}
\definecolor{four}{HTML}{48BFE3}
\definecolor{five}{HTML}{64DFDF}

\definecolor{Gray}{gray}{0.9}
\newcommand{\acronym}{{\color{mycolor} real-world AL}\xspace}

\makeatletter
\def\LT@makecaption#1#2#3{%
\LT@mcol\LT@cols c{\hbox to\z@{\hss\parbox[t]\LTcapwidth{%
\footnotesize\bgroup\par\centering\@IEEEtabletopskipstrut{\normalfont\footnotesize #2}\\{\normalfont\footnotesize\scshape #3}\par\addvspace{0.5\baselineskip}\egroup\endgraf%
\@IEEEtablecaptionsepspace}%
\hss}}}
\makeatother
\newcolumntype{P}[1]{>{\raggedright\arraybackslash}p{#1}}
\newcolumntype{L}[1]{>{\raggedright\let\newline\\\arraybackslash\hspace{0pt}}m{#1}}
\newcolumntype{C}[1]{>{\centering\let\newline\\\arraybackslash\hspace{0pt}}m{#1}}
\newcolumntype{R}[1]{>{\raggedleft\let\newline\\\arraybackslash\hspace{0pt}}m{#1}}
\newcommand{\cmark}{\ding{51}}%
\newcommand{\xmark}{\ding{55}}%
\newcolumntype{Y}{>{\raggedright\arraybackslash}X}
\newcolumntype{Z}{>{\centering\arraybackslash}X}

\newsavebox{\ieeealgbox}

\pgfplotsset{compat=1.16}
\makeatletter
\def\@cline#1-#2\@nil{%
  \omit
  \@multicnt#1%
  \advance\@multispan\m@ne
  \ifnum\@multicnt=\@ne\@firstofone{&\omit}\fi
  \@multicnt#2%
  \advance\@multicnt-#1%
  \advance\@multispan\@ne
  \leaders\hrule\@height\arrayrulewidth\hfill
  \cr
  \noalign{\nobreak\vskip-\arrayrulewidth}}
\makeatother

\definecolor{orcidlogocol}{HTML}{A6CE39}
\tikzset{
    orcidlogo/.pic={
        \fill[orcidlogocol] svg{M256,128c0,70.7-57.3,128-128,128C57.3,256,0,198.7,0,128C0,57.3,57.3,0,128,0C198.7,0,256,57.3,256,128z};
        \fill[white] svg{M86.3,186.2H70.9V79.1h15.4v48.4V186.2z}
        svg{M108.9,79.1h41.6c39.6,0,57,28.3,57,53.6c0,27.5-21.5,53.6-56.8,53.6h-41.8V79.1z M124.3,172.4h24.5c34.9,0,42.9-26.5,42.9-39.7c0-21.5-13.7-39.7-43.7-39.7h-23.7V172.4z}
        svg{M88.7,56.8c0,5.5-4.5,10.1-10.1,10.1c-5.6,0-10.1-4.6-10.1-10.1c0-5.6,4.5-10.1,10.1-10.1C84.2,46.7,88.7,51.3,88.7,56.8z};
    }
}

\newcommand\orcidicon[1]{\href{https://orcid.org/#1}{\mbox{\scalerel*{
                \begin{tikzpicture}[yscale=-1,transform shape]
                \pic{orcidlogo};
                \end{tikzpicture}
            }{|}}}}
            
\usepackage{hyperref}

\usepackage[resetlabels]{multibib}
\newcites{supplement}{References}

\begin{document}

\title{A Survey on Cost Types, Interaction Schemes, and Annotator Performance Models in Selection Algorithms for Active Learning in Classification}

\author{

Marek Herde$^{\textsuperscript{\orcidicon{0000-0003-4908-122X}}}$, Denis Huseljic$^{\textsuperscript{\orcidicon{0000-0001-6207-1494}}}$, Bernhard Sick$^{\textsuperscript{\orcidicon{0000-0001-9467-656X
}}}$, \IEEEmembership{Member, IEEE}, Adrian Calma$^{\textsuperscript{\orcidicon{0000-0003-4380-7018
}}}$
\thanks{M. Herde, D. Huseljic, B. Sick, and A. Calma are with the department
of Intelligent Embedded Systems, University of Kassel, Germany (e-mail: \{marek.herde $|$ dhuseljic $|$ bsick $|$ adrian.calma\}@uni-kassel.de).}% <-this % stops a space
\thanks{This research was supported by the CIL project at the University of Kassel under internal funding P/710 and P/1082.}% <-this % stops a space
\thanks{We thank Daniel Kottke, Tuan Pham Minh, Lukas Rauch, and Robert Monarch for their comments that greatly improved this survey.}
}

{\maketitle}

\begin{abstract}
Pool-based active learning (AL) aims to optimize the annotation process (i.e., labeling) as the acquisition of annotations is often time-consuming and therefore expensive. 
For this purpose, an AL strategy queries annotations intelligently from annotators to train a high-performance classification model at a low annotation cost.
Traditional AL strategies operate in an idealized framework.
They assume a single, omniscient annotator who never gets tired and charges uniformly regardless of query difficulty.  
However, in real-world applications, we often face human annotators, e.g., crowd or in-house workers, 
who make annotation mistakes and can be reluctant to respond if tired or faced with complex queries.
Recently, a wide range of novel AL strategies has been proposed to address these issues.
They differ in at least one of the following three central aspects from traditional AL:
(1)~They explicitly consider (multiple) human annotators whose performances can be affected by various factors, such as missing expertise.
(2)~They generalize the interaction with human annotators by considering different query and annotation types, such as asking an annotator for feedback on an inferred classification rule.
(3)~They take more complex cost schemes regarding annotations and misclassifications into account.
This survey provides an overview of these AL strategies and refers to them as \acronym.
Therefore, we introduce a general \acronym strategy as part of a learning cycle and use its elements, e.g., the query and annotator selection algorithm, to categorize about 60 \acronym strategies.
Finally, we outline possible directions for future research in the field of AL. 
\end{abstract}


\section{Introduction}
\label{sec:introduction}
\IEEEPARstart{I}NFORMATION and communication technology has become an integral part of humans' lives and supports us embedded in our surroundings~\cite{Haddon2004}. 
In particular, improving computational power and the ease of collecting a plethora of data has promoted \textit{machine learning} (ML)~\cite{Edwards2015,Roh2019}.
Nowadays, ML models are employed in various fields~\cite{Larranaga2018}, ranging from recommender systems~\cite{Zhang2019}, text classification~\cite{Kadhim2019}, and speech recognition~\cite{Chiu2018} to object detection in videos~\cite{Zhao2019}.
In this survey, we consider ML for building classification models.
They learn from data sets consisting of instances and their corresponding annotations (e.g., class labels, membership probabilities, etc.).
However, annotating instances may be costly and time-consuming since it is often manually executed by annotators. 

In general, an annotator is an information or knowledge source such as a human, the Internet, or a simulation system~\cite{Hanika2019} and can annotate various types of queries.
In this survey, we focus on human annotators.
Other commonly used terms are oracle~\cite{Du2010}, expert~\cite{Wallace2011}, worker~\cite{Zhao2014}, teacher~\cite{Dekel2012}, and labeler~\cite{Zheng2010}.
A large group of (human) annotators who do not necessarily know each other is also named a crowd~\cite{Estelles2012}. 
The exact characteristics of a crowd, e.g., the number and heterogeneity of the annotators, depend on the requirements of the crowdsourcing initiative at hand.

\begin{figure*}[!h]
    \centering
    \begin{forest}
        for tree={% style of tree nodes
            %draw, semithick, rounded corners,
            every tree node={rounded corners=4pt, font=\footnotesize, text width = 40mm, text=mycolor, draw=mycolor},
            every leaf node={rounded corners=4pt, font=\footnotesize, text width = 95mm, text=mycolor, draw=mycolor},%{font=\tiny, text width = 50mm, text=black, draw=black},
            top color = white!20,
            bottom color = white!40,
            text badly centered,% <-- "align=center" doesn't work
            % style of tree (edges, distances, direction)
            %edge = {draw=blue!25!black, semithick},
            parent anchor = east, 
            child anchor = west,
            grow = east,
            s sep = 2mm,    % sibling distance
            l sep = 4mm,    % level distance
        }
        [\textbf{Real-world \\ Active Learning} \\ Objective and Learning Cycles (Section~\ref{sec:real_world_active_learning}), text width = 25mm, draw = {black, thick}, text=black
            [\textbf{Selection Algorithms:} \\ Which query is presented \\to which annotator? \\ (Section~\ref{sec:selection_strategies}), draw = {four, thick}, edge = {four, thick}, text=black
                [\textbf{Joint Selection:} \\ Pairs of queries and annotators are jointly selected. \\ (Section~\ref{subsec:joint_selection}), draw = {four, thick}, edge = {four, thick}, text=black
                ]
                [\textbf{Sequential Selection:} \\ Queries are selected in the first step and annotators in the second step. \\ (Section~\ref{subsec:sequential_selection}), draw = {four, thick}, edge = {four, thick}, text=black
                ]
            ]
            [\textbf{Annotator Performance Models:} \\  Who is the best annotator? \\ (Section~\ref{sec:annotator_models}), draw = {five, thick}, edge = {five, thick}, text=black
                [\textbf{Annotator Performance:} \\ Annotators differ in their performances regarding annotation quality. \\ (Section~\ref{subsec:annotator_performance}), draw = {five, thick}, edge = {five, thick}, text=black
                ]
                [\textbf{Influence Factors:} \\ Annotators are influenced by several factors during the annotations process. \\ (Section~\ref{subsec:influence_factors}), draw = {five, thick}, edge = {five, thick}, text=black
                ]
            ]
            [\textbf{Interaction Schemes:} \\ How to interact with annotators? \\ (Section~\ref{sec:types_of_queries_and_annotations}), draw = {two, thick}, edge = {two, thick}, text=black
                [\textbf{Annotation Types:} \\ There are different ways of providing learning information as annotations. \\ (Section~\ref{subsec:annotation_types}), draw = {two, thick}, edge = {two, thick}, text=black
                ]
                [\textbf{Query Types:} \\ There are different ways of querying an annotator for learning information. \\ (Section~\ref{subsec:query_types}), draw = {two, thick}, edge = {two, thick}, text=black
                ]
            ]
            [\textbf{Cost Types:} \\ Which costs are to be considered? \\ (Section~\ref{sec:types_of_cost}), draw = {one, thick}, edge = {one, thick}, text=black
                [\textbf{Annotation Cost:} \\ Costs may be imbalanced and they may depend on the annotator or query. \\ (Section~\ref{subsec:annotation_cost}), draw = {one, thick}, edge = {one, thick}, text=black
                ]
                [\textbf{Misclassification Cost:} \\ Costs may be imbalanced and they may depend on the type of misclassification. \\ (Section~\ref{subsec:misclassification_cost}), draw = {one, thick}, edge = {one, thick}, text=black
                ]
            ]
        ]
    \end{forest}
    \caption{Overview of \acronym and structure of this survey's main body: The blue nodes of the tree name the different topics of \acronym identified in this survey. Additionally, they provide a brief summary of each topic and reference the specific section with more details.}
    \label{fig:overview}
\end{figure*}
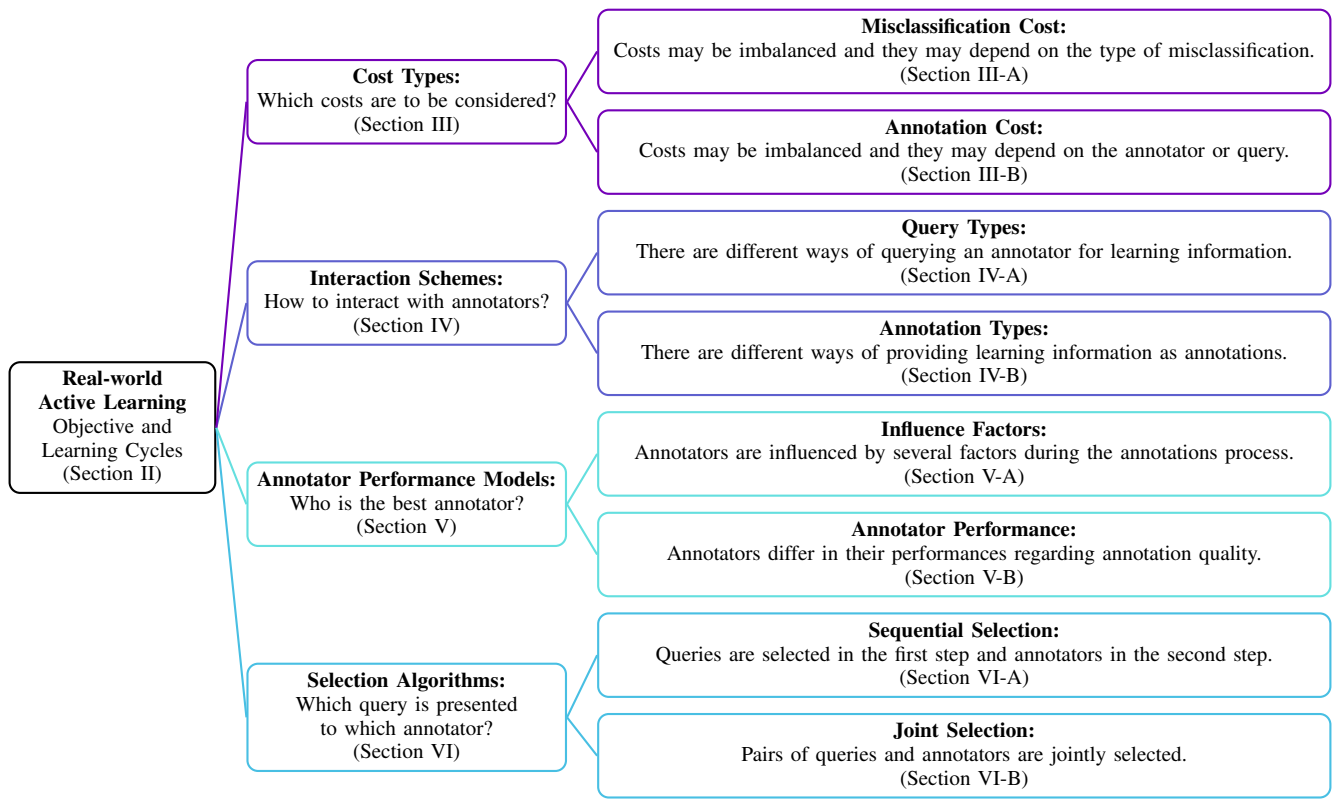

\textit{Active learning} (AL) is a subfield of human-in-the-loop learning~\cite{Munro2019} and interactive ML~\cite{Holzinger2016,Teso2020}, which directly and iteratively interacts with human annotators.
It aims at reducing annotation and misclassification cost~\cite{Settles2010}. 
Thus, an AL strategy queries annotations for instances from which the classification model is expected to learn the most~\cite{Aggarwal2014}. 
As a result, the classification model trained on an actively selected subset of annotated instances reaches in average a superior performance to a model trained on a randomly selected subset.
AL strategies have been successfully employed in several applications, e.g., malware detection~\cite{Nissim2014}, waste classification~\cite{Ahmed2020}, classification of medical images~\cite{Hoi2006}, and training of robots~\cite{Herde2018}.
However, many of these AL strategies make three central assumptions that limit their practical use~\cite{Settles2011}, and we refer to them as traditional AL.
\begin{itemize}
\item[(1)] There is a single omnipresent and omniscient annotator providing the correct annotation for each query at any time. This assumption conflicts with the available options of annotation acquisitions.
In particular, crowdsourcing represents a popular way to obtain data annotations~\cite{Howe2006}.
However, on crowdsourcing platforms, e.g., Amazon's Mechanical Turk~\cite{Paolacci2010,Buhrmester2011}, CloudResearch (formerly TurkPrime)~\cite{Litman2016}, and Prolific Academic~\cite{Peer2017}, multiple error-prone annotators have to be considered~\cite{Rodrigo2019}. 
Otherwise, annotation mistakes (e.g., noisy class labels) will degrade the classification model's performance~\cite{Zhu2004,Saez2014}.   
\item[(2)] The cost of an annotation is constant across the queries. This assumption is violated in cases such as biomedical citation screening~\cite{Wallace2011}, in which articles are to be classified as relevant or irrelevant for a particular research topic.
The time to annotate an article depends on its length, complexity, and the queried annotator~\cite{Arora2009}.
Hence, the cost varies across pairs of articles and annotators.
\item[(3)] Each query requests the class label of a specific instance. This assumption ignores the possibility of designing more general and effective queries~\cite{Settles2011,Angluin1988}.
Some of these queries also require annotations to be more complex than simple class labels.
We avoid confusion regarding the terms label and annotation by defining an annotation as the most general reply to a query, e.g., an annotator could answer a query with ``I have no idea!''. 
Correspondingly, a class label is a specific example of an annotation.
\end{itemize}

Various concepts have been proposed to overcome the limitations above. These include collaborative interactive learning~\cite{Calma2016,Bahle2016} and proactive learning~\cite{Donmez2008b,Donmez2010b}. We summarize their main differences to traditional AL through the three following aspects:
\begin{itemize}
    \item[(1)] Instead of assuming a single omniscient and omnipresent annotator, they consider (multiple) human annotators whose performances can be affected by various factors, e.g., missing expertise, fatigue, and malicious behavior.
    \item[(2)] Instead of repeatedly querying class labels of instances, they generalize the interaction with human annotators by considering different types of queries and annotations, such as asking an annotator for feedback on an inferred classification rule. 
    \item[(3)] Instead of assuming uniform cost, they take more complex cost schemes regarding annotations and misclassifications into account.
\end{itemize}
\textbf{In this survey}, we provide an overview of existing AL strategies taking at least one of the three aspects into account and refer to them as \acronym.
We limit the scope by including only strategies for classification in the pool-based AL setting~\cite{Settles2010} because it is the most researched AL field. 
However, many implications of this survey go beyond this scope and are emphasized in the outlook.
Based on these prerequisites, this survey makes the following contributions:
\begin{itemize}
	\item We formalize the objective of a \acronym strategy as the optimal annotation sequence to a cost-sensitive classification problem.
	\item We propose a taxonomy of existing cost types, interaction schemes, annotator performance models, and selection algorithms to compare different \acronym strategies.
	\item We give a comprehensive comparison of about 60 \acronym strategies and analyze them regarding their handling of error-prone annotators, usage of query and annotation types, consideration of imbalanced misclassification and annotation cost, and query-annotator selection.
	\item We identify five unsolved challenges in the \acronym setting and formulate them as future research directions.
\end{itemize}
We structure this survey's main body according to Fig.~\ref{fig:overview} that gives an overview of the main topics reviewed in this survey. The four sections \ref{sec:types_of_cost}--\ref{sec:selection_strategies} are accompanied by a respective tabular literature overview of \acronym strategies, including detailed analyses in this survey's appendices. Based on these literature overviews, we formulate challenges in the setting of \acronym and beyond in Section~\ref{sec:challenges}.  We conclude this survey in Section~\ref{sec:conclusion}.
% In Section~\ref{sec:real_world_active_learning}, we describe the problem setting of \acronym, provide a brief overview of traditional AL and show the structural difference to \acronym.
% A literature overview of query and annotation types is given in Section~\ref{sec:types_of_queries_and_annotations}.
% Then, we discuss existing implementations of \acronym strategies' elements, namely, the query utility and the annotator performance measure, and the selection algorithm, in Section~\ref{sec:query-utility}, Section~\ref{sec:annotator_models}, and Section~\ref{sec:selection_strategies}.
% Finally, we formulate future research directions with reference to \acronym in Section~\ref{sec:challenges} and conclude this survey in Section~\ref{sec:conclusion}.
% Fig.~\ref{fig:overview} gives a graphical content overview of the strategies and topics reviewed in this survey.

\section{Real-world Active Learning}
\label{sec:real_world_active_learning}
In this section, we introduce the problem setting of \acronym.
A real-world application illustrates a possible scenario violating the assumptions of traditional AL and thus indicating the need for \acronym strategies. 
In the context of this application, we also explain the notation used throughout this survey.
Moreover, we formalize the objective of \acronym as the optimal solution to a cost-sensitive classification problem and present learning cycles finding greedy approximations of this solution.

\subsection{Motivating Application}
\begin{figure}[h!]
	\centering
	\includegraphics[width=\columnwidth]{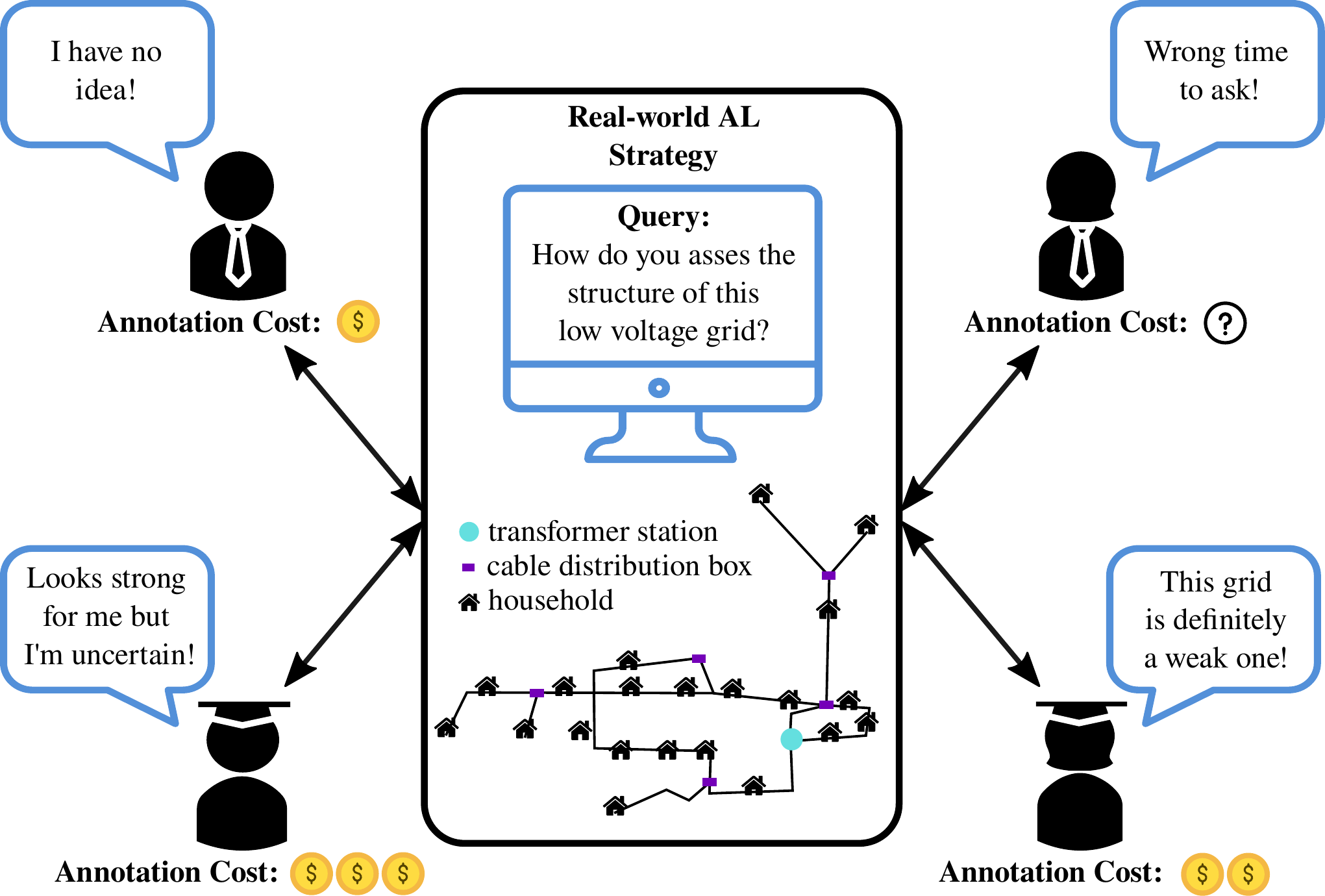}
	\caption{Example of \acronym: The AL strategy queries four human annotators to assess the structure of a low-voltage grid regarding its hosting capacity for distributed generators. Their answers show possible issues when employing AL in real-world applications. The structure of a grid is assessed through ordinal class labels leading to non-uniform misclassification costs. The annotators can have conflicting opinions or may have no time to process a query. An annotator could also be reluctant to answer a query due to its difficulty. Moreover, one annotator may demand more money for an annotation than another. Sometimes the annotation costs are even unknown in advance (e.g., if annotation time is the major cost factor).}
	\label{fig:grid}
\end{figure}

An example of a practical use case requiring the employment of a \acronym strategy is the classification of low-voltage grids described in~\cite{Breker2015,Breker2018}. 
They connect most consumers, e.g., households, to the electrical power system, and an illustration of such a grid is given in Fig.~\ref{fig:grid}. 

Formerly, the power system was designed to transport energy from a few central generators (plants) to the consumers. 
However, recent developments are characterized by an increasing number of installed distributed generators, particularly photovoltaic generators, in low-voltage grids~\cite{Puttgen2003}.
These distributed generators may provoke, e.g., an overload of electrical components, and violate critical voltage values within a grid. 
Assessing the hosting capacity of low-voltage grids for distributed generation supports the responsible distribution system operator in deciding for which low-voltage grid an investment in the infrastructure could be most beneficial such that its sustainable operational reliability is guaranteed~\cite{Breker2015}.

In this context, a significant challenge is the high complexity of low-voltage grids.
Therefore, multiple annotators with heterogeneous background knowledge are requested to provide annotations, such as strong, weak, etc., classifying the hosting capacities of low-voltage grids (cf. Fig.~\ref{fig:grid}).
To do so, the annotators have access to a grid diagram and corresponding tabular information.
The provided annotations, i.e., ordered classes and confidence assessments in this case, are prone to error because of missing expertise, for example.
Moreover, the annotations are expensive because the annotators have to investigate the grids to generate an annotation regarding the hosting capacity. 
A \acronym strategy can save time and money by training a classification model to categorize each possible low-voltage grid's hosting capacity automatically.

As a result, a representative question of this survey would be: \textit{``How to design a \acronym strategy for solving problems such as the classification of low-voltage grids?''}.

\subsection{Formalization of Problem Setting}
%For answering the previously stated question, we need to formalize the \acronym setting.
An instance is described by a  $D$-dimensional feature vector ${\mathbf{x} = (x_{1},\dots,x_{D})^{\mathrm{T}}}, D \in \mathbb{N}$.
It is drawn from the distribution $\Pr(X) = \Pr(X_1, \dots, X_D)$ defined over a $D$-dimensional feature (input) space $\Omega_{X}$, where $X_d$ denotes the random variable of the feature $d \in \{1, \dots, D\}$ and $X$ is used as short-cut for the $D$~dimensional random variable representing all features.
The observed multi-set of identically and independently distributed instances is given by ${\mathcal{X} = \{\mathbf{x}_1,\dots,\mathbf{x}_N\} \subseteq \Omega_{X}}$.
For example, the instance of the low-voltage grid illustrated in~Fig.~\ref{fig:grid} may take the form
\begin{equation}
	\label{eq:feature_vector}
	\mathbf{x}_{n} = \begin{pmatrix} x_{n1} \\ x_{n2} \\ \vdots \\ x_{nD} \end{pmatrix}
	=\begin{pmatrix} \footnotesize{\texttt{\# transformer stations}} \\ \footnotesize{\texttt{\# cable distribution boxes}} \\ \vdots \\ \footnotesize{\texttt{\# house connections}}\end{pmatrix}.
\end{equation}
Each instance $\mathbf{x}_n$ belongs to a true class $y_n \in \Omega_{Y}$ sampled from the categorical distribution $\Pr(Y \mid X=\mathbf{x}_n)$ with $Y$ denoting  the random variable of the true class labels. In total, there are $|\Omega_{Y}| = C \in \mathbb{N}_{\geq 2}$ classes. 
The multi-set of true class labels for the observed instances in $\mathcal{X}$ is denoted as $\mathcal{Y} = \{y_1,\dots,y_N\}$.
%The optimal but unknown classification function to be learned is defined by $y:~\Omega_{X} \rightarrow \Omega_{Y}$, where the set of class labels is given by $\Omega_{Y}~=~\{1,\dots,N\}$. 
Regarding the classification of low-voltage grids, the class labels would have an ordinal structure ranging from a \texttt{very weak}~($Y~=~1~\in~\Omega_{Y}$) to a \texttt{very strong}~($Y~=~5~\in~\Omega_{Y}$) hosting capacity.

As pointed out, there is no omniscient and omnipresent annotator in most applications.
In the context of \acronym, we work with (multiple) error-prone annotators who we summarize in the set~$\mathcal{A} = \{a_1, \dots, a_M\}$.
Each annotator can be queried to provide annotations.
An annotation is not restricted to be a specific class label $y \in \Omega_{Y}$, but all kinds of annotations are allowed, e.g., confidence scores~\cite{Song2018}, probabilistic labels~\cite{Calma2018a}, or rejecting to answer a query~\cite{Zhong2015}.
The space of possible annotations is summarized by the set~$\Omega_{Z}$, e.g., $\Omega_Z = [0, 1]$ if probabilistic class labels are expected for a binary classification problem. The (multivariate) random variable for the annotations of annotator~$a_m$ is denoted as~$Z_m$.

A query cannot only ask for the class label of a specific instance~$\mathbf{x}_n$, but more general queries such as ``Do instance~$\mathbf{x}_n$ and instance~$\mathbf{x}_m$ belong to the same class?'' can be formulated~\cite{Fu2011}. To learn from queries and annotations, a classification model requires appropriate mathematical representations of them. An exemplary representation of the query given above would be $q = \{\mathbf{x}_n, \mathbf{x}_m\}$. The mathematical representations of all possible queries are summarized in a set  $\mathcal{Q}_\mathcal{X}$, which depends on the underlying classification problem and the set of observed instances~$\mathcal{X}$~\cite{Kane2017}. 
Due to this dependency, we can interpret the queries as random events, and $Q$ denotes the associated random variable.
In most cases, a query asks for the class label of a specific instance such that we can define $\mathcal{Q}_\mathcal{X} = \mathcal{X}$ as query set.

The task of a \acronym strategy is to generate a sequence for the execution of the annotation process, which is assumed to consist of countable distinct (time) steps.
In other words, a sequence answers the question: ``Which annotator has to answer which query at which time step?''.
Accordingly, we define a sequence as a function~${\mathcal{S}: \mathbb{N} \rightarrow \mathcal{P}({\mathcal{Q}_\mathcal{X} \times \mathcal{A}}})$, such that ${(q_l, a_m) \in \mathcal{S}(t)}$ induces an annotation of query~$q_l$ by annotator~$a_m$ at time step~${t \in \mathbb{N}}$. The annotation behavior of an annotator can be modeled through a conditional distribution ${\Pr(Z_m \mid Q=q_l, t)}$ from which $z_{lm}^{(t)} \in \Omega_Z$ is drawn as annotation of annotator~$a_m$ for query $q_l$, i.e., ${z_{lm}^{(t)} \sim \Pr(Z_m \mid Q=q_l, t)}$.
As a result, annotators are not compulsorily deterministic in their decisions. Still, decisions might also change throughout the annotation process, e.g., if an annotator gets tired during the annotation process~\cite{Donmez2010}.
% Since a query may be annotated by multiple annotators, we define a vector of annotations per query~$q_l$ after $t$~time steps as
% \begin{align}
%     \mathbf{z}_l^{(t)} &= \left(z_{l1}, \dots, z_{lM}\right)^\mathrm{T}\\ \text{ with } z_{lm} &= \begin{cases} z \in \Omega_Z \text{ if } \exists t^\prime \in \{1, \dots, t\}: (q_l, a_m) \in \mathcal{S}(t^\prime), \\
%     \odot \text{ otherwise.}\end{cases}\nonumber
% \end{align}
% We write $z_{lm} = \odot$ to indicate that an annotator $a_m$ has not annotated query $q_l$ yet. The dependence on $\mathcal{S}$ is omitted to keep the notation uncluttered.
An annotation process executed until the beginning of the time step $t$ according to a sequence $\mathcal{S}$ leads to a data set 
\begin{align}
	\label{eq:data-set}
	\mathcal{D}{(t)} = \left\{(q_l, a_m, z_{lm}^{(t^\prime)})\right. \Bigm| & t^\prime \in \mathbb{N} \wedge \exists t^\prime  < t: (q_l, a_m) \in \mathcal{S}(t^\prime) \nonumber \\  & \left. \wedge \, z_{lm}^{(t^\prime)} \sim \Pr(Z_m \mid Q=q, t^\prime)\right\}
\end{align}
consisting of triplets of a query, an annotator, and an annotation. 
We define the end of a sequence $\mathcal{S}$ as the last time step at which an annotation has been performed, i.e., where the selection is empty:
\begin{equation}
	t_\mathcal{S} = \max(\{t \mid \mathcal{S}(t) \neq \emptyset \wedge t \in \mathbb{N}\}).
\end{equation}

On a data set~$\mathcal{D}{(t)}$, a classification model described by its parameters~$\boldsymbol{\theta}$ can be trained. We denote the resulting parameters of the classification model by~$\boldsymbol{\theta}_{\mathcal{D}{(t)}}$.
For example, these parameters would correspond to weights in the case of a neural network~\cite{Jain1996} taken as a classification model.
The trained classification model predicts class labels for given instances, where the prediction for an instance $\mathbf{x} \in \Omega_X$ is denoted by $\hat{y}(\mathbf{x} \mid \boldsymbol{\theta}_{\mathcal{D}{(t)}}) \in \Omega_Y$.
In many cases, the classification model can predict the class label of an instance and estimate the probabilities of class memberships.
In this case, we denote the estimated class membership probability that a given instance~$\mathbf{x}$ belongs to class $y$ by ${\Pr(Y=y \mid X=\mathbf{x}, \boldsymbol{\theta}_{\mathcal{D}{(t)}})}$.

\subsection{Objective}
Given the formalized problem setting and generalizing the objective definitions in ~\cite{Donmez2008b,Kapoor2007} toward all query and annotation types including complex cost schemes, we formulate the objective of \acronym as determining the optimal annotation sequence for a cost-sensitive classification problem:
\begin{align} 
	\label{eq:objective}
	\mathcal{S}^* = &\argmin\limits_{\mathcal{S} \in \Omega_S} \left[ \text{MC}(\boldsymbol{\theta}_{\mathcal{D}{(t_\mathcal{S} + 1)}} \mid \boldsymbol{\kappa}) + \text{AC}(\mathcal{D}(t_\mathcal{S} + 1) \mid \boldsymbol{\nu})\right] \\
	&\text{ subject to the constraints } \mathcal{C}, \nonumber
\end{align}
where $\Omega_S$ denotes the set of all potential sequences.
MC and AC are the misclassification and annotation cost, respectively.
We expect them to be on the same scale. Otherwise, extra normalization might be necessary. 
The optimal annotation sequence~$\mathcal{S}^*$ minimizes the total cost while satisfying all constraints~$\mathcal{C}$. A common constraint is a maximum annotation budget $B  \in \mathbb{R}_{>0}$, i.e., ${\mathcal{C} = \{\text{AC}(\mathcal{D}(t_\mathcal{S} + 1) \mid \boldsymbol{\nu} )\leq B\}}$. The total cost is decomposed into MC and AC, where the vector~$\boldsymbol{\kappa}$ encodes given hyperparameters for computing the MC, e.g, a cost matrix, and the vector~$\boldsymbol{\nu}$ represents the hyperparameters for computing AC, e.g., wages of the annotators. 
We provide a more detailed discussion on different cost schemes in the setting of \acronym in Section ~\ref{sec:types_of_cost}.

Since it is difficult to find the optimal annotation sequence~$\mathcal{S}^*$ given by Eq.~\ref{eq:objective} in advance~\cite{Donmez2008b}, an AL strategy aims to approximate the optimal solution through a greedy approach.
Therefore, the annotation sequence~$\mathcal{S}$ is defined iteratively at run time by executing a cycle where one iteration corresponds to a single time step.
%Hence, we neglect the explicit notation of step $t$ for ease of notation in the following. 
We start with the description of such a cycle for traditional AL.
Subsequently, we restructure it to fit the setting of \acronym.

\subsection{Traditional Active Learning Cycle}
\label{sec:tradional_pal}
\begin{figure}[h!]
	\centering
	\includegraphics[width=\columnwidth]{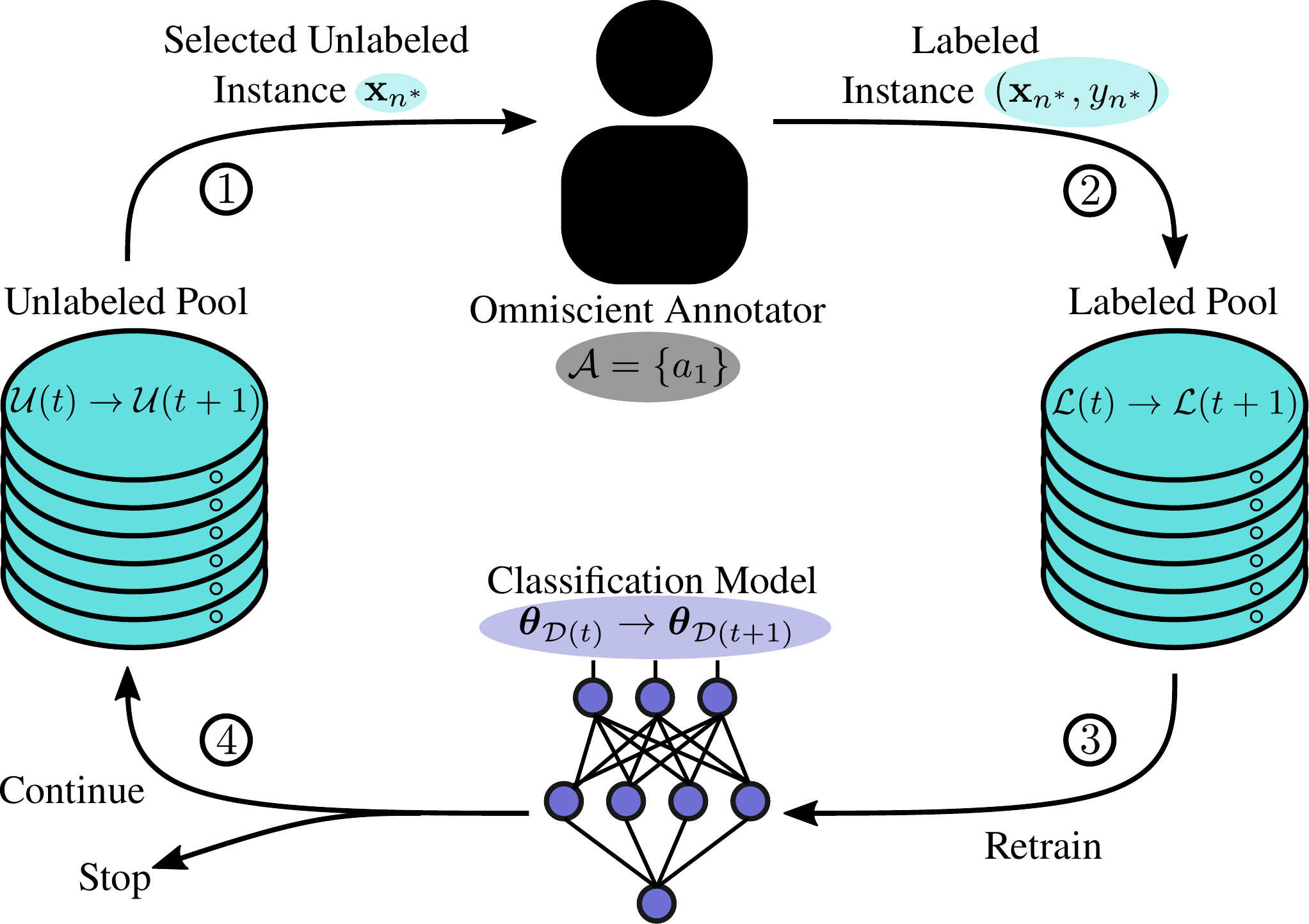}
	\caption{Traditional AL cycle according to~\cite{Settles2010}: (1) At the start of the iteration cycle~$t$, the traditional AL strategy selects an unlabeled instance $\mathbf{x}_{n^*}$ from the unlabeled pool: $\mathcal{U}(t+1) = \mathcal{U}(t) \setminus \{\mathbf{x}_{n^*}\}$. (2) Subsequently, the instance is presented to the omniscient annotator $\mathcal{A} = \{a_1\}$ who provides its true class label $y_{n^*}$. The resulting instance-label pair is inserted into the labeled pool: $\mathcal{L}(t+1) = \mathcal{L}(t) \cup \{(\mathbf{x}_{n^*}, y_{n^*})\}$, (3) on which the classification model is retrained by updating its parameters $\boldsymbol{\theta}_{\mathcal{L}(t)} \rightarrow \boldsymbol{\theta}_{\mathcal{L}(t+1)}$. (4) At the end of the cycle, the traditional AL strategy decides whether to continue or to stop learning. This decision is made by a so-called stopping criterion~\cite{Zhu2010,Altschuler2019,Scharei2018}, which is part of ongoing research and not within this survey's scope.}
	\label{fig:pal_cycle}
\end{figure}

In traditional AL, an omniscient and omnipresent annotator~$\mathcal{A}=\{a_1\}$ is assumed to be available~\cite{Settles2010}.
Moreover, a query expects the class label of an instance such that the set of queries can be represented by $\mathcal{Q}_\mathcal{X} = \mathcal{X}$ and the set of annotations is given by the set of classes, i.e., $\Omega_{Z} = \Omega_{Y}$.
Traditional AL strategies differ between the labeled (annotated) set $\mathcal{L}(t) = \{(\mathbf{x}_n, y_n) \mid (\mathbf{x}_n, a_1, y_n) \in \mathcal{D}(t)\}$ and the unlabeled (non-annotated) set \mbox{$\mathcal{U}(t) = \{\mathbf{x}_n \mid \mathbf{x}_n \in \mathcal{X} \wedge (\mathbf{x}_n, y_n) \notin \mathcal{L}(t)\}$} obtained after executing the \mbox{$(t-1)$-th} iteration cycle. 
The main idea is to develop a strategy intelligently selecting instances from the unlabeled pool $\mathcal{U}(t)$ to which the annotator~$a_1$ assigns true class labels.
Due to the omniscience of this annotator~$a_1$, the annotation distribution satisfies ${\Pr(Z_1 = y_n \mid X=\mathbf{x}_n, t) = 1}$ for all iteration cycles $t \in \mathbb{N}$ and observed instances $\mathbf{x}_n \in \mathcal{X}$.
Fig.~\ref{fig:pal_cycle} summarizes the entire selection procedure as a cycle.

The selection of an instance is based on a so-called utility measure $\phi: \mathcal{X} \rightarrow \mathbb{R}$~\cite{Fu2013} estimating the utilities of the observed instances $\mathcal{X}$ regarding the classification model to be trained.
In general, the unlabeled instance with the maximum utility is selected in iteration cycle $t$: 
\begin{equation}
    \label{eq:selection}
	\mathbf{x}_{n^*} = \argmax_{\mathbf{x}_n \in \mathcal{U}(t)} \left[\phi(\mathbf{x}_n \mid \boldsymbol{\theta}_{\mathcal{L}(t)})\right].
\end{equation}
There are many approaches computing instances' utilities.
In the following, we briefly describe two fundamental concepts:
\begin{itemize}
	\item The simplest concept of utility measures is \textit{uncertainty sampling} (US)~\cite{Lewis1994}, which usually requires an instance's class membership probabilities estimated by the classification model to be trained. 
	Alternatively, distances to decision boundaries~\cite{Tong2002} are used as proxies of them.
	US ranks all instances in the unlabeled pool~$\mathcal{U}({t})$ based on an uncertainty measure and queries the label for the instance with the maximum uncertainty regarding its class information. 
	A common uncertainty measure is the entropy $H$~\cite{Shannon1948} of the class distribution such that an instance's utility estimated is computed as
	\begin{equation}
	    \label{eq:us}
		\phi_{\text{US}}(\mathbf{x}_n \mid \boldsymbol{\theta}_{\mathcal{L}({t})}) = H[\Pr(Y \mid X=\mathbf{x}_n, \boldsymbol{\theta}_{\mathcal{L}({t})})].
	\end{equation}
	\item The decision-theoretic framework \textit{expected error reduction} (EER)~\cite{Roy2001} estimates the performance of the classification model. 
	Therefor, EER assumes that the instances in the unlabeled pool $\mathcal{U}({t})$ form a validation set. 
	For each unlabeled instance, the classification model's expected error is computed on this validation set by retraining the classification model with each combination of the given unlabeled instance and its possible class label.
	The multiple retraining procedures of the classification model lead to high computational complexity.
	The resulting estimate of the negative expected error defines the utility measure. 
	Correspondingly, EER selects the instance leading to the minimum estimated error.
% 	\item The utility measure concept named \textit{query-by-committee} (QBC)~\cite{Gilad-Bachrach2005} maintains a committee of classification models that are trained on different subsets of the labeled pool $\mathcal{L}({t})$. 
% 	As a result, they may produce competing classification hypotheses.
% 	The unlabeled instance with maximum disagreement among the committee's classification models is considered the most useful and selected for annotation.
\end{itemize}
One of the main challenges regarding the design of utility measures is the exploration-exploitation trade-off.
On the one hand, we aim to select instances near the classification model's decision boundary to refine it (exploitation).
On the other hand, we aim to select instances in unknown regions (exploration)~\cite{Osugi2005}.
More advanced AL strategies balance this trade-off by considering distances to the decision boundaries, density, class distribution estimates~\cite{Donmez2007,Reitmaier2013,Calma2018b}, or using a Bayesian approach~\cite{Kottke2021}. 

In batch mode AL~\cite{Hoi2006}, we must consider the diversity of instances since a batch of instances is selected in each learning iteration cycle.
%In this case, the potential query set $\mathcal{Q}_\mathcal{X}$ would consist of all subsets of observed instances $\mathcal{X}$ with fixed batch size.
However, a detailed analysis of instance utility measures in the traditional AL setting is beyond the scope of this survey, and a more detailed discussion on them is given in~\cite{Settles2010,Aggarwal2014,Fu2013,Kumar2020}.

\subsection{Real-world Active Learning Cycle}
\begin{figure}[h!]
	\centering
	\includegraphics[width=\columnwidth]{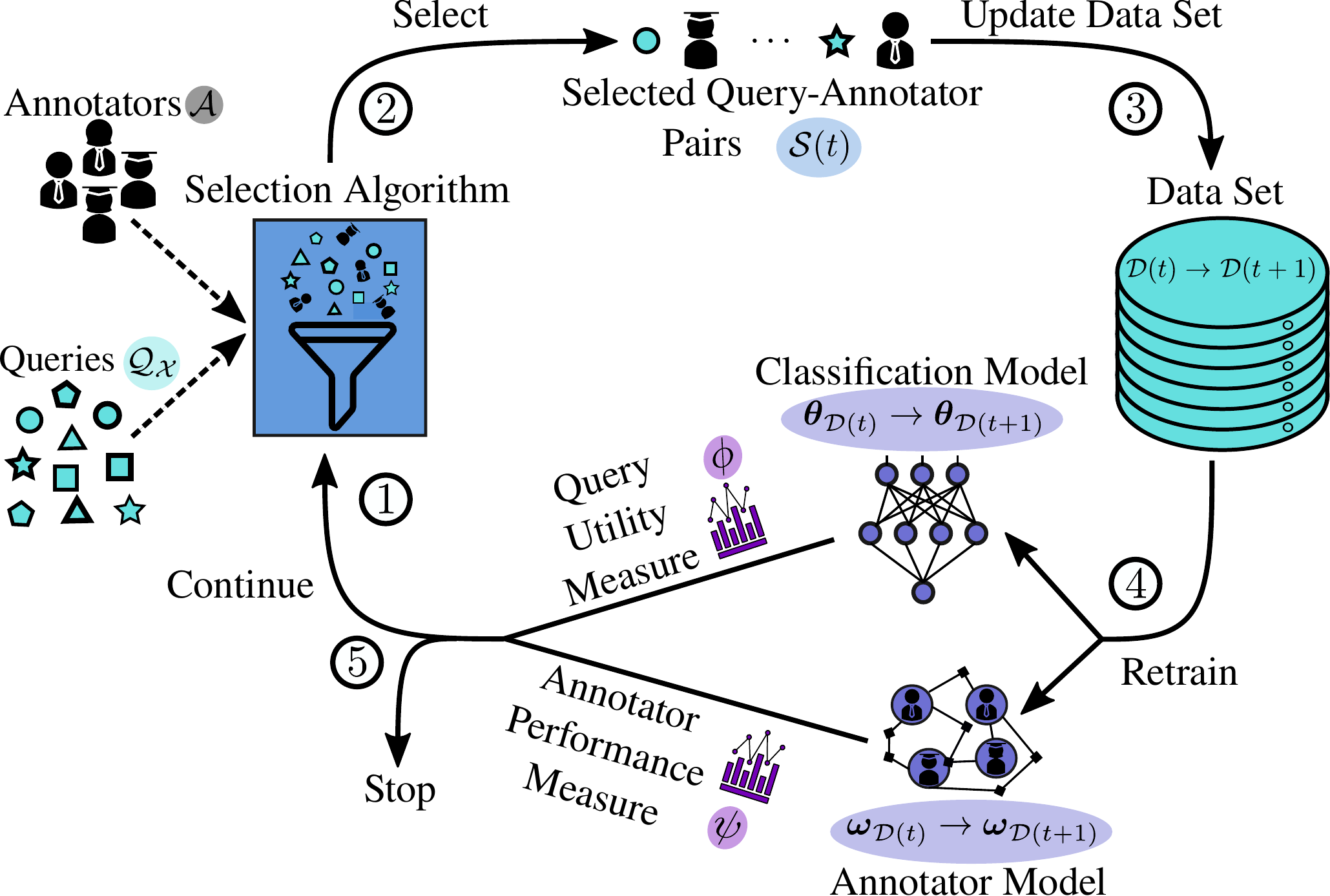}
	\caption{Proposed \acronym cycle: (1)~At the start of the iteration cycle~$t$, the classification and annotator model, both trained on the current data set~$\mathcal{D}(t)$, provide information regarding the query utility and the annotator performance measure of the \acronym strategy. (2)~Based on both measures, the \acronym strategy's selection algorithm specifies a set of query-annotator pairs $\mathcal{S}(t) \subseteq \mathcal{Q}_\mathcal{X} \times \mathcal{A}$.
	Each pair ${(q_l, a_m) \in \mathcal{S}(t)}$ initiates an annotation of query $q_l$ by annotator $a_m$. (3)~The annotations are inserted into the data set: $\mathcal{D}(t+1) = \mathcal{D}(t) \cup \{(q_l, a_m, z_{lm}^{(t)}) \mid (q_l, a_m) \in \mathcal{S}(t)\}$. (4)~Then, the classification and annotator model are retrained on the updated data set ($\boldsymbol{\theta}_{\mathcal{D}(t)} \rightarrow \boldsymbol{\theta}_{\mathcal{D}(t+1)}$ and $\boldsymbol{\omega}_{\mathcal{D}(t)} \rightarrow \boldsymbol{\omega}_{\mathcal{D}(t+1)}$). (5)~At the end of the iteration, the \acronym strategy decides whether to stop or to continue learning.}
	\label{fig:active_learner}
\end{figure}

The traditional AL cycle depicted in Fig.~\ref{fig:pal_cycle} has to be adjusted to fit the setting of \acronym.
%Furthermore, these adjustments lead to additional requirements.
%In this survey, we frame a \acronym strategy embedded in a learning system.
Our resulting cycle, including the \acronym strategy's elements (i.e., query utility measure, annotator performance measure, and selection algorithm), is shown in Fig.~\ref{fig:active_learner}.

The \textbf{query utility measure} $\phi: \mathcal{Q}_\mathcal{X} \rightarrow \mathcal{R}_\phi$ is an element being already part of the traditional AL setting.
However, in the \acronym setting, not only can the class labels of non-annotated (unlabeled) instances be queried, but more general queries can be selected for annotation.
This also includes a re-annotation of instances, known as repeated labeling~\cite{Ipeirotis2014}, re-labeling~\cite{Lin2016a}, or backward instance labeling~\cite{Zhang2015}.
Hence, the strict distinction into a non-annotated (unlabeled) set~$\mathcal{U}(t)$ and an annotated (labeled) set $\mathcal{L}(t)$ is often not adequate anymore.
As a result, the utility measure~$\phi$ needs to be adapted to quantify the utility of more general queries.
Another adaption concerns the form of the output of the utility measure. 
Instead of computing a single score per query, i.e., $\mathcal{R}_\phi \subseteq \mathbb{R}$, a utility measure may provide a more general description for each query, e.g., a distribution, which can then be combined with annotator performance estimates~\cite{Yan2012b,Herde2021}.
We provide an overview of query utility measures for different query and annotation types in Section~\ref{sec:types_of_queries_and_annotations}.

The \textbf{annotator performance measure} $\psi: \mathcal{Q}_\mathcal{X} \times \mathcal{A} \rightarrow \mathcal{R}_\psi$ represents a novel element compared to traditional AL strategies and is defined through an annotator model.
Similar to a classification model, an annotator model has parameters~$\boldsymbol{\omega}_{\mathcal{D}(t)}$ learned from a data set $\mathcal{D}(t)$.
Its main task concerns the estimation of the performance~${\psi(q_l, a_m \mid \boldsymbol{\omega}_{\mathcal{D}(t)})} \in \mathcal{R}_\psi$ of an annotator $a_m$ regarding a query~$q_l$~\cite{Yan2012b}, e.g., the probability for providing a correct annotation.
In most cases, ${\psi(q_l, a_m \mid \boldsymbol{\omega}_{\mathcal{D}(t)})}$ is a point estimate, i.e., $\mathcal{R}_\psi \subseteq \mathbb{R}$,  but there are also annotator models estimating probability distributions over annotator performances~\cite{Yan2012b,Herde2021}.
Moreover, an annotator model may account for improvements and deteriorations of annotators' performances, e.g., when an annotator learns or gets exhausted.
The annotator performance may also be affected by collaboration mechanisms between the annotators, e.g., the best annotator is asked to teach the worst annotator~\cite{Fang2012}.
We provide an overview of annotator performance measures in Section~\ref{sec:annotator_models}.

A \acronym strategy is completed by the \textbf{selection algorithm} as the final element.
It updates the annotation sequence $\mathcal{S}$ by selecting query-annotator pairs in each iteration cycle~$t$.
This selection is specified by choosing a subset of query-annotator pairs ${\mathcal{S}(t) \subseteq \mathcal{Q}_\mathcal{X} \times \mathcal{A}}$.
Therefor, it assesses potential query-annotator pairs through the query utility and annotator performance measure.
If the set $\mathcal{S}(t)$ contains multiple queries, we face similar challenges as in batch mode AL, e.g., selecting diverse queries.
We provide an overview of selection algorithms in Section~\ref{sec:selection_strategies}.

AC and MC are modeled in AL literature by designing cost-sensitive variants of query utility measures, annotator performance measures, or selection algorithms. We provide an overview in Section~\ref{sec:types_of_cost}.

\section{Cost Types}
\label{sec:types_of_cost}
MC and AC are the most crucial cost types in the real-world setting, and we will summarize typical schemes of them in this section.
There exist several additional types of cost when solving a classification problem, e.g., cost of computation (e.g., renting a graphics processing unit) and cost of test (e.g., getting the results of a blood test). They are described as a taxonomy in~\cite{Turney2002}.
At the end of this section, we present a literature overview of \acronym strategies explicitly modeling MC and/or AC.

\subsection{Misclassification Cost}
\label{subsec:misclassification_cost}
Mistakes of the classification model induce MC (the first summand in Eq.~\ref{eq:objective}). In the literature, we identified three cost schemes and describe them in increasing order complexity in the following:

\textbf{Uniform MC:} Each classification error is charged at an equal cost. The classification model's performance is inversely proportional to the misclassification rate~\cite{Seliya2009}, i.e., the proportion of misclassified instances. This cost scheme is the simplest one and is assumed by traditional AL strategies. 

\textbf{Class-dependent MC:} This cost scheme is probably the most common one in cost-sensitive classification~\cite{Zhou2010, Elkan2001}. The cost of a classification error is defined by means of a cost matrix/table~$\mathbf{C} \in \mathbb{R}_{\geq 0}^{C \times C}$, where an entry~$\mathbf{C}[y, y^\prime]$ in row~$y$ and column~$y^\prime$ denotes the cost of predicting the class label~$\hat{y}(\mathbf{x}_n \mid \boldsymbol{\theta}_{\mathcal{D}(t)})=y^\prime$, when the instance $\mathbf{x}_n$ actually belongs to class~$y_n = y$.
Our grid classification example could use the mean absolute error on class numbers as a typical cost measure for ordinal classes~\cite{Baccianella2009}. 
It would be implemented through $\mathbf{C}[y, y^\prime]=|y-y^\prime|$.
In some applications, the cost matrix is extended by adding an extra column representing cases where the classification model is too uncertain and rejects predicting a class label (known as reject option~\cite{Bishop2006}).

\textbf{Instance-dependent MC:} Costs of classification errors depend on specific characteristics of instances. An example is fraud detection, where the amount of money involved in a particular case has an essential impact on MC~\cite{Chan1999}. 
For our grid classification example, it would be more expensive if many households were affected by overloading a low-voltage grid.
Consequently, the feature \texttt{\# house connections} in Eq.~\ref{eq:feature_vector} is to play a central role when computing the cost of misclassifying a grid. 

MC can be computed as the expectation regarding the true (but unknown) joint distribution $\Pr(X, Y)$ of instances and class labels~\cite{Bishop2006}. For example, class-dependent MC is computed according to
\begin{equation}
	 \text{MC}(\boldsymbol{\theta}_{\mathcal{D}(t)} \mid \mathbf{C}) = \E_{\Pr(X=\mathbf{x}, Y=y)}\left[\mathbf{C}[y, \hat{y}(\mathbf{x}  \mid  \boldsymbol{\theta}_{\mathcal{D}(t)}]\right].
\end{equation}
In practice, the exact computation of MC is often infeasible due to the limited size of test data.
Furthermore, in the \acronym setting, its estimation based on a separate set of instances is challenging because of a sampling bias (arising from the active data acquisition)~\cite{Kottke2019} and the lack of known ground truth class labels (arising from the error-proneness of the annotators).
Nevertheless, some \acronym strategies take imbalanced, i.e., class- or instance-dependent, MC into account.

\subsection{Annotation Cost}
\label{subsec:annotation_cost}
AC (the second summand in Eq.~\ref{eq:objective}) arises from the work effort of the annotators who have to invest time to decide on appropriate annotations for the posed queries.
The exact specification of AC depends on the underlying cost scheme.
In the literature, we identified four different schemes and describe them in increasing order of complexity in the following:

\textbf{Uniform AC}: The cost of obtaining an annotation is constant for each query and independent of the queried annotator. Correspondingly, the AC is proportional to the number of acquired annotations. This cost scheme is the simplest one and is frequently used. In particular, it is often employed in crowdsourcing environments, where the requester sets a constant pay rate per query. This means the qualification of an annotator and the time spent on annotating a query have no impact on the AC.

\textbf{Annotator-dependent AC}: In this cost scheme, the AC explicitly depends on the queried annotator. This setting is typical when annotators with different qualifications receive different earnings per query, e.g., annotators with different levels of expertise. 
Of course, there is typically no guarantee that expensive annotators provide more accurate annotations~\cite{Zheng2010}.

\textbf{Query-dependent AC}: Since there may be more or less difficult queries, the cost of annotating a query may depend on the query itself. For example, assessing the hosting capacity of a large and complex low-voltage grid may require more time than assessing a small and simple grid. 
Another example is the annotation of voice mails, where the duration of a voice mail is used as a proxy of the AC, e.g., 0.01 US dollar per second~\cite{Kapoor2007}.
For the classification of documents, the number of words or characters in a document is often correlated to the AC~\cite{Arora2009}.
Additionally, the query type affects the AC.
For example, comparing two instances and deciding whether both belong to the same class is often easier than assigning an instance to one of many classes~\cite{Joshi2010,Joshi2012}.

\textbf{Query- and annotator-dependent AC}: If the query and annotator-dependent cost schemes are considered, the AC varies across the pairs of query and annotator~\cite{Settles2008a}.
This cost scheme fits scenarios in which annotators are paid according to their individual hourly wages and the annotation time depends on the query~\cite{Wallace2011}.

The exact computation of AC depends on the underlying scheme.
If we exemplary assume annotator-dependent AC with $\boldsymbol{\nu} = \{\nu_1, \dots, \nu_M\}$ and $\nu_m > 0$ representing the payment per query for annotator~$a_m$, we would obtain 
\begin{align}
    \text{AC}(\mathcal{D}(t) \mid \boldsymbol{\nu}) = \sum_{m=1}^M \nu_m \cdot N_m^{(t)}, \\
    \label{eq:num-annotations}
    N_m^{(t)} = \sum_{(q, a, z) \in \mathcal{D}(t)} \delta(a \doteq a_m),
\end{align}
where $\doteq$ denotes a Boolean comparison and the indicator function~${\delta: \{\text{false}, \text{true}\} \rightarrow \{0, 1\}}$ returns one if the argument is true and zero otherwise. Correspondingly, $N_m^{(t)} \in \mathbb{N}$ is the number of annotations provided by annotator $a_m$ until the start of step $t$.
In certain scenarios, such an exact specification of the AC is infeasible. This is when the annotation time is the major cost factor and is not known in advance.
Therefore, an AL strategy is required to estimate the AC before querying an annotator.

\subsection{Literature Overview}
Table~\ref{tab:cost-sensitive-al} gives a literature overview of \acronym strategies, explicitly modeling imbalanced MC or AC.
The first part of this table lists strategies being MC-sensitive, i.e., class-dependent or instance-dependent. 
The second part summarizes strategies taking imbalanced AC into account, i.e., annotator- and/or query-dependent.
Each strategy is categorized according to its cost scheme, the type of classification problem (binary vs. multi-class), and its predefined or estimated required cost information (cost matrix, annotation time, etc.).
Additionally, we provide a brief description of each strategy's main idea.
A more in-depth analysis of them is provided in the appendices of this survey.

\begin{table*}[!p] \centering
    \caption{Literature overview of cost-sensitive \acronym strategies.}
    \label{tab:cost-sensitive-al}
    \footnotesize
    \begin{tabularx}{\textwidth}{|L{0.16\textwidth}|p{0.21\textwidth}|p{0.15\textwidth}|p{0.38\textwidth}|}
        \toprule
        \multicolumn{1}{|c|}{\textbf{Strategy}} & \multicolumn{1}{c|}{\textbf{Cost Scheme}} & \textbf{Classification Problem} & \multicolumn{1}{c|}{\textbf{Cost Information}}\\
        \midrule
        \hline
        \rowcolor{Gray}
        \multicolumn{4}{|c|}{{Misclassification Cost (MC)}} \\ 
        \hline
        \multirowcell{3}[0pt][l]{\citet{Margineantu2005}, \\ \citet{Joshi2010,Joshi2012}} & class-dependent MC & multi-class & cost matrix (predefined)\\
        \cline{2-4} 
        & \multicolumn{3}{X|}{These strategies compute the expected MC on the annotated set. Therefor, they simulate the annotation of an instance and its                             addition to the classification model's training set. We can interpret this approach as a cost-sensitive variant of EER.} \\
        \hline
        \multirowcell{3}[0pt][l]{\citet{Liu2009}} & class-dependent MC & multi-class & cost matrix  (predefined)\\
        \cline{2-4} 
        & \multicolumn{3}{X|}{This strategy extends traditional US by making use of self-supervised training. Therefor, a cost-sensitive classification model is trained on instances annotated by annotators and instances annotated by a cost-insensitive classification model. } \\
        \hline
        \multirowcell{3}[0pt][l]{\citet{Chen2013}} & class-dependent MC & multi-class & cost matrix  (predefined)\\
        \cline{2-4} 
        & \multicolumn{3}{X|}{The first variant of this strategy computes the maximum expected MC of an instance. In contrast, the second variant computes the cost-weighted minimum margin between the two predictions with the lowest estimated MCs.} \\
        \hline
        \multirowcell{3}[0pt][l]{\citet{Krempl2015}} & class-dependent MC & binary & cost ratio of false negative vs. false positive  (predefined)\\
        \cline{2-4} 
        & \multicolumn{3}{X|}{This strategy computes the density-weighted expected MC reduction in an instance's neighborhood within the feature space. Therefor, it simulates the annotation of an instance and its addition to the classification model's training set.} \\
        \hline
        \multirowcell{2}[0pt][l]{\citet{Kaeding2015}} & class-dependent MC & multi-class & cost function (predefined)\\
        \cline{2-4} 
        & \multicolumn{3}{X|}{This strategy computes the classification model's expected change by simulating the annotation of an instance.} \\
        \hline
        \multirowcell{3}[0pt][l]{\citet{Nguyen2015}} & class-dependent MC & binary & cost matrix (predefined) \\
        \cline{2-4} 
        & \multicolumn{3}{X|}{This strategy computes the expected MC reduction when obtaining an annotation from an error-prone annotator and from an infallible expert. Therefore, it employs a cost-sensitive variant of EER.} \\
        \hline
        \multirowcell{3}[0pt][l]{\citet{Huang2016a}} & class-dependent MC & multi-class & cost matrix  (predefined)\\
        \cline{2-4} 
        & \multicolumn{3}{X|}{This strategy is based on a cost embedding approach, which transfers the MC information into a distance measure of a latent space. Utilities are defined as expected MCs that are represented through distances in the latent space.} \\
        \hline
        \multirowcell{3}[0pt][l]{\citet{Min2019}} & class-dependent MC & multi-class & cost matrix (predefined)\\
        \cline{2-4} 
        & \multicolumn{3}{X|}{This strategy queries only annotations for instances with MCs being higher than their respective ACs. The MCs are estimated through a cost-sensitive $k$-nearest neighbor model.} \\
        \hline
        \multirowcell{3}[0pt][l]{\citet{Wu2019}, \\ \citet{Wang2019}} & class-dependent MC & binary & cost matrix (predefined)\\
        \cline{2-4} 
        & \multicolumn{3}{X|}{These strategies employ a density-based clustering technique to construct a master tree of instances. In an iterative process, this master tree is subdivided into blocks and for each block an estimated MC-optimal number of instances are annotated. } \\
        \hline
        \multirowcell{3}[0pt][l]{\citeauthor{Krishnamurthy2017} \\ \cite{Krishnamurthy2017,Krishnamurthy2019}} & instance-dependent MC & multi-class & cost of predicting a class label for an instance (estimated) \\
        \cline{2-4} 
        & \multicolumn{3}{X|}{These strategies query MC information per instance and class from an annotator. Their idea is to query the actual MC information for the class label, for which an instance has the largest estimated MC range.} \\
        \hline
        \rowcolor{Gray}
        \multicolumn{4}{|c|}{{Annotation Cost (AC)}} \\
        \hline
        \multirowcell{2}[0pt][l]{\citet{Zheng2010}, \\ \citet{Chakraborty2020}} & annotator-dependent AC & mutli-class & cost of querying an annotator (predefined) \\
        \cline{2-4} 
        & \multicolumn{3}{X|}{These strategies solve an optimization problem to specify a subset of annotators with low ACs and high performances.} \\
        \hline
        \multirowcell{2}[0pt][l]{\citeauthor{Moon2014} \\ \cite{Moon2014}, \citet{Huang2017}} & annotator-dependent AC & multi-class & cost of querying an annotator (predefined) \\
        \cline{2-4} 
        & \multicolumn{3}{X|}{These strategies compute the annotator performance per AC unit to prefer annotators with high performances and low ACs.} \\
        \hline
        \multirowcell{3}[0pt][l]{\citet{Nguyen2015}} & annotator-dependent AC & multi-class & cost ratio of querying crowd worker vs. expert (predefined) \\
        \cline{2-4} 
        & \multicolumn{3}{X|}{This strategy normalizes the utility of querying an expert or crowd worker by their respective ACs to find a trade-off between expensive but correct expert annotations and cheap but error-prone crowd worker annotations. } \\
        \hline
        \multirowcell{3}[0pt][l]{\citet{Margineantu2005}, \\ \citeauthor{Donmez2008b} \\ \cite{Donmez2008b,Donmez2010b}} & query-dependent AC & multi-class & cost of annotating a query (predefined) \\
        \cline{2-4} 
        & \multicolumn{3}{X|}{These strategies subtract the query's individual AC from its utility to prefer highly useful queries with low ACs. These ACs are assumed to be known in advance for each query or to follow a predefined model.} \\
        \hline
        \multirowcell{3}[0pt][l]{\citet{Joshi2010,Joshi2012}} & query-dependent AC & multi-class & number of comparisons per query (estimated) \\
        \cline{2-4} 
        & \multicolumn{3}{X|}{These strategies use multiple comparison queries to reveal the class label of a non-annotated instance. The expected number of comparisons required to reveal an instance's class label is used as an AC proxy and subtracted from an instance's utility.} \\
        \hline
        \multirowcell{3}[0pt][l]{\citet{Tsou2019}} & query-dependent AC & multi-class & cost of annotating a query (predefined) \\
        \cline{2-4} 
        & \multicolumn{3}{X|}{This strategy builds a decision tree throughout the AL process. It computes the average AC of the already annotated instances in each leaf of this tree. These AC estimates are used to normalize the query utilities of the instances in the respective leaves.} \\
        \hline
        \multirowcell{3.6}[0pt][l]{\citet{Settles2008a}, \\ \citet{Haertel2008}, \\ \citet{Tomanek2010}, \\ \citet{Wallace2010}} & query-dependent AC & multi-class & annotation time per query (estimated) \\
        \cline{2-4} 
        & \multicolumn{3}{X|}{These strategies estimate the annotation time per query. Therefore, they use either historical data in form of logged annotation times or employ prior knowledge regarding a domain, e.g., the number of words when annotating text. The estimated annotation times are  considered by normalizing query utility or employing a linear rank combination of utilities and annotation times.} \\
        \hline
        \multirowcell{3.8}[0pt][l]{\citet{Wallace2011}} & \multirowcell{2}[0pt][l]{annotator-, query-dependent AC} & \multirowcell{2}[0pt][l]{multi-class} & annotation time per query (estimated) + cost per time unit for each annotator (predefined)\\
        \cline{2-4} 
        & \multicolumn{3}{X|}{This strategy computes ACs by multiplying the annotation times (estimated through the number of words in a document) with the respective salaries (predefined) of the annotators. Annotators with low estimated ACs are more often queried. } \\
        \hline
        \multirowcell{3}[0pt][l]{\citet{Arora2009}} & annotator-, query-dependent AC & multi-class & annotation time per query-annotator pair (estimated) \\
        \cline{2-4} 
        & \multicolumn{3}{X|}{This strategy estimates the annotation time as a function of the annotator and the query. Therefor, it uses features to describe the query and the annotator in combination with historical data in form of logged annotation times.} \\
        \hline
        \bottomrule
    \end{tabularx}
\end{table*}

\section{Interaction Schemes}
\label{sec:types_of_queries_and_annotations}
Interaction with human annotators forms an essential part of AL.
In this survey, we focus on the AL typical query-annotation-based interaction. 
For this purpose, we provide an overview of different query types and annotation types based on the literature.
At the end of this section, we present a literature overview of existing \acronym strategies using different combinations of queries and annotations as interaction schemes.

\subsection{Query Types}
\label{subsec:query_types}
\begin{figure*}[h!]
	\centering
	\subfloat[Example of an instance query.]{
		\includegraphics[width=0.32\textwidth]{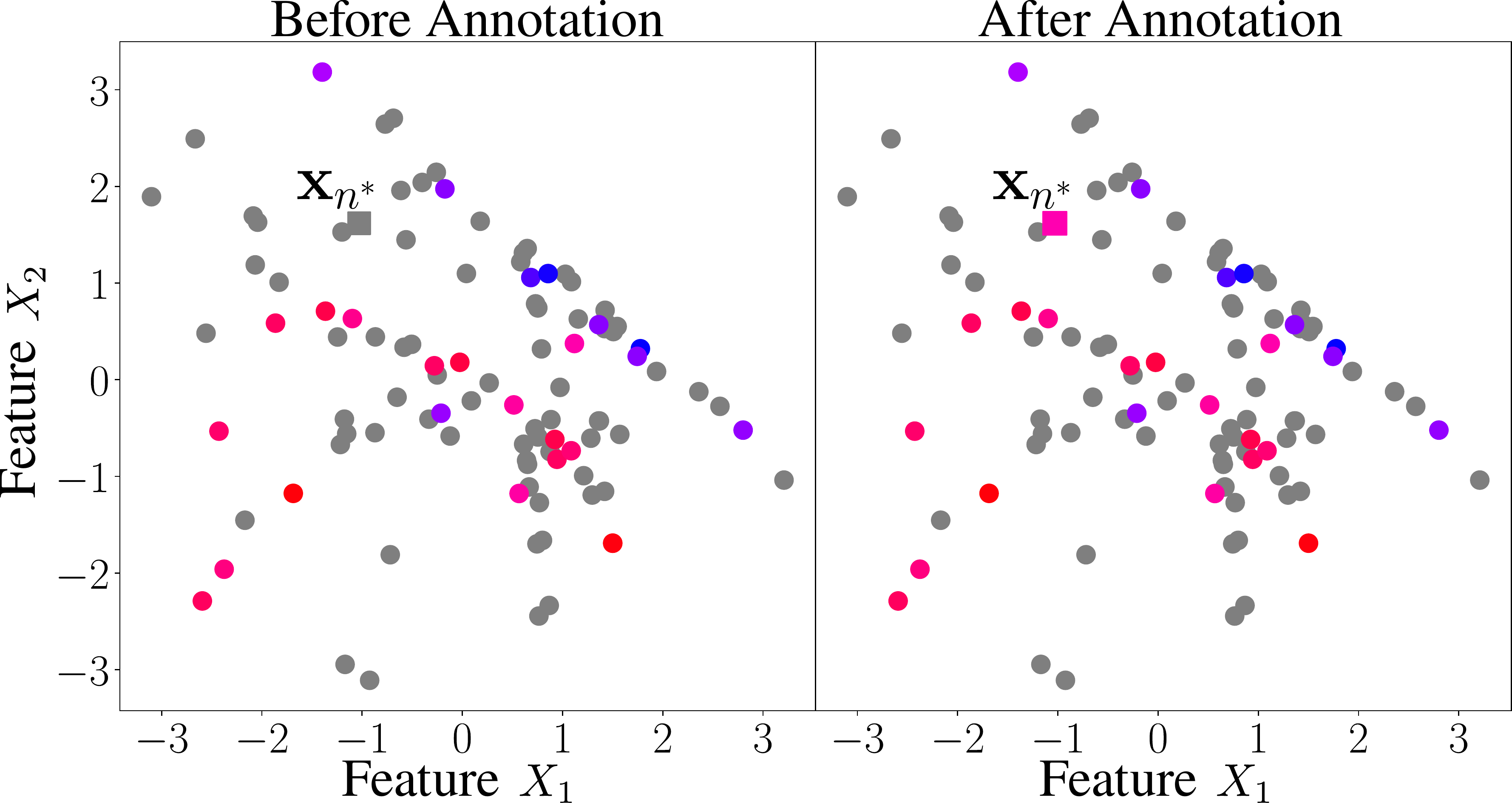}}\label{fig:instance_based_query}
	\subfloat[Example of a region query.]{
		\includegraphics[width=0.32\textwidth]{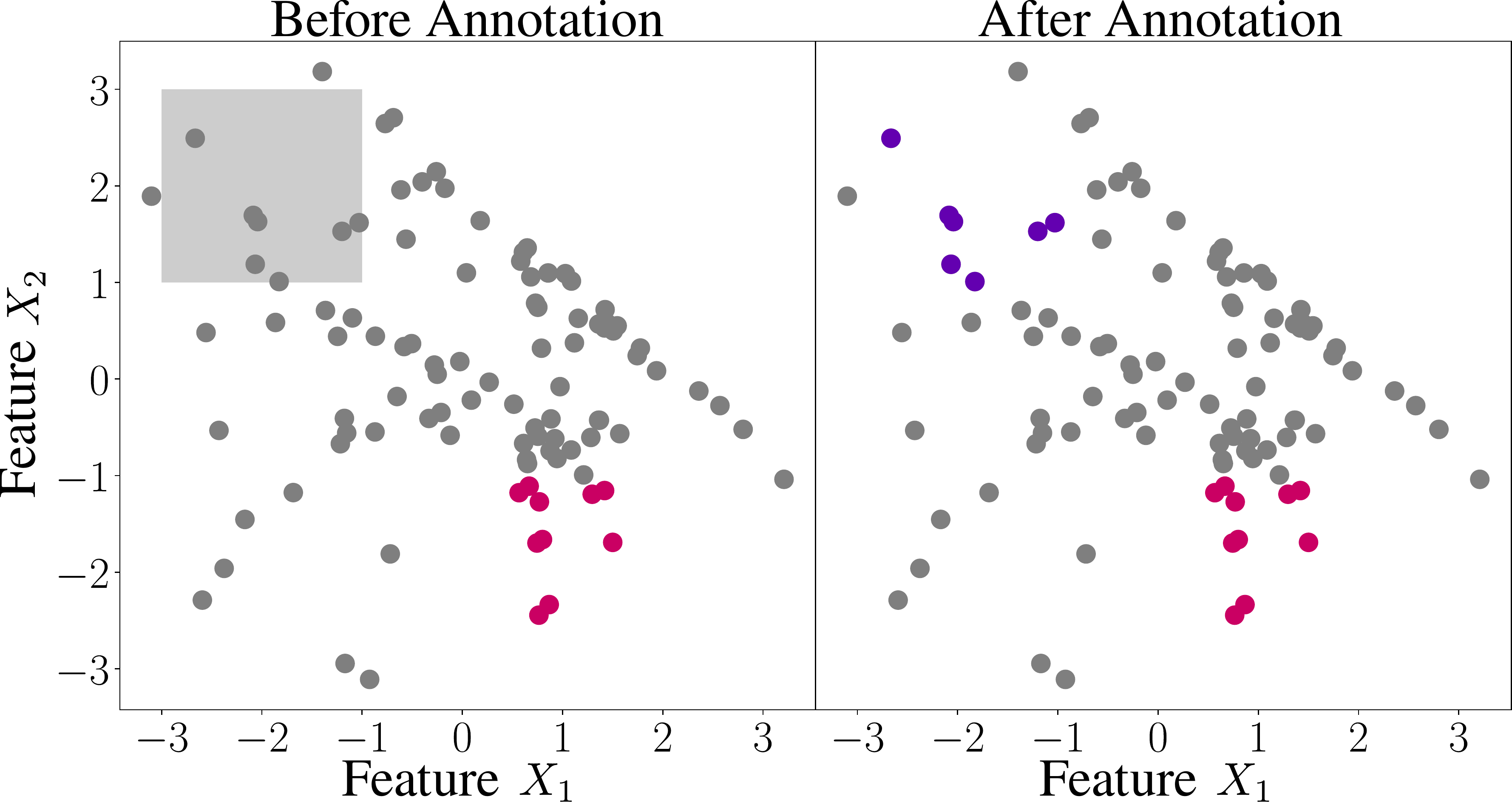}}\label{fig:region_based_query}
	\subfloat[Example of a comparison query.]{
		\includegraphics[width=0.32\textwidth]{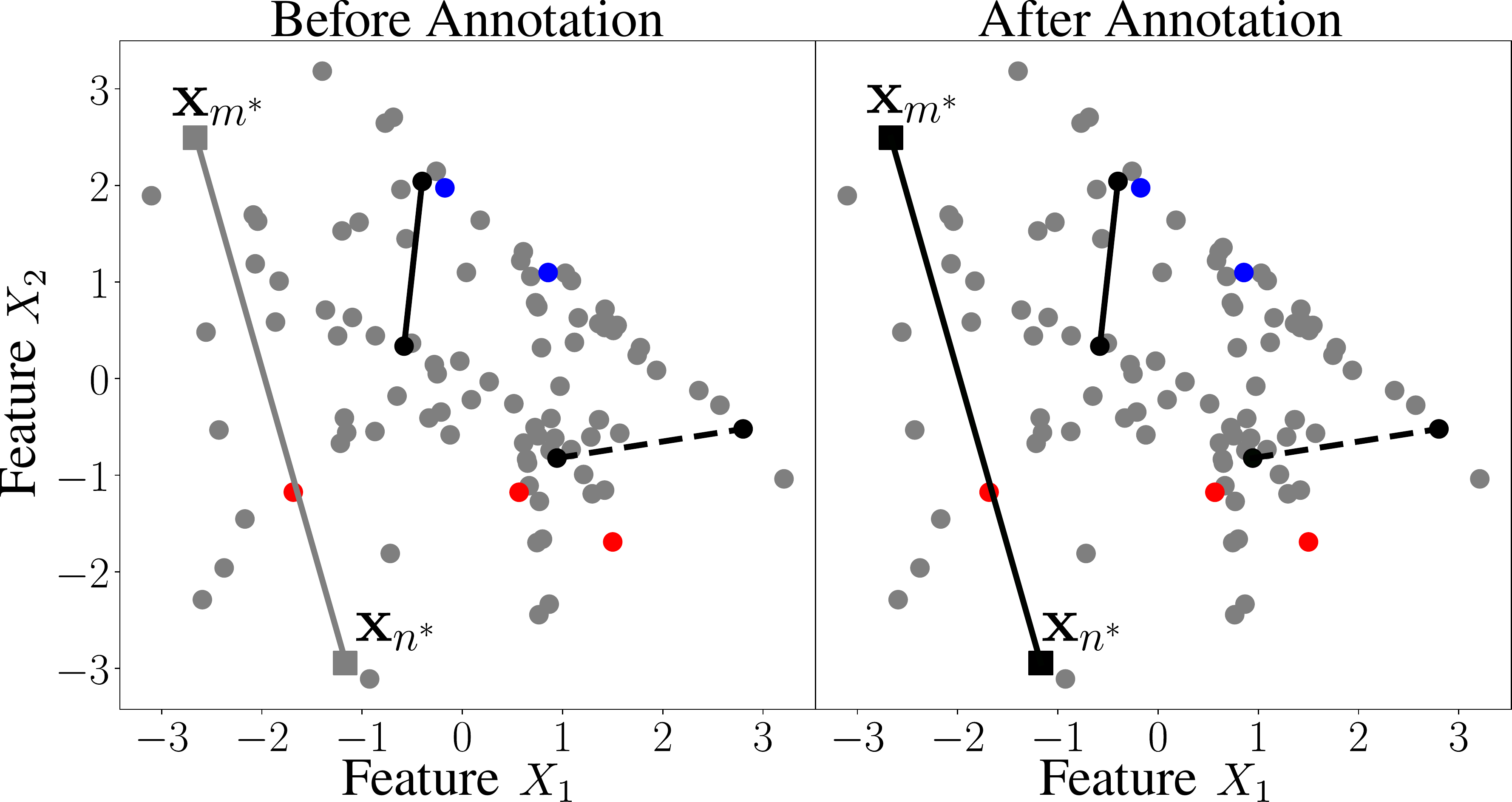}}\label{fig:comparison_based_query}
	
	\vspace*{-0.5em}
	\subfloat{\includegraphics[width=0.45\textwidth]{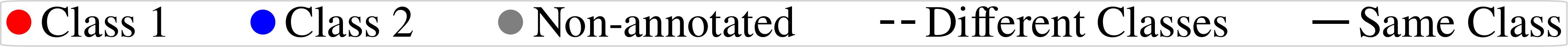}}\label{fig:query_types_legend}
	
	\caption{Illustration of query types within the feature space: For a binary classification problem, the two-dimensional instances of an artificially generated set~$\mathcal{X} \subset \mathbb{R}^2$ are plotted according to their feature values. Probabilistic annotations are depicted by using the corresponding proportions of the red and blue colors. Figure~\ref{fig:query_types}(a) illustrates an instance query by marking the selected instance $\mathbf{x}_{n^*}$ for which the class membership probability for the blue class is expected as an annotation. A region query defining the region $X_1 \in [-3, -1] \wedge X_2 \in [1, 3]$ is depicted by the gray rectangle in Fig.~\ref{fig:query_types}(b). Again, the class membership probability for the blue class represents the annotation. Figure~\ref{fig:query_types}(c) illustrates a comparison query requesting whether the instances~$\mathbf{x}_{n^*}$ and $\mathbf{x}_{m^*}$ belong to the same class (solid gray line). A solid black line connects instances belonging to the same class, and a dashed black line indicates that instances are assigned to different classes.}
	\label{fig:query_types}
\end{figure*}

The set of possible queries $\mathcal{Q}_\mathcal{X}$ specifies how a \acronym strategy can interact with the available annotators~$\mathcal{A}$.
Depending on the underlying classification problem, there are different possibilities to design these queries.
In the literature, we identified the following three most common query types:

\textbf{Instance queries} ask for information on a specific instance~$\mathbf{x}_n$ as illustrated in Fig.~\ref{fig:query_types}(a) and is the most common query type. Next to class labels, a query may request additional information. Concrete examples are presented in~\cite{Calma2018a,Ni2012}, where annotators are asked for confidence scores interpreted as proxies of an instance's class membership probabilities.

\textbf{Region queries} do not query information regarding a specific instance, but ask annotators to provide information about an entire region in the feature space~\cite{Du2009}. For this purpose, the query is to be formulated in an appropriate and human-readable representation~\cite{Luo2019}. A common way to achieve this requirement involves formulating premises of sharp or possibilistic classification rules by defining conditions on the value ranges of features~\cite{Rashidi2011}. An example of such a region query is depicted in Fig.~\ref{fig:query_types}(b). Although a region query provides class information about many instances, this type of query differs from batch mode AL, where each instance of a selected batch is annotated individually~\cite{Hoi2006}.

\textbf{Comparison queries} enhance the learning process by obtaining relative information between instances~\cite{Kane2017}. 
For example, the comparison query, illustrated in Fig.~\ref{fig:query_types}(c), compares two instances~$\mathbf{x}_n$ and $\mathbf{x}_m$ by requesting whether they belong to the same class or not~\cite{Fu2011,Fu2014}. Regarding the ordinal grid classification example, another conceivable comparison query may ask which of the two grid instances $\mathbf{x}_n$ and $\mathbf{x}_m$ has a superior hosting capacity.

Going beyond these three query types, we will present our own proposals for query types as future research directions in Section~\ref{sec:challenges}.

\subsection{Annotation Types}
\label{subsec:annotation_types}
Usually, the type of an annotation depends on the query itself.
In this survey, we differentiate between the following three annotation types:

\textbf{Distinct annotations} are the simplest form of annotations. They represent categorical information without the scope of interpretation. Most AL strategies use them to encode class labels.
Other AL strategies expect a simple \texttt{yes} or \texttt{no} as a distinct annotation~\cite{Fu2011,Kane2017}. Furthermore, they can encode a sorting of instances in case of a comparison query~\cite{Qian2015}.
	
\textbf{Soft annotations} allow for the representation of continuous information. They are often inaccurate and subjective. Many AL strategies use them to obtain information on the confidence of a provided class label by requesting a numerical value in a continuous confidence interval~\cite{Song2018,Ni2012}. Another example is the use of probabilistic labels as gradual annotations~\cite{Calma2018a,Sandrock2019}, which enhanced the classification performance for certain tasks, e.g., in the medical domain~\cite{Nguyen2014a}.
	
\textbf{Explanatory annotations} are the most informative type of annotations. Instead of only communicating a distinct or soft decision, an explanatory annotation also explains why a certain decision has been made. An exemplary explanation would be: ``The instance $\mathbf{x}_n$ does not belong to the positive class because its feature value $x_{nd}$ is too low.''~\cite{Biswas2013}.

\subsection{Literature Overview}
A query mostly requests information of a specific kind.
Accordingly, the query and annotation types are closely coupled.
Table~\ref{tab:query_annotation_types} gives a literature overview of existing combinations of queries and annotations as interaction schemes. The query ``To which class does instance~$\mathbf{x}_n$ belong?'' known already from the traditional AL setting is excluded.
A more in-depth analysis of the \acronym strategies in Table~\ref{tab:query_annotation_types} with a focus on their query utility measures is provided in the appendices of this survey.

\begin{table*}
    \footnotesize
    \caption{Literature overview of combinations of queries and annotations employed by \acronym strategies.}
    \label{tab:query_annotation_types}
    \begin{tabular}{|P{0.23\textwidth}|P{0.325\textwidth}|P{0.37\textwidth}|}
    	\toprule
    	\multicolumn{1}{|c|}{\textbf{Strategy}} & \multicolumn{1}{c}{\textbf{Query}}  & \multicolumn{1}{|c|}{\textbf{Annotation}} \\
    	\midrule
    	\hline
    	\rowcolor{Gray}
        \multicolumn{3}{|c|}{{Instance Queries}} \\ \hline
    	\citet{Hu2019} & Does instance $\mathbf{x}_n\in\mathcal{X}$ belong to concept~$\mathcal{K} \subset \Omega_{Y}$? & distinct annotations: $\Omega_Z = \{\texttt{yes}, \texttt{no}\}$ \\
    	\hline
    	\multirowcell{2}[0pt][l]{\citet{Bhattacharya2019}} & To which class in $\left\{y^{(1)}, \dots, y^{(n)}\right\} \subset \Omega_{Y}$ does instance ${\mathbf{x}_n \in \mathcal{X}}$ belong? & \multirowcell{2}[0pt][l]{distinct annotations: $\Omega_Z = \Omega_{Y}$} \\
    	\hline
    	\citet{Cebron2012} & To which class does instance ${\mathbf{x}_n \in \mathcal{X}}$ not belong? & distinct annotations: $\Omega_Z = \mathcal{P}({\Omega_{Y}})$ \\
    	\hline
    	\citet{Donmez2008b,Donmez2010b}, \citet{Wallace2011,Fang2013e,Zhong2015,Kaeding2015} & \multirowcell{3.5}[0pt][l]{Provided that you are confident: What is the class \\ label of instance ${\mathbf{x}_n \in \mathcal{X}}$?} & \multirowcell{3.5}[0pt][l]{distinct annotations: ${\Omega_Z = \Omega_{Y} \cup \{\texttt{uncertain}\}}$}\\
    	\hline
    	\multirowcell{2}[0pt][l]{\citet{Donmez2008b,Donmez2010b}, \\\citet{Ni2012,Calma2018a}} & What is the class label of instance $\mathbf{x}_n \in \mathcal{X}$ and how confident are you? & soft annotations: ${\Omega_Z = \Omega_{Y} \times \Omega_{C}}$ where $\Omega_{C}$ denotes the set of possible confidence scores \\
    	\hline
    	\multirowcell{2}[0pt][l]{\citet{Song2018}} & How confident are you that instance $\mathbf{x}_n \in \mathcal{X}$ belongs to the positive class?   & soft annotations: ${\Omega_Z = [-1, 1]}$ with $z \in \Omega_z$ indicating the confidence that $\mathbf{x}_n$ belongs to the positive class  \\ 
    	\hline
    	\multirowcell{2}[0pt][l]{\citet{Biswas2013}} & Does instance ${\mathbf{x}_n \in \mathcal{X}}$ belong to class ${y \in \Omega_Y}$? If this is not the case, can you explain the reason? & explanatory annotations: $\Omega_Z = \{\texttt{yes}\} \cup \Omega_{E}$ with $\Omega_{E}$ representing the set of explanations \\
    	\hline
    	\multirowcell{2}[0pt][l]{\citet{Teso2019}} & Does instance ${\mathbf{x}_n \in \mathcal{X}}$ belong to class ${y \in \Omega_Y}$ because of explanation $e \in \Omega_E$? &  explanatory annotations: $\Omega_Z = \{\texttt{yes}\} \cup \Omega_Y \cup \Omega_{E}$ with $\Omega_{E}$ representing the set of explanations \\
    	\hline
    	\rowcolor{Gray}
    	\multicolumn{3}{|c|}{{Region Queries}} \\ \hline
    	\multirowcell{2}[0pt][l]{\citet{Druck2009,Settles2011a}} & For which classes is a positive feature value $X_d > 0$ highly indicative? & distinct annotations: ${\Omega_Z = \mathcal{P}({\Omega_{Y}}})$ with $z \subseteq \Omega_Y$ indicating the set of possible classes\\
    	\hline
    	\multirowcell{2.75}[0pt][l]{\citet{Du2009}} & What is the proportion of positive instances in the region described by the constellation of categorical features, e.g., ${X_1 \doteq 1 \wedge X_4 \doteq 0 \wedge X_8 \doteq 0}$? & \multirow{2}{=}[-0.65cm]{soft annotations: ${\Omega_Z = [0, 1]}$ with $z \in \Omega_Z$ indicating the proportion of positive instances}  \\
    	\cline{1-2}
    	\citet{Du2010,Hauskrecht2018,Luo2018,Luo2019}, \citet{Rashidi2011,Haque2013} & \multirowcell{2.75}[0pt][l]{What is the proportion of positive instances \\ in the region described by the feature constellation, \\e.g., ${X_1 \in [0, 2] \wedge X_2 \leq 10 \wedge X_3 \doteq 3}$?} & \\
    	\hline
    	\rowcolor{Gray}
    	\multicolumn{3}{|c|}{{Comparison Queries}} \\ \hline
    	\multirowcell{2}[0pt][l]{\citet{Fu2011,Fu2014}, \\ \citet{Joshi2010,Joshi2012}} & Do instance $\mathbf{x}_n \in \mathcal{X}$ and instance  $\mathbf{x}_m \in \mathcal{X}$ belong to the same class? & \multirowcell{2}[0pt][l]{distinct annotations: ${\Omega_Z = \{\texttt{yes}, \texttt{no}\}}$} \\
    	\hline
    	\multirowcell{2}[0pt][l]{\citet{Xiong2015}} & Is instance $\mathbf{x}_n \in \mathcal{X}$ more similar to instance $\mathbf{x}_m \in \mathcal{X}$ than instance $\mathbf{x}_o \in \mathcal{X}$? & \multirowcell{2}[0pt][l]{distinct annotations: $\Omega_{Z} = \{\texttt{yes}, \texttt{no}, \texttt{uncertain}\}$}\\
    	\hline
    	\multirowcell{2}[0pt][l]{\citet{Kane2017,Xu2017}, \\ \citet{Hopkins2020}} & Is instance $\mathbf{x}_n \in \mathcal{X}$ more likely to belong to the positive class than instance $\mathbf{x}_m \in \mathcal{X}$? & \multirowcell{2}[0pt][l]{distinct annotations: ${\Omega_Z = \{\texttt{yes}, \texttt{no}\}}$} \\
    	\hline
    	\multirowcell{2.75}[0pt][l]{\citet{Qian2015}} & What is the decreasing order of the instances $\{\mathbf{x}_n, \mathbf{x}_m, \mathbf{x}_o\} \subset \mathcal{X}$ regarding their similarities to instance $\mathbf{x}_p \in \mathcal{X}$? & \multirowcell{3}[0pt][l]{distinct annotations: $\Omega_Z$ consists of all possible ordering \\ of the available instances, e.g., ${(\mathbf{x}_m, \mathbf{x}_o, \mathbf{x}_n)} \in \Omega_{Z}$} \\
    	%\hline
    	%\multirowcell{2}[0pt][l]{\citet{Liang2020}} & How would you differentiate between the class $y \in \Omega_Y$ and class $y^\prime \in \Omega_Y$? & explanatory annotations: $\Omega_Z = \mathcal{E}$ with $\mathcal{E}$ as a set of natural-language explanations to differ between two classes \\
    	\hline
    	\bottomrule
    \end{tabular}
\end{table*}

\section{Annotator Performance Models}
\label{sec:annotator_models}
The error-proneness of annotators poses a major challenge in \acronym~\cite{Calma2016}.
In this section, we discuss the typical factors influencing the performance of error-prone annotators.
Moreover, we identify three different types of annotator performance.
At the end of this section, we present a literature overview of existing annotator performance models.

\subsection{Influence Factors}
\label{subsec:influence_factors}
We refer to ``annotator performance'' as a general term for the quality of the annotations obtained from an annotator.
There is no clear definition of this term, but there exist several concrete interpretations, e.g., label accuracy~\cite{Donmez2009}, confidence~\cite{Ni2012}, 
%abstention rate~\cite{Yan2016},
uncertainty~\cite{Fang2012}, reliability~\cite{Li2019}, etc.
Such an interpretation is closely coupled to the annotation type and the expected optimal annotation of a query.

The annotator performance may be affected by various factors~\cite{Daniel2018,Jin2020}, and the most prominent ones identified in the AL literature are given in the following:

The \textbf{domain knowledge} of annotators has an essential impact on their performances~\cite{Kazai2013}. Insufficient knowledge leads to a deterioration of the annotator performance. In complex tasks, such as assessing the hosting capacity of a low-voltage grid, a certain level of domain knowledge is indispensable.

The \textbf{query difficulty} affects the probability of obtaining an optimal annotation~\cite{Zhao2014,Whitehill2009,Beigman2010}. For example, in recognition of hand-written digits, it is often more challenging to differentiate between the digits 1 and 7 than discriminating between the digits 1 and 8~\cite{Calma2018a}. Next to the subject of a query, also its type can be crucial for the performance of an annotator~\cite{Joshi2012}.

The \textbf{ability for a reliable self-assessment} of annotators plays a central role, particularly in scenarios where queries ask for confidence scores as annotations~\cite{Ni2012}. Although empirical studies~\cite{Wallace2011,Calma2018} have shown that annotators can reliably estimate their performances in some domains, the Dunning-Kruger-effect~\cite{Kruger1999} states that, in particular, unskilled annotators provide not only erroneous annotations, but they also cannot realize their mistakes. This effect has also been confirmed in a large-scale crowd-sourcing study~\cite{Gadiraju2017}.

\textbf{Motivation or level of interest} of an annotator may influence the elaborateness during the annotation process. For example, in a crowdsourcing study analyzed in~\cite{Kazai2013}, more interested annotators performed superiorly.

The \textbf{payment} of an annotator may have a significant impact on the annotator performance, such that well-paid annotators provide more high-quality annotations. In a crowdsourcing environment, the improvement of the annotation quality has been confirmed by increasing the pay from 0.10\$ to 0.25\$ per query~\cite{Kazai2013}.

The annotator has to be \textbf{concentrated} when annotating a query~\cite{Calma2017}. Otherwise, annotation mistakes arise because of missing mindfulness or tiredness.

A constant stream of queries of the same type may be annoying for the annotator~\cite{Amershi2014}. Therefore, the \textbf{way of interaction} between the AL strategy and an annotator may influence the annotation results and needs to be designed appropriately. For example, different interaction schemes can lead to different degrees of an annotator's enjoyability, as experimentally shown in~\cite{Cakmak2010}. 

The \textbf{learning aptitudes} of annotators are also crucial for their performances. For example, one could teach the annotators to provide high-quality annotations~\cite{Daniel2018}.

The \textbf{collaboration} between annotators is also interlinked with their performances. Incorporating corresponding mechanisms for collaboration can strongly improve the annotation quality~\cite{Chang2017}.

\subsection{Annotator Performance Types}
\label{subsec:annotator_performance}
\begin{figure}[h!]
    \captionsetup[subfigure]{labelformat=empty}
	\centering
	\subfloat[Uniform]{
        \centering
            \resizebox{.29\columnwidth}{!}{
                \begin{tikzpicture}
                    	% Nodes
                    	\node[obs](Q){$Q$};
                    	\node[latent, above = of Q](Z){$Z$};
                    	\node[latent, right = of Z](P_m){$P_m$};
                    	\node[obs, right = of Q](Z_m){$Z_m$};
                    	\node[const, right = of P_m, xshift=-0.5cm](t){$t$};
                    	% Edges
                    	\draw[->, thick] (Z) -- (Z_m);
                    	\draw[->, thick] (P_m) -- (Z_m);
                    	\draw[->, thick, dashed] (t) -- (P_m);
                    	\draw[->, thick] (Q) -- (Z);
                \end{tikzpicture}
            }
		}
		\label{fig:uniform_-performance}
	\subfloat[Annotation-dependent]{
        \centering
            \resizebox{.29\columnwidth}{!}{
                \begin{tikzpicture}
                	% Nodes
                	\node[obs](Q){$Q$};
                	\node[latent, above = of Q](Z){$Z$};
                	\node[latent, right = of Z](P_m){$P_m$};
                	\node[obs, right = of Q](Z_m){$Z_m$};
                	\node[const, right = of P_m, xshift=-0.5cm](t){$t$};
                	% Edges
                	\draw[->, thick] (Z) -- (Z_m);
                	\draw[->, thick] (Z) -- (P_m);
                	%\draw[->, thick] (Q) -- (Z_m);
                	\draw[->, thick] (P_m) -- (Z_m);
                	\draw[->, thick, dashed] (t) -- (P_m);
                	\draw[->, thick] (Q) -- (Z);
                \end{tikzpicture}
            }
		}
		\label{fig:annotation-dependent-performance}
	\subfloat[Query-dependent]{
        \centering
            \resizebox{.29\columnwidth}{!}{
                \begin{tikzpicture}
                	% Nodes
                	\node[obs](Q){$Q$};
                	\node[latent, above = of Q](Z){$Z$};
                	\node[latent, right = of Z](P_m){$P_m$};
                	\node[obs, right = of Q](Z_m){$Z_m$};
                	\node[const, right = of P_m, xshift=-0.5cm](t){$t$};
                	% Edges
                	\draw[->, thick] (Z) -- (Z_m);
                	\draw[->, thick] (Z) -- (P_m);
                	\draw[->, thick] (Q) -- (P_m);
                	\draw[->, thick] (P_m) -- (Z_m);
                	\draw[->, thick, dashed] (t) -- (P_m);
                	\draw[->, thick] (Q) -- (Z);
                	\draw[->, thick] (Q) -- (Z_m);
                \end{tikzpicture}
            }
		}
		\label{fig:query-dependent-performance}
	\caption{Statistical models of annotator performance types: Following the idea of~\citet{Frenay2014}, we present three different annotator performance types as graphical models. There are four random variables depicted as nodes: $Q$ is the query, $Z$ is the optimal annotation, $Z_m$ is the annotation provided by annotator $a_m$, and $P_m$ is the variable indicating the performance of the annotator $a_m$. In the simplest case, $P_m$ is a binary variable to represent whether an annotator provides the optimal annotation ($P_m=1$) or not ($P_m=0$). We denote observed variables by shading the corresponding nodes, whereas the other nodes represent latent variables. The variable $t$ is a deterministic parameter denoting the time. Arrows represent statistical dependencies, e.g., the optimal annotation always depends on the underlying query. The dashed arrow between the annotator performance variable $P_m$ and the time $t$ indicates an optional dependency. If this dependency is considered, the annotator performance is time-varying~\cite{Donmez2010}. Otherwise, it is assumed to be persistent.}
	\label{fig:annotation-performance-types}
\end{figure}
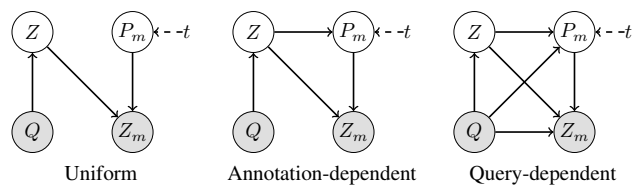

\begin{figure}[!h]
    \centering
    \includegraphics[width=\columnwidth]{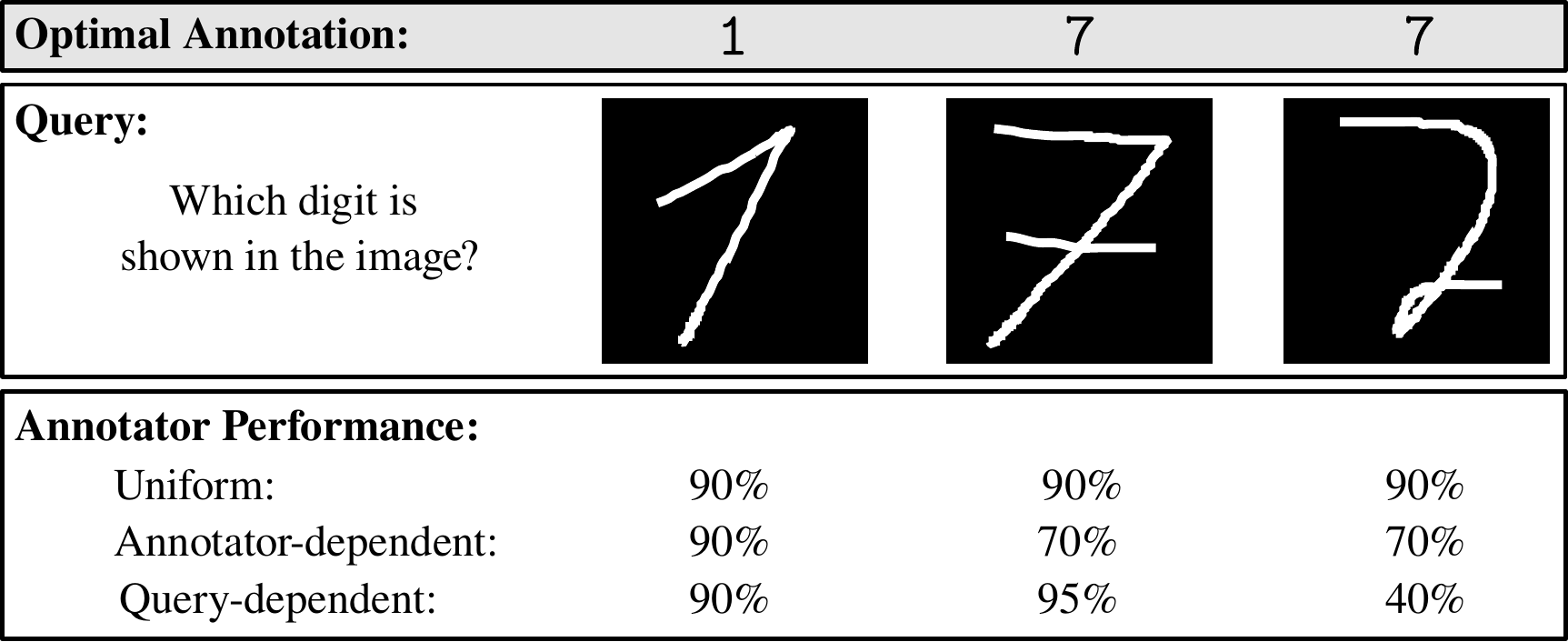}
    \caption{Illustration of annotator performance types: There are three images of hand-written digits. Assuming a uniform annotator performance, an annotator has an equal chance of correct digit recognition for each of the three images. In the case of annotation-dependent performance values, the chance of recognizing a digit correctly depends on its true class as optimal annotation, e.g., the annotator is better at recognizing digit~\texttt{1} than digit~\texttt{7}. The assumption of query-dependent performance values is more general and realistic. For example, the annotator has a low chance of recognizing the right digit due to its unclear writing. In the case of time-varying annotator performances, the chance of correct digit recognition can change over time, e.g., the chance may increase due to the learning progress of an annotator~\cite{Fang2012,Settles2008a} or may decrease because of exhaustion or emerging boredom~\cite{Cakmak2010}.}
    \label{fig:digit-example}
\end{figure}

Modeling and quantifying the influence of each of the previously listed factors on annotator performance is infeasible.
Instead, existing annotator models abstract from these factors to estimate annotator performance.
In the literature, we identified three different types of annotator performances. 
Therefor, we generalize the class label noise taxonomy, presented by~\citet{Frenay2014}, to the setting of \acronym by including queries and annotations instead of instances and classes. 
The resulting statistical taxonomy of annotator performance types is presented in Fig.~\ref{fig:annotation-performance-types}, and we provide more details in the following:

\textbf{Uniform annotator performance:} The annotator performance depends only on the characteristics of the annotator. As a result, the query itself or the query's optimal annotation has no influence. 
An example is given in Fig.~\ref{fig:digit-example}, where an annotator has the constant probability of $90\%$ to recognize a hand-written digit correctly. 

\textbf{Annotation-dependent annotator performance:} The annotator performance depends next to the annotator's characteristics on the optimal annotation for a query. An example is given in Fig.~\ref{fig:digit-example}, where an annotator is better at identifying the digit~1 (constant correctness probability of $90$\%) than the digit~7 (constant correctness probability of $70$\%) in images of hand-written digits. 

\textbf{Query-dependent annotator performance:} The annotator performance depends on the annotator's characteristics, the query, and the optimal annotation. An example is given in Fig.~\ref{fig:digit-example}, where an annotator has a low probability to correctly identify the third digit as 7 because it can be misinterpreted as the digit~2.
	
As an additional dimension, possible temporal dependencies regarding annotator performance can be taken into account.
Therefore, we differ between \textbf{persistent} and \textbf{time-varying} annotator performance.
In the first case, the annotator performance is constant during the entire annotation process. 
In the latter case, the annotator performance may increase due to the learning progress of an annotator~\cite{Fang2012,Settles2008a} or may decrease because of exhaustion or emerging boredom~\cite{Cakmak2010}.

\subsection{Literature Overview}
During the AL process, the performances of the annotators are estimated by annotator models.
Table~\ref{tab:annoator_models_part_1} provides a literature overview, including a categorization of those models. 
Next to the assumptions regarding the type of annotator performance, we use several other factors to categorize different annotator models.
In particular, the query and annotation types described in Section~\ref{sec:types_of_queries_and_annotations} are essential properties of an annotator model.
However, to the best of our knowledge, existing annotator models focus on instance queries such that no column for the query type is present in Table~\ref{tab:annoator_models_part_1}.
As a further category, we differentiate between the assumed relation of the annotators. 
In the case of multiple annotators, they are either independent or collaborative.
If a model can work with a single annotator, the term single is denoted for this category.
Furthermore, we indicate in Table~\ref{tab:annoator_models_part_1} whether an annotator model allows for the integration of prior knowledge regarding the performances of annotators.
Additionally, we provide a brief description of each annotator model's main idea.
A more in-depth analysis of these annotator models is provided in the appendices of this survey.

\newcommand{\methodDescriptionColumnWidth}{0.775\textwidth}
\begin{table*}[!ph]
    \footnotesize
    \caption{Part I: Literature overview of annotator models employed by \acronym strategies.}
    	\label{tab:annoator_models_part_1}
    \begin{tabularx}{\textwidth}{|L{0.175\textwidth}|L{0.35\textwidth}|L{0.15\textwidth}|L{0.125\textwidth}|Y|}
    	\toprule 
    	\multicolumn{1}{|c|}{\textbf{Strategy}} & \multicolumn{1}{C{0.35\textwidth}|}{\textbf{Annotation Type}} &  \multicolumn{1}{C{0.15\textwidth}|}{\textbf{Temporal Annotator Performance}} & \multicolumn{1}{C{0.125\textwidth}}{\textbf{Annotator Relation}} & \multicolumn{1}{|C{0.075\textwidth}|}{\textbf{Prior Knowledge}} \\ 
    	\midrule
    	\hline

    	\rowcolor{Gray}
    	\multicolumn{5}{|c|}{{Uniform Annotator Performance}} \\ \hline
    	\multirowcell{4}[0pt][l]{\citet{Donmez2009}, \\ \citet{Zheng2010}}  & distinct: class labels & persistent & independent & yes \\ \cline{2-5} 
    	& \multicolumn{4}{p{\methodDescriptionColumnWidth}|}{This annotator model estimates the true class label of an instance by means of majority voting. Following the interval estimation method~\cite{Kaelbling1993}, the majority votes are then used to evaluate the upper bound of the fraction of correctly annotated instances as performance estimate for each annotator.}\\
    	\hline
    	\multirowcell{4}[0pt][l]{\citet{Donmez2010}}  & distinct: class labels & time-varying & independent & no\\ \cline{2-5} 
    	& \multicolumn{4}{p{\methodDescriptionColumnWidth}|}{This annotator model models the quality of each annotator as a time-varying latent state sequence. For this purpose, it assumes that the change in the annotation quality from one to the next state follows a Gaussian distribution with a zero-mean and a known variance, which is shared among all annotators.}\\
    	\hline
    	\multirowcell{4}[0pt][l]{\citet{Long2013,Long2016}, \\ \citet{Long2015}}  & distinct: binary class labels & persistent & independent & no\\ \cline{2-5} 
    	& \multicolumn{4}{p{\methodDescriptionColumnWidth}|}{This annotator model, based on a probabilistic model with Gaussian processes~\cite{Rasmussen2003}, estimates a single performance value per annotator by comparing the provided annotations to the estimated true annotations. The performance value of an annotator indicates the probability that this annotator assigns the correct class label to an instance.}\\
    	\hline

    	\rowcolor{Gray}
    	\multicolumn{5}{|c|}{{Annotation-dependent Annotator Performance}} \\ \hline
    	\multirowcell{4.5}[0pt][l]{\citet{Wu2013b}}  & distinct: binary class labels & persistent & independent & yes \\ \cline{2-5} 
    	& \multicolumn{4}{p{\methodDescriptionColumnWidth}|}{This annotator model, based on the logistic regression model proposed by~\citet{Raykar2010a}, estimates the performance of an annotator in dependence of an instance's (unknown) true class label. Using the maximum a posteriori criterion, the model's training follows the expectation-maximization algorithm~\cite{Moon1996} which iteratively estimates the true class labels (expectation-step) to evaluate the probability of a correct annotation for each class-annotator pair (maximization-step).}\\
    	\hline 
    	\multirowcell{4}[0pt][l]{\citet{Rodrigues2014}}  & distinct: binary class labels & persistent & independent & no\\ \cline{2-5} 
    	& \multicolumn{4}{p{\methodDescriptionColumnWidth}|}{This annotator model, based on a Gaussian processes~\cite{Rasmussen2003} framework and expectation propagation~\cite{Minka2001}, estimates the class-dependent specificity and sensitivity of each each annotator by comparing the provided annotations to the estimated true annotations.}\\
    	\hline
    	\multirowcell{4}[0pt][l]{\citet{Moon2014}}  & distinct: class labels & persistent & independent & no\\ \cline{2-5} 
    	& \multicolumn{4}{p{\methodDescriptionColumnWidth}|}{This annotator model expects an initial set of instances annotated by each annotator. The true class labels of these instances are estimated through majority voting. Subsequently, the performance of an annotator is computed as the annotation accuracy, i.e., estimated fraction of correct annotations, per class.}\\
    	\hline 
    	\multirowcell{4}[0pt][l]{\citet{Nguyen2015}}  & distinct: class labels & persistent & independent & yes\\ \cline{2-5} 
    	& \multicolumn{4}{p{\methodDescriptionColumnWidth}|}{This annotator model differs between infallible experts and error-prone crowd workers. The performance of the latter ones is estimated by comparing their provided class labels with the expert class labels. For this purpose, the model computes a confusion matrix including a Bayesian prior for the group of crowd workers.}\\
    	\hline 
    
    	\rowcolor{Gray}
    	\multicolumn{5}{|c|}{{Query-dependent Annotator Performance}} \\ 
    	\hline
    	\multirowcell{3}[0pt][l]{\citet{Wallace2011}}  & distinct: binary class labels and \texttt{uncertain} & persistent &  independent & yes\\ \cline{2-5}
    	& \multicolumn{4}{p{\methodDescriptionColumnWidth}|}{This annotator model relies on domain information in form of annotators' pay grades. Therefore, it assumes that the pay grades of the annotator are highly correlated with their annotator performances.}\\
    	\hline
    	\multirowcell{5}[0pt][l]{\citeauthor{Donmez2008b} \\ \cite{Donmez2008b,Donmez2010b}} & soft: class labels and confidence scores & persistent & independent & no\\ \cline{2-5}
    	& \multicolumn{4}{p{\methodDescriptionColumnWidth}|}{This annotator model uses the annotators' confidence scores as proxies of their annotator performances. Using $k$-means clustering~\cite{Xu2005}, an annotator is queried to annotate the $k \in \mathbb{N}$ instances closest to the respective $k$~cluster centroids. It is assumed that instances belonging to a cluster, whose centroid has a high-confidence annotation, will be accurately annotated by the corresponding annotator.}\\
    	\hline
    	\multirowcell{4}[0pt][l]{\citet{Du2010}}  & distinct: binary class labels & persistent & single & no\\ \cline{2-5} 
    	& \multicolumn{4}{p{\methodDescriptionColumnWidth}|}{Since the classification model is trained under a single annotator's supervision, this annotator model assumes that the classification model behaves similarly to the annotator. As a result, the annotator performance estimates near the classification model's decision boundary are lower than in regions where the classification model is certain.}\\
    	\hline 
    	\multirowcell{5}[0pt][l]{\citet{Yan2012b,Yan2011}}  & distinct: binary class labels & persistent &  independent & no\\* \cline{2-5}
    	& \multicolumn{4}{p{\methodDescriptionColumnWidth}|}{This annotator model, based on a logistic regression model proposed in~\cite{Yan2010}, estimates the performance of an annotator in dependence of an instance and its true class label. Using the maximum likelihood criterion, the model's training follows the expectation-maximization algorithm~\cite{Moon1996} which iteratively estimates the true class labels (expectation-step) and uses them to evaluate the annotator performance for each instance-annotator pair (maximization-step).}\\*
    	\hline
    	\multirowcell{5}[0pt][l]{\citet{Ni2012}}  & soft: binary class labels and confidence scores & persistent &  single/independent & no\\ \cline{2-5}
    	& \multicolumn{4}{p{\methodDescriptionColumnWidth}|}{This annotator model uses the confidence scores provided by the annotators as proxies of their annotator performances. For non-annotated instances, these scores are estimated using the (inverse-)distance-weighted $k$-nearest-neighbor rule~\cite{Dudani1976}. Accordingly, an annotator's performance for an instance is defined as the weighted mean confidence score of its $k$ nearest neighbors being already annotated by this annotator.}\\
    	\hline
    	\multicolumn{5}{|r|}{{\textsc{Continued on the next page.}}} \\ 
    	\hline
    	\bottomrule
    \end{tabularx}
\end{table*}

\begin{table*}[!t]
    \footnotesize
    \ContinuedFloat
    \caption{Part II: Literature overview of annotator models employed by \acronym strategies.}
    \label{tab:annoator_models_part_2}
     \begin{tabularx}{\textwidth}{|L{0.175\textwidth}|L{0.35\textwidth}|L{0.15\textwidth}|L{0.125\textwidth}|Y|}
    	\toprule 
    	\multicolumn{1}{|c|}{\textbf{Strategy}} & \multicolumn{1}{C{0.35\textwidth}|}{\textbf{Annotation Type}} &  \multicolumn{1}{C{0.15\textwidth}|}{\textbf{Temporal Annotator Performance}} & \multicolumn{1}{C{0.125\textwidth}}{\textbf{Annotator Relation}} & \multicolumn{1}{|C{0.075\textwidth}|}{\textbf{Prior Knowledge}} \\ 
    	\midrule
    	\hline
    	\rowcolor{Gray}
    	\multicolumn{5}{|c|}{{Query-dependent Annotator Performance}} \\ 
    	\hline
    	\multirowcell{6}[0pt][l]{\citet{Fang2012}}  & distinct: binary class labels & time-varying & collaborative & no\\ \cline{2-5} 
    	& \multicolumn{4}{p{\methodDescriptionColumnWidth}|}{This annotator model interprets the performance as uncertainty of an annotator regarding high-level concepts, e.g., sports, politics, and culture in case of document classification. These concepts are latent variables and modeled through a Gaussian mixture model~\cite{Gershman2012}. An instance may belong to multiple concepts. Using the maximum likelihood criterion, the model's training follows the expectation-maximization algorithm~\cite{Moon1996}, which iteratively estimates the true class labels (expectation-step) and takes them as basis for evaluating an annotator's uncertainty in annotating an instance (maximization-step).}\\
    	\hline
    	\multirowcell{4}[0pt][l]{\citet{Fang2013e}}  & distinct: binary class labels and \texttt{uncertain} & persistent & single/independent & no \\ \cline{2-5} 
    	& \multicolumn{4}{p{\methodDescriptionColumnWidth}|}{This annotator model expects the annotator to provide $\texttt{uncertain}$ as annotation, if the annotator does not know an instance's true class label. Using this information, the model characterizes the performance of the annotator by training a classifier to estimate the probability whether an instance will not belong to the annotator's uncertain knowledge set.}\\
    	\hline
    	\multirowcell{5}[0pt][l]{\citet{Fang2013,Fang2014}}  & distinct: binary class labels & persistent & independent & no \\ \cline{2-5} 
    	& \multicolumn{4}{p{\methodDescriptionColumnWidth}|}{This annotator model assumes that the performance of an annotator depends on a high-level representation of an instance's features and the instance's true class label. This dependency is indirectly modeled by introducing a latent variable for the expertise of each annotator. The expertise of an annotator is then computed as weighted linear combination of the instance's high level features.}\\
    	\hline 
    	\multirowcell{5}[0pt][l]{\citet{Zhao2014}}  & distinct: binary class labels & persistent & independent & no \\ \cline{2-5} 
    	& \multicolumn{4}{p{\methodDescriptionColumnWidth}|}{This annotator model estimates the annotator performance through two latent variables, namely, the query difficulty and the query-independent expertise of an annotator. For example, for annotators with high expertise or for easy queries, the probability of providing the true class label is high. The query difficulty and annotator expertise are latent and therefore iteratively estimated through the expectation-maximization algorithm~\cite{Moon1996}.}\\
    	\hline
    	\multirowcell{5.5}[0pt][l]{\citet{Zhong2015}, \\ \citet{Kaeding2015}}  & distinct: class labels and \texttt{uncertain} & persistent & single/independent & no \\ \cline{2-5}
    	& \multicolumn{4}{p{\methodDescriptionColumnWidth}|}{These annotator model allow an annotator to provide  \texttt{uncertain} as annotation in case of a lack of knowledge regarding an instance's class membership, otherwise she/he provides a class label. The instances annotated  with class labels (positive class) and the ones annotated with \texttt{uncertain} (negative class) form a binary classification problem. They are used to train an annotator model, i.e., a support vector machine~\cite{Tong2002} in~\cite{Zhong2015} and Gaussian processes~\cite{Rasmussen2003} in~\cite{Kaeding2015}. It predicts whether an annotator has sufficient knowledge to annotate an instance (positive class) or not (negative class).}\\
    	\hline 
    	\multirowcell{4}[0pt][l]{\citet{Huang2017}}  & distinct: class labels & persistent & independent & no \\ \cline{2-5} 
    	& \multicolumn{4}{p{\methodDescriptionColumnWidth}|}{This annotator model expects an initial set of instances with true class labels and annotations of each annotator. The model assumes that an annotator has a similar performance on similar instances. Therefore, it estimates an annotator's performance for an instance by computing the annotation accuracy regarding the instance's nearest neighbors in the initial set.}\\
    	\hline
    	\multirowcell{4.5}[0pt][l]{\citet{Yang2018}}  & distinct: class labels & persistent & independent & no \\ \cline{2-5} 
    	& \multicolumn{4}{p{\methodDescriptionColumnWidth}|}{This annotator model learns a low-dimensional embedding for each annotator to capture the annotator's expertise regarding latent topics. Additionally, an embedding for each instance is learned as representation by the latent topics. Both embeddings are combined to estimate the performance of an annotator. Since these embeddings are latent variables, they are learned through the expectation-maximization algorithm~\cite{Moon1996}.}\\
    	\hline
    	\multirowcell{4.5}[0pt][l]{\citet{Chakraborty2020}}  & distinct: class labels & persistent & independent & no \\ \cline{2-5} 
    	& \multicolumn{4}{p{\methodDescriptionColumnWidth}|}{This annotator model expects an initial set of instances with true class labels and annotations of each annotator. Since the true class labels are known in this set, the mistakes of each annotator can be determined on this set. A binary logistic regression classifier is then trained for each annotator separately. The trained logistic regression model of an annotator estimates her/his performance as the probability of obtaining a correct annotation for a certain instance.}\\
    	\hline
    	\multirowcell{4.5}[0pt][l]{\citet{Herde2021}}  & distinct: class labels & persistent & independent & yes\\ \cline{2-5} 
    	& \multicolumn{4}{p{\methodDescriptionColumnWidth}|}{This annotator model estimates the performance of an annotator for a certain instance in form of a Beta distribution. This distribution is parameterized by the number of estimated false and true annotations in the local neighborhood of an instance. A false or true annotation of an annotator is identified by comparing the annotations of a single annotator to the predictions of a classifier trained with the annotations of the other annotators.}\\
    	\hline
        \bottomrule
    \end{tabularx}
\end{table*}

\section{Selection Algorithms}
\label{sec:selection_strategies}
The selection of query-annotator pairs is based on a selection algorithm.
It uses the query utility measure $\phi$ and the annotator performance measure $\psi$ as basis to specify~${\mathcal{S}(t) \subseteq \mathcal{Q}_\mathcal{X} \times \mathcal{A}}$ as the set of query-annotator pairs in each AL iteration cycle~$t \in \mathbb{N}$.
In this context, we differentiate between two types of selection algorithms, explained in the following.
At the end of this section, we present a literature overview of existing selection algorithms.

\subsection{Sequential Selection of Queries and Annotators}
\label{subsec:sequential_selection}
Sequential selection of queries and annotators is made in two steps. 
In the first step, one or multiple (in the case of batch mode AL) queries with the highest utilities are selected.
In a second step, corresponding annotators are selected and assigned to the respective queries, e.g., a predefined number of the annotators with the highest estimated performances per query~\cite{Long2013}.
Ideally, the selected annotators lead to low AC while providing high accuracy annotations.
The main motivation for a sequential selection is to emphasize useful queries by selecting them in advance of the annotators.
Moreover, the issue of annotator selection reduces to determining a ranking of the annotators regarding a selected query.
As a result, not the exact but only the relative differences between the performances of the annotators are crucial for the annotator selection.

\subsection{Joint Selection of Queries and Annotators}
\label{subsec:joint_selection}
Selecting queries without considering the annotator's performances can result in low-quality annotations because there is no guarantee that at least one annotator has a sufficient performance regarding a selected query~\cite{Zhong2015}.
This problem can be resolved by applying a selection algorithm jointly selecting queries and annotators.
For this purpose, the query utility and the annotator performance measure are to be combined appropriately, e.g., by taking their product~\cite{Huang2017}.
Compared to the sequential selection of queries and annotators, the joint selection comes with higher computational complexity. Instead of computing the annotator performance estimates only for the selected queries, the annotator performance estimates are required for each possible query.
Moreover, exact estimates regarding the annotator performance are more crucial since the annotator performance estimates are directly integrated into the selection criterion.
If these estimates are unreliable, not only the annotator selection will be negatively affected but also the combination with the query selection.

\subsection{Literature Overview}
\label{subsec:selection-algorithms}
Table~\ref{tab:selection-algorithms} provides an overview of selection algorithms employed by existing \acronym strategies, which select query-annotator pairs.
Next to the differentiation between a sequential and joint selection of queries and annotators, the number of selected queries and annotators per learning cycle is of interest.
Selecting only a single query-annotator pair is often easier than selecting a batch of query-annotator pairs.
In the latter case, the selection algorithm must ensure that queries are diverse.
Otherwise, redundant information is queried.
Moreover, multiple annotators are to be distributed across queries.
To differentiate between both settings, we denote either single or batch for the query and annotator selection categories in Table~\ref{tab:selection-algorithms}. 
A few selection algorithms consider criteria beyond annotator performance and query utility, e.g., a collaboration between annotators.
We denote these criteria accordingly in Table~\ref{tab:selection-algorithms}.
Additionally, we provide a brief description of each selection algorithm's main idea.
A more in-depth analysis of them is provided in the appendices of this survey.

\begin{table*}[!ht] \centering
    \caption{Literature overview of selection algorithms employed by \acronym strategies.}
    \label{tab:selection-algorithms}
    \footnotesize
    \begin{tabularx}{\textwidth}{|L{0.23\textwidth}|p{0.18\textwidth}|p{0.18\textwidth}|p{0.31\textwidth}|}
        \toprule
        \multicolumn{1}{|c|}{\textbf{Strategy}} & \multicolumn{1}{c|}{\textbf{Query Selection}} & \multicolumn{1}{c}{\textbf{Annotator Selection}} & \multicolumn{1}{|c|}{\textbf{Criteria Beyond Utility and Performance}}\\
        \midrule
        \hline
        \rowcolor{Gray}
        \multicolumn{4}{|c|}{{Sequential Selection}} \\ 
        \hline
        \multirowcell{3}[0pt][l]{\citet{Ni2012,Wu2013b}, \\ \citet{Rodrigues2014}, \citeauthor{Fang2013} \\ \cite{Fang2013,Fang2014}, \citet{Zhong2015}} & single & single & none\\
        \cline{2-4}
        & \multicolumn{3}{X|}{These strategies select the query with the highest estimated utility. Subsequently, they select the annotators with the highest estimated performances regarding the annotation of this query.} \\
        \hline
        \multirowcell{3.6}[0pt][l]{\citet{Wallace2011}} & single & single & workload of annotators\\
        \cline{2-4}
        & \multicolumn{3}{X|}{This strategy selects either a non-annotated query with the highest estimated utility or a query for re-annotation. The annotator selection follows a categorical distribution whose parameters reflect a certain objective, e.g., balancing the annotation workload among annotators.} \\
        \hline
        \multirowcell{3.6}[0pt][l]{\citet{Zhao2014}} & single & single & none\\
        \cline{2-4}
        & \multicolumn{3}{X|}{This strategy selects the query with the highest estimated utility.  The annotator selection follows one of two options. On the one hand, an annotator can be selected with a probability proportional to her/his estimated performance. One the other hand, the estimated best annotator is either selected with a pre-defined probability or a random one.} \\
        \hline
        \multirowcell{3}[0pt][l]{\citet{Donmez2009}} & single & batch & none\\
        \cline{2-4}
        & \multicolumn{3}{X|}{This strategy selects the query with the highest estimated utility. Subsequently, it selects an adaptive number of annotators with the highest estimated performances.} \\
        \hline
        \multirowcell{3.6}[0pt][l]{\citet{Zheng2010}} & single & batch & none\\
        \cline{2-4}
        & \multicolumn{3}{X|}{This strategy selects the query with the highest estimated utility. In an exploration phase, it assigns an adaptive number of annotators with the highest estimated performances to this query. In the subsequent exploitation phase, a fixed subset of annotators with low ACs and high performances is determined and always selected.} \\
        \hline
        \multirowcell{3.6}[0pt][l]{\citet{Fang2012}} & single & batch & collaboration between annotators\\
        \cline{2-4}
        & \multicolumn{3}{X|}{This strategy selects the query with the highest estimated utility. Subsequently, it selects not only the annotator with the highest estimated performance but additionally the annotator with the lowest estimated performance. This way, the estimated best annotator can teach the estimated worst annotator.} \\
        \hline
        \multirowcell{3}[0pt][l]{\citet{Long2013,Long2016}, \\ \citet{Long2015}} & single & batch & none\\
        \cline{2-4}
        & \multicolumn{3}{X|}{These strategies select the query with the highest estimated utility. Subsequently, they select a pre-defined number of annotators with the highest estimated performances.} \\
        \hline
        \multirowcell{3}[0pt][l]{\citet{Yang2018}} & batch & batch & none\\
        \cline{2-4}
        & \multicolumn{3}{X|}{This strategy selects a pre-defined number of queries with the highest estimated utilities. Subsequently, it assigns to each of these selected queries the respective annotator with the highest estimated performance.} \\
        \hline
        \rowcolor{Gray}
        \multicolumn{4}{|c|}{{Joint Selection}} \\ 
        \hline
        \multirowcell{3.3}[0pt][l]{\citet{Donmez2008b,Donmez2010b}, \\ \citet{Moon2014}, \\ \citet{Huang2017}} & single & single & none\\
        \cline{2-4}
        & \multicolumn{3}{X|}{\vspace{-0.25cm}These strategies select the query-annotator pair whose product of estimated query utility and annotator performance is the highest. \vspace{0.1cm}} \\
        \hline
        \multirowcell{3}[0pt][l]{\citet{Yan2011}} & single & single & none\\
        \cline{2-4}
        & \multicolumn{3}{X|}{This strategy jointly selects a query and annotator by solving a linearly constrained and bi-convex optimization problem. Its goal is to find the optimal trade-off between a highly useful query and a high-performance annotator.} \\
        \hline
        \multirowcell{3}[0pt][l]{\citet{Yan2012b}} & single & single & none\\
        \cline{2-4}
        & \multicolumn{3}{X|}{This strategy combines the query utility information and annotator performance information through a mutual information criterion~\cite{Cover1991} as the joint selection criterion for a query-annotator pair.} \\
        \hline
        \multirowcell{3}[0pt][l]{\citet{Nguyen2015}, \\ \citet{Herde2021}} & single & single & none\\
        \cline{2-4}
        & \multicolumn{3}{X|}{These strategies jointly select a query and annotator by incorporating the estimated performance of an annotator (group) into the query utility measure quantifying the performance gain of the classification model.} \\
        \hline
        \multirowcell{3}[0pt][l]{\citet{Chakraborty2020}} & batch & batch & query diversity\\
        \cline{2-4}
        & \multicolumn{3}{X|}{This strategy jointly selects a batch of query-annotator pairs by solving a linear programming problem. Its solution balances the trade-off between useful queries, accurate annotators, and a small redundancy between these queries.} \\
        \hline
        \bottomrule
    \end{tabularx}
\end{table*}

\section{Future Research Directions}
\label{sec:challenges}
This section proposes some future research directions resulting from analyzing the \acronym strategies discussed in the previous sections.
We structure them into three categories to distinguish between challenges that strongly relate to this survey and those that go partially beyond it.
Although we define these addressable challenges separately, they are not entirely solvable without taking a holistic view.

\subsection{Active Learning for Classification}
\label{subsec:al-for-classification}
\textbf{Multi-criteria cost functions:} 
The majority of existing \acronym strategies minimize the number of queries and misclassifications.
However, in real-world applications, the ACs are often unknown in advance and may be query- and annotator-dependent.
Furthermore, the computation of the MC is related to the application at hand. 
Therefore, an AL strategy needs to accept a user-defined objective function as input.
This function needs to account for additional criteria, such as balancing the workload between annotators~\cite{Wallace2011}.

\textbf{Novel query types and a combination of them:}
Present AL strategies focus on collecting novel information relevant to the classification model. 
However, a query may not only improve the classification model but additionally the queried annotator~\cite{Ghai2020}.
For example, a strategy could ask ``Are you certain that instance $\mathbf{x}_n \in \mathcal{X}$ belongs to class $y \in \Omega_{Y}$? Previously, you stated that the similar instance $\mathbf{x}_m \in \mathcal{X}$ belongs to class $y^\prime \in \Omega_{Y}$?''. 
Such a query may help the annotator to learn from previous annotation mistakes. 
Moreover, most pool-based AL strategies query class information of instances.
However, recently, \citet{Liang2020} proposed the strategy \textit{active learning with contrastive natural language explanations} (ALICE).
It uses queries of the form ``How would you differentiate between the class $y \in \Omega_Y$ and class $y^\prime \in \Omega_Y$?'' in combination with explanatory annotations.
As a result, ALICE does not need a pool of non-annotated instances but only a small initial training set.
Next to novel query types, future strategies may combine different query types to enhance interaction with annotators further.

\textbf{Batch selection of diverse queries and annotators:}
Deep learning model's generalization capabilities depend on a vast amount of data.
Therefore, annotating single queries per AL cycle may be inappropriate~\cite{Sener2018}.
Instead, a batch of diverse and useful queries is to be selected per AL cycle.
Such a batch maximizes usefulness by avoiding redundancies.
In a multi-annotator setting, assigning appropriate annotators to these queries is an additional challenge.
For example, assigning all queries in a batch to a single annotator can be harmful because it could bias the performance estimates of the other annotators~\cite{Rodrigues2014}.

\textbf{Advanced annotator performance estimation:}
Existing annotator models are limited in their application due to their assumptions.
On the one hand, most of them assume persistent annotator performances and thus disregard, e.g., learning capabilities, collaboration, or signs of fatigue.
On the other hand, they do not incorporate background knowledge about the annotators, e.g., interests, skills, level of education, age, etc.
Such knowledge may improve the selection of annotators~\cite{Difallah2013}.

\textbf{Realistic Evaluation:}
Evaluating \acronym strategies is more complex than assessing traditional AL strategies.
In particular, the simulation of realistic experimental settings represents a challenge.
For example, there is a need to collect real-world data sets processed by multiple annotators to verify the performance of AL strategies in multi-annotator settings.
When collecting such data sets, it is infeasible to present each possible query to each annotator.
Therefore, a further research direction is the simulation of annotators for different query types.
Moreover, an AL strategy may be evaluated in a real-world system~\cite{Baldridge2009} in addition to simulated experiments on benchmark data sets to verify its effectiveness regarding real-world applications.

\subsection{Active Learning Issues Beyond Classification}
Although we focused on AL strategies for classification in this survey, 
their analysis provides insights beyond a classification setting.
If we exemplify object detection in images, similar challenges arise when employing AL strategies.
For example, relying on the number of annotated images as AC is not representative.
Instead, the number of objects within an image is more appropriate~\cite{Shen2020} because annotating images with many objects is more time-intensive.
Another example for object detection is the handling of error-prone annotators, where the AL strategy has additionally to assess the quality of provided bounding box annotations.

A challenge affecting pool-based AL with multiple annotators is the asynchronous nature of the annotation process~\cite{Chang2017}.
This results from different working speeds of annotators, i.e., some annotators process queries faster than others.
Due to this asynchronous nature, the selection of query-annotator pairs must be adaptive regarding the working states of the annotators.
This is, in particular, true for stream-based AL.

Techniques of explainable artificial intelligence may further improve the interaction between annotators and AL strategy.
For example, the ML model can visualize its decision-making process such that a human annotator can monitor the model's learning progress and correct wrong decisions~\cite{Ghai2020}.

\subsection{Active Learning Issues Beyond Artificial Intelligence}
Deploying AL strategies into real-world applications not only raises challenges in the scope of artificial intelligence but also involves research beyond it.
One example is graphical user interfaces of the annotation process, which are crucial for the efficiency of the AL process. 
Studies have shown that an appropriate user interface design strongly decreases the annotation time and thus AC~\cite{Settles2011a,Papadopoulos2017}.
Another example is the design of queries and annotations from a psychological perspective.
On the one hand, queries are to be formulated neutral without a bias toward a specific annotation.
On the other hand, annotations are to be comparable, particularly when asking for the annotators' self-assessments. 

Another future research direction is integrating AL into further little to no explored application areas to exploit its full potential.
For example, it can be employed in material science to actively design experiments in a more systematic way~\cite{Lookman2019} or for automatic program repair~\cite{Boehme2020} to save cost and time.
Another example would be the review process in science, where AL can select appropriate reviewers as annotators for articles.
Therefor, one could use feedback from authors of past conferences and the reviewers' background knowledge to train annotator models.

\section{Conclusion}
\label{sec:conclusion}
At the start of this survey, we pointed out unrealistic assumptions as disadvantages of traditional AL strategies.
Based on that, we identified three crucial requirements for \acronym strategies, i.e., estimating costs, asking alternative queries, and modeling annotator performances.
Subsequently, we formalized the objective for classification tasks as the specification of the optimal annotation sequence leading to minimum MC and AC.
Additionally, we proposed a novel AL cycle that generalizes the settings of the majority of existing \acronym strategies.
A strategy is part of a learning system in this cycle and comprises a query utility measure, an annotator performance measure, and a selection algorithm.
We provided tabular literature overviews of existing \acronym strategies regarding their cost types, their query- and annotation-based interaction, their handling of error-prone annotators, and their selection of query-annotator pairs.
In addition, we analyzed the \acronym strategies in more detail and embedded them in our unifying mathematical notation in the appendices.
These analyses resulted in the formulation of future research directions in the field of AL.

\bibliographystyle{IEEEtranN}
\bibliography{main}

\begin{IEEEbiography}
    [{\includegraphics[width=1in,height=1.25in,clip,keepaspectratio]{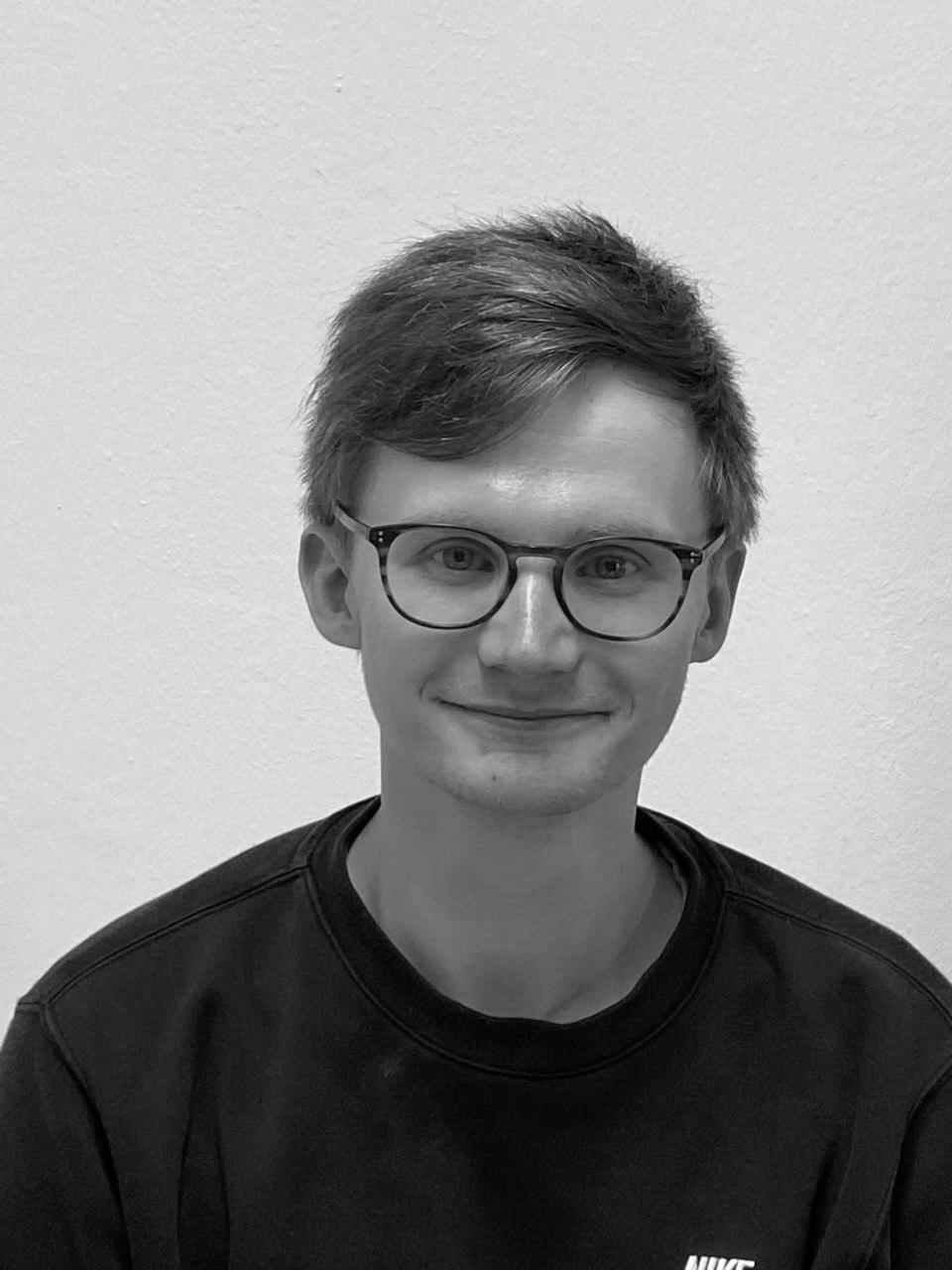}}]{Marek Herde} received his B.Sc. and M.Sc. degrees in computer science from the Univ.\ of Kassel, Germany. Currently, he is also pursuing his Ph.D. degree in computer science there. His research focuses on active learning, deep learning, and methods for learning from error-prone annotators.
\end{IEEEbiography}
\vspace*{-2\baselineskip} % \vskip -2\baselineskip plus -1fil
\begin{IEEEbiography}
    [{\includegraphics[width=1in,height=1.25in,clip,keepaspectratio]{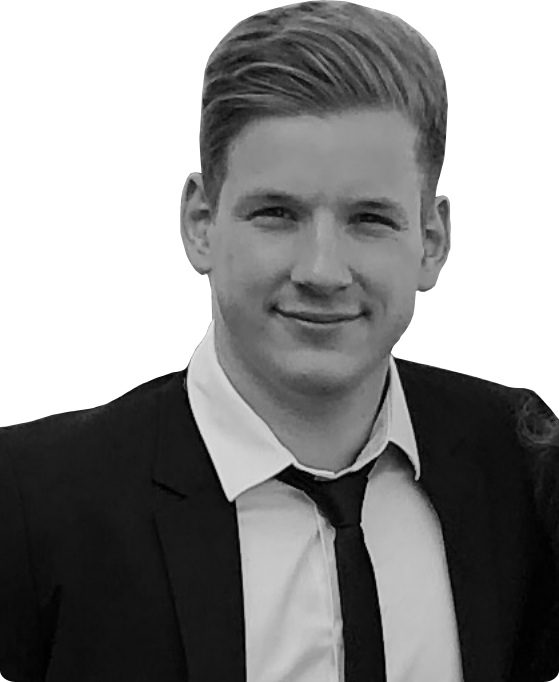}}]{Denis Huseljic} received his B.Sc. and M.Sc. degrees in computer science from the Univ.\ of Kassel, Germany. Currently, he is also pursuing his Ph.D. degree in computer science there. His research focuses on active learning, deep learning for computer vision, and methods for uncertainty estimation.
\end{IEEEbiography}
\vspace*{-2\baselineskip}%\vskip -2\baselineskip plus -1fil
\begin{IEEEbiography}
    [{\includegraphics[width=1in,height=1.25in,clip,keepaspectratio]{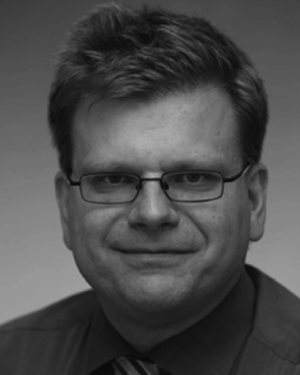}}]{Bernhard Sick}received his Diploma, Ph.D., and Habilitation degrees from the Univ.\ of Passau, Germany.
    %, in 1992, 1999, and 2004, respectively.
    He is currently a full professor for Intelligent Embedded Systems at the 
    %with the faculty for electrical engineering / computer science,
    Univ.\ of Kassel, Germany. His research comprises data science and machine learning with applications, e.g., in renewable energies, autonomous driving, physics/materials science. He authored more than 200 peer-reviewed publications in these areas. He 
    %holds one patent and 
    received several theses, best paper, teaching, and inventor awards. He is  a member of IEEE and GI.
    %the IEEE Systems, Man, and Cybernetics Society, Computer Society, and Computational Intelligence Society as well as GI. % (Gesellschaft f\"ur Informatik). 
    %He is an Associate Editor of the IEEE Transactions on Cybernetics.
\end{IEEEbiography}
\vspace*{-2\baselineskip}
\begin{IEEEbiography}
    [{\includegraphics[width=1in,height=1.25in,clip,keepaspectratio]{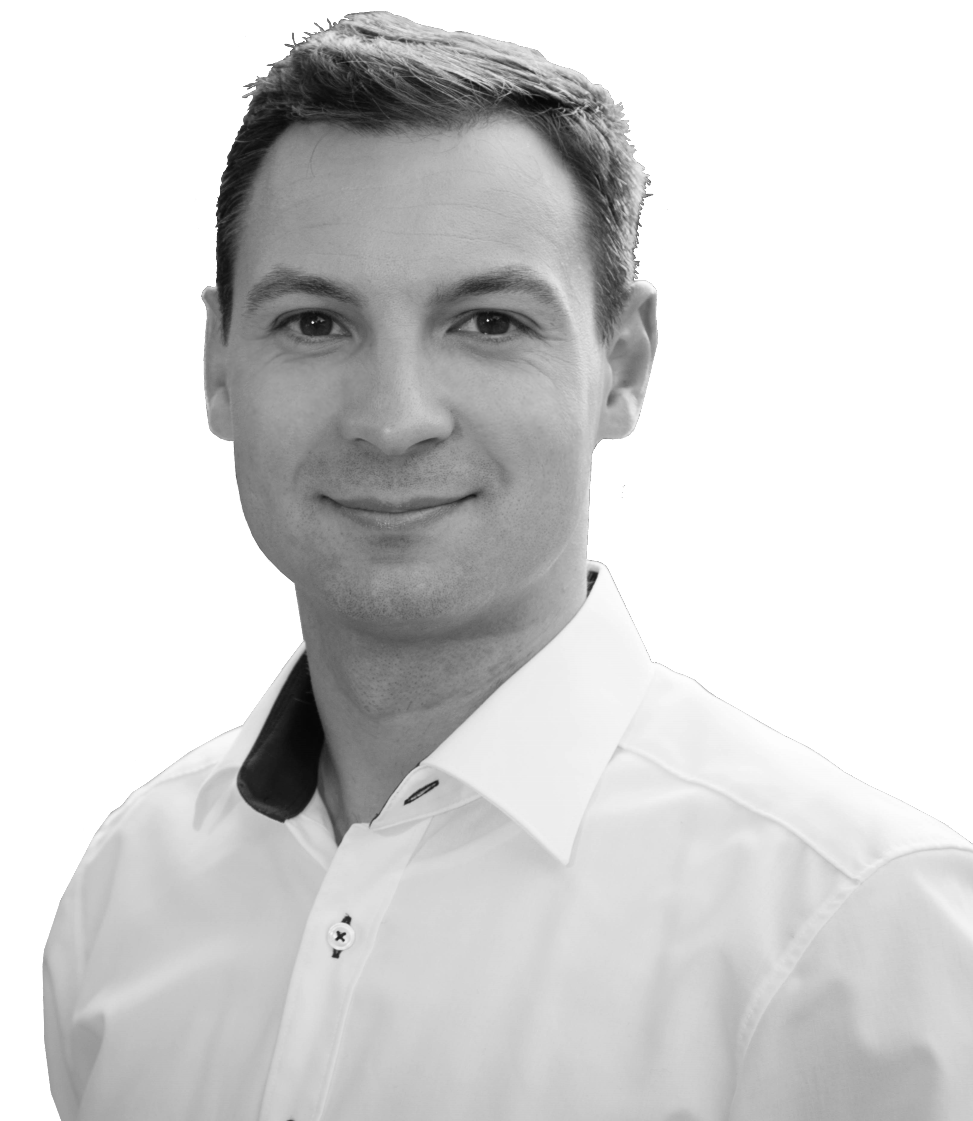}}]{Adrian Calma} received his B.Sc., M.Sc., and Ph.D. degrees in computer science from the Univ.\ of Kassel, Germany. He is keen on applying active learning techniques to real-world problems. His research focuses on developing active learning techniques handling error-prone annotators. Currently, he is a Fellow in Intelligent Embedded Systems Lab at the Univ.\ of Kassel, where he is working on improving precision farming methods with active learning.
\end{IEEEbiography}
\vfill

\title{Appendices of \\ A Survey on Cost Types, Interaction Schemes, and Annotator Performance Models in Selection Algorithms for Active Learning in Classification}

\author{
	Marek Herde$^{\textsuperscript{\orcidicon{0000-0003-4908-122X}}}$, Denis Huseljic$^{\textsuperscript{\orcidicon{0000-0001-6207-1494}}}$, Bernhard Sick$^{\textsuperscript{\orcidicon{0000-0001-9467-656X
	}}}$, \IEEEmembership{Member, IEEE}, Adrian Calma$^{\textsuperscript{\orcidicon{0000-0003-4380-7018
	}}}$
	\thanks{M. Herde, D. Huseljic, B. Sick, and A. Calma are with the department
		of Intelligent Embedded Systems, University of Kassel, Germany (e-mail: \{marek.herde $|$ dhuseljic $|$ bsick $|$ adrian.calma\}@uni-kassel.de).}% <-this % stops a space
	\thanks{This research was supported by the CIL project at the University of Kassel under internal funding P/710 and P/1082.}% <-this % stops a space
	\thanks{We thank Daniel Kottke, Tuan Pham Minh, Lukas Rauch, and Robert Monarch for their comments that greatly improved this survey.}
}

\maketitle

\setcounter{equation}{0}
\setcounter{figure}{0}
\setcounter{table}{0}
\setcounter{page}{1}

\appendices
\section*{General}
\label{appendix:general}
\IEEEPARstart{T}HE following appendices provide a more in-depth analysis of the \acronym strategies reviewed in the associated survey.
This analysis includes discussing the \acronym strategies regarding their cost types in Appendix~\ref{appendix:cost-types}, their interaction schemes in Appendix~\ref{appendix:interaction-schemes}, their annotator performance models in Appendix~\ref{appendix:annotator-performance-models}, and their selection algorithms in Appendix~\ref{appendix:selection-algorithms}.
Table~\ref{tab:abbreviation} lists essential abbreviations, and Table~\ref{tab:mathematical-notation} explains the mathematical notation used throughout this survey.
For ease of notation, we do not explicitly denote step~$t$ if it is not required.
Table~\ref{tab:strategies-overview} lists all analyzed \acronym strategies, including their acronyms where N/A denotes not available.
The crosses and check-marks indicate to which of the four research aspects these strategies contributed.
If more than one reference is given for a strategy, the year of publication refers to the most recent one.

\begin{table}[!h]
	\centering
	\caption{List of essential abbreviations.}
	\label{tab:abbreviation}
	\begin{tabular}{|R{0.345\columnwidth}|L{0.555\columnwidth}|}
		\toprule
		\textbf{Abbreviation} & \textbf{Meaning} \\
		\midrule
		\hline
		ML & machine learning \\
		AL & active learning \\
		AC & annotation cost \\
		MC & misclassification cost \\
		US & uncertainty sampling \\
		EER & expected error reduction \\
		EM & expectation-maximization \\
		EP & expectation propagation \\
		GMM & Gaussian mixture model \\
		SVM & support vector machine \\
		NN & nearest neighbors \\
		\hline 
		\bottomrule
	\end{tabular}
\end{table}

\begin{table}[!h]
	\ContinuedFloat
	\centering
	\caption{Part I: Overview of mathematical notation.}
	\label{tab:mathematical-notation}
	\begin{tabular}{|R{0.345\columnwidth}|L{0.555\columnwidth}|}
		\toprule
		\textbf{Symbol} & \textbf{Meaning} \\
		\midrule
		\hline
		\rowcolor{Gray}
		\multicolumn{2}{|c|}{Data Spaces} \\ 
		\hline
		$\Omega_X$ & feature/input space of possible instances \\
		$\Omega_Y$ & set of possible class labels \\
		$\Omega_Z$ & set of possible annotations \\
		$\Omega_C$ & set of possible confidence scores \\
		$\Omega_E$ & set of possible explanations \\
		$\Omega_S$ & set of possible annotation sequences \\
		\hline 
		\rowcolor{Gray}
		\multicolumn{2}{|c|}{Dimensions} \\ 
		\hline
		$N \in \mathbb{N}$ & number of instances \\
		$D \in \mathbb{N}$ & number of features \\
		$C \in \mathbb{N}$ & number of classes \\
		$M \in \mathbb{N}$ & number of annotators \\
		$L, O \in \mathbb{N}$ & context-sensitive number of dimensions \\
		\hline
		\rowcolor{Gray}
		\multicolumn{2}{|c|}{Data Entities} \\ 
		\hline
		$\mathcal{X} \define \{\mathbf{x}_1, \dots, \mathbf{x}_N\}$ & set of observed instances \\
		$\mathcal{Y} \define \{y_1, \dots, y_N\}$ & set of true class labels \\
		$\mathcal{A} \define \{a_1, \dots, a_M\}$ & set of error-prone annotators \\
		$\mathcal{Q}_\mathcal{X} \define \{q_1, q_2, \dots\}$ & set of all possible queries \\
		$\mathbf{x} = (x_1, \dots, x_D)^\mathrm{T} \in \Omega_X$ & feature vector  \\
		$\widetilde{\mathbf{x}} $  & (non-linear) transformation of $\mathbf{x}$ \\
		$y \in \Omega_Y$ & class label \\
		$q \in \mathcal{Q}_\mathcal{X}$ & query \\
		$z \in \Omega_Z$ & annotation \\
		$c \in \Omega_Y$ & confidence score \\
		\hline 
		\rowcolor{Gray}
		\multicolumn{2}{|c|}{Random Variables} \\ 
		\hline
		$X \define (X_1, \dots, X_D)$ & random variables of all features \\
		$Y$ & random variable of true class labels \\
		$Z \define (Z_1, \dots, Z_M)$ & random variables of annotations \\
		$Q$ & random variable of queries \\
		$A$ & random variable of annotators \\
		$P \define (P_1, \dots, P_M)$ & random variables of annotators' performances \\
		\hline 
		\rowcolor{Gray}
		\multicolumn{2}{|c|}{Annotation Process} \\ 
		\hline
		$\mathcal{S}: \mathbb{N} \to \mathcal{P}({\mathcal{Q}_\mathcal{X} \times \mathcal{A}})$ & sequence of the annotation process \\
		$\mathcal{S}^*$ & optimal annotation sequence \\
		$t \in \mathbb{N}$ & (time) step \\
		$t_\mathcal{S}$ & last step of annotation sequence $\mathcal{S}$\\
		\hline
		\multicolumn{2}{|r|}{{\textsc{Continued on the next page.}}} \\ 
		\hline
		\bottomrule
	\end{tabular}
\end{table}

\begin{table}[!h]
	\ContinuedFloat
	\centering
	\caption{Part II: Overview of mathematical notation.}
	\begin{tabular}{|R{0.345\columnwidth}|L{0.555\columnwidth}|}
		\toprule
		\textbf{Symbol} & \textbf{Meaning} \\
		\midrule
		\hline
		\rowcolor{Gray}
		\multicolumn{2}{|c|}{Annotation Process} \\ 
		\hline
		$\mathcal{C}$ & constraints of the annotation process \\ 
		$\mathcal{D}(t)$ & data set obtained at begin of step $t$\\
		$\mathcal{U}(t)$ & non-annotated data set at begin of step $t$ \\
		$\mathcal{L}(t)$ & annotated data set at begin of step $t$ \\
		\multirowcell{1}[0pt][r]{$z_{lm}^{(t)}$} & annotation for query $q_l$ by annotator $a_m$ obtained during step $t$ \\
		\hline
		\rowcolor{Gray}
		\multicolumn{2}{|c|}{Costs} \\ 
		\hline
		$B \in \mathbb{R}_{>0}$ & annotation budget \\
		$\mathbf{C} \in \mathbb{R}_{\geq 0}^{C \times C}$ & cost matrix \\
		${N_m}^{(t)} \in \mathbb{N}$ & number of annotations of annotator~$a_m$ in data set~$D(t)$ \\
		$\text{MC}(\boldsymbol{\theta}_{\mathcal{D}(t)} \mid \boldsymbol{\kappa})$ & misclassification cost + hyperparameters $\boldsymbol{\kappa}$ \\
		$\text{AC}(\mathcal{D}(t) \mid \boldsymbol{\nu})$ & annotation cost + hyperparameters $\boldsymbol{\nu}$ \\
		$\boldsymbol{\nu}_m \in \mathbb{R}_{>0}$ & cost of obtaining an annotation from annotator $a_m$ \\
		$\boldsymbol{\nu}_l \in \mathbb{R}_{>0}$ & cost of obtaining an annotation from annotator query $q_l$ \\
		$\boldsymbol{\nu}_{lm} \in \mathbb{R}_{>0}$ & cost of obtaining an annotation from annotator $a_m$ for query $q_l$ \\
		$\nu_{\text{max}} \in \mathbb{R}_{>0}$ & user-defined maximum annotation cost \\
		\hline 
		\rowcolor{Gray}
		\multicolumn{2}{|c|}{Elements of Real-world AL Strategies} \\ 
		\hline
		$\boldsymbol{\theta}$ & parameters of classification model \\
		$\boldsymbol{\theta}_{\mathcal{D}(t)}$ & classification model trained on $\mathcal{D}(t)$ \\
		$\boldsymbol{\omega}$ & parameters of the annotator model \\
		$\boldsymbol{\omega}_{\mathcal{D}(t)}$ & annotator model trained on $\mathcal{D}(t)$ \\
		$\hat{y}(\mathbf{x} \mid \boldsymbol{\theta}_{\mathcal{D}(t)})$ & prediction of classification model for $\mathbf{x}$ \\
		$\hat{y}^{(i)}(\mathbf{x} \mid \boldsymbol{\theta}_{\mathcal{D}(t)}) \in \Omega_Y$ & prediction leading to the $i$-th lowest MC \\ 
		$\hat{y}^{(n_i)} \in \Omega_Y$ & prediction with $i$-th highest probability for~$\mathbf{x}_n$ \\
		$\text{Pr}(Y=y \mid X=\mathbf{x}, \boldsymbol{\theta}_{\mathcal{D}(t)})$ & class membership probability of class $y$\\
		$\phi: \mathcal{Q}_\mathcal{X} \to \mathcal{R}_\phi $ & query utility measure \\
		$\psi: \mathcal{Q}_\mathcal{X} \times \mathcal{A} \to \mathcal{R}_\psi $ & annotator performance measure \\
		\hline
		\rowcolor{Gray}
		\multicolumn{2}{|c|}{Other and Strategy-specific Symbols} \\ 
		\hline
		$\delta: \{\text{true}, \text{false}\} \rightarrow \{0, 1\}$ & indicator function \\
		$\mathcal{P}({\mathcal{M}})$ & power set of an arbitrary set $\mathcal{M}$ \\
		$\doteq, \not\doteq$ & Boolean comparison \\
		$||\cdot||$ & user-defined distance function \\
		$\sigma: \mathbb{R} \rightarrow [0, 1]$ & logistic function \\
		$H: [0, 1]^C \rightarrow \mathbb{R}$ & entropy function \\
		$\mathrm{rank}: \mathbb{R} \rightarrow \mathbb{N}$ & ranking function \\
		$r_1, \dots, r_O: \Omega_X \rightarrow \mathbb{R}$ & relative attribute predictors \\
		$\mathbf{u}_y \in \mathbb{R}^O$ & embedding of class $y$ \\
		$\mathbf{\hat{u}}(\mathbf{x} \mid \boldsymbol{\theta}_{\mathcal{D}(t)}) \in \mathbb{R}^O$  & prediction of multi-target regression model \\
		$k \in \mathbb{N}$ & number of NN \\
		$\mathcal{N}_{\mathbf{x}}, \mathcal{E}_{\mathbf{x}} \subset \mathcal{X}$ & set of similar/dissimilar instances regarding $\mathbf{x}$\\
		$\mathcal{N}^k_\mathbf{x} \subset \mathcal{X}$ & $k$-nearest annotated neighbors of $\mathbf{x}$ \\
		$\mathcal{N}_{\mathbf{x},m}^{k} \subset \mathcal{X}$ & $k$-nearest neighbors of $\mathbf{x}$ annotated by $a_m$ \\
		$\mathcal{N}_{\mathbf{x}, \mathcal{D}_{\text{init}}}^{k} \subset \mathcal{X}$ & $k$-nearest fully annotated neighbors of $\mathbf{x}$ in the data set $\mathcal{D}_{\text{init}}$ \\
		$\mathbf{D}_n \in [0, 1]^{O \times O}$ & matrix of pairwise absolute differences of the top $O$ predicted class-membership probabilities of $\mathbf{x}_n$ \\
		$\mathbf{S} \in \mathbb{R}^{N \times N}$ & similarity matrix of instances $\mathcal{X}$ \\
		\hline
		\bottomrule
	\end{tabular}
\end{table}

\begin{table}[!h]
	\ContinuedFloat
	\centering
	\caption{Part III: Overview of mathematical notation.}
	\begin{tabular}{|R{0.345\columnwidth}|L{0.555\columnwidth}|}
		\toprule
		\textbf{Symbol} & \textbf{Meaning} \\
		\midrule
		\hline
		\rowcolor{Gray}
		\multicolumn{2}{|c|}{Other and Strategy-specific Symbols} \\ 
		\hline
		$\mathcal{A}^\prime_\text{acc} \in [0, 1]$ & estimated accuracy of the annotators' $\mathcal{A}^\prime \subseteq \mathcal{A}$ majority vote \\
		$\alpha_0, \alpha_1 \in \mathbb{R}$ & coefficients of a linear transformation \\
		$c_\text{min} \in [0.5, 1]$ & hyperparameter for minimum certainty threshold \\
		$\delta_d \in {0, 1}$ & indicator whether the feature $X_d$ has a positive value or not \\  
		$\rho \in [0, 1], \lambda \in (0, 1)$ & list of context-sensitive hyperparameters \\
		$\epsilon \in [0, 1)$ & hyperparameter of $\epsilon$-greedy annotator selection \\
		$\rho \in (0.5, 1)$ & quantile of t-student distribution \\
		$\mathcal{R} \subseteq \{1, \dots, D\}$ & index set of features \\
		$\mathcal{I} \subseteq \{1, \dots, N\}$ & index set of observed instances \\
		$\mathcal{K} = \{\mathcal{K}_1, \dots, \mathcal{K}_O\}$, {$\mathcal{K} \subseteq \mathcal{P}({\Omega_Y})$} & set of composite classes \\
		$\mathcal{G}_i \subseteq \mathcal{X}$ & instance being part of the region defined through a region query $q_i$ \\
		$\mathcal{D}_{\text{init}}$ & initiall fully annotated data set \\
		$e_m \in \mathbb{R}$ & expertise of annotator $a_m$ \\
		$d_n \in \mathbb{R}_{>0}$ & difficulty of annotating instance $\mathbf{x}_n$ \\
		$\mathbf{w} = (w_1, \dots, w_O)^\mathrm{T}$, {$\mathbf{w} \in \mathbb{R}^O$} & context-sensitive weight vector \\
		$\mathbf{p} = (p_1, \dots, p_C)^\mathrm{T}$, {$\mathbf{p} \in [0, 1]^C$} & vector of class probabilities \\
		$\mathbf{f}_{nm} \in \mathbb{R}_{\geq 0}^2$ & kernel frequency estimates regarding instance $\mathbf{x}_n$ and annotator $a_m$ \\
		$\boldsymbol{\beta} \in \mathbb{R}_{>0}^2$ & prior parameters for Beta distribution \\
		\hline
		\bottomrule
	\end{tabular}
\end{table}

\begin{table*}[!p]
	\caption{Part I: List of \acronym strategies reviewed in this survey.}
	\label{tab:strategies-overview}
	\footnotesize
	\begin{tabularx}{\textwidth}{|L{.225\textwidth}|c|L{.165\textwidth}|C{.04\textwidth}|C{.075\textwidth}|C{.069\textwidth}|C{.069\textwidth}|Y|}
		\toprule
		\multicolumn{1}{|c|}{\textbf{Strategy}} & \textbf{Year} & \multicolumn{1}{c|}{\textbf{Acronym}} & \textbf{Cost Types} & \textbf{Interaction Scheme} & \textbf{Annotator Model} & \textbf{Selection Algorithm} & \multicolumn{1}{c|}{\textbf{Appendix}} \\
		\midrule
		\hline
		\rowcolor{Gray}
		\citetsupplement{supplementary-Herde2021}                           & 2021 & MaPAL                              & \xmark & \xmark & \cmark & \cmark & \ref{subappendix:query-dependent-annotator-performance}, \ref{subappendix:joint-selection} \\
		\hline
		\citetsupplement{supplementary-Hopkins2020}                         & 2020 & N/A                                & \xmark & \cmark & \xmark & \xmark & \ref{subappendix:comparison_queries} \\
		\hline
		\rowcolor{Gray}
		\citetsupplement{supplementary-Chakraborty2020}                     & 2020 & N/A                                & \cmark & \xmark & \cmark & \cmark & \ref{subappendix:annotation-cost}, \ref{subappendix:query-dependent-annotator-performance}, \ref{subappendix:joint-selection} \\
		\hline
		\citetsupplement{supplementary-Min2019}                             & 2019 & TALK                               & \cmark & \xmark & \xmark & \xmark & \ref{subappendix:misclassification-cost} \\
		\hline
		\rowcolor{Gray}
		\citetsupplement{supplementary-Wu2019}                              & 2019 & CADU                               & \cmark & \xmark & \xmark & \xmark & \ref{subappendix:misclassification-cost} \\
		\hline
		\citetsupplement{supplementary-Wang2019}                            & 2019 & CATS                               & \cmark & \xmark & \xmark & \xmark & \ref{subappendix:misclassification-cost} \\
		\hline
		\rowcolor{Gray}
		\citetsupplement{supplementary-Krishnamurthy2017,supplementary-Krishnamurthy2019} & 2019 & COAL                               & \cmark & \xmark & \xmark & \xmark & 
		\ref{subappendix:misclassification-cost} \\
		\hline
		\citetsupplement{supplementary-Tsou2019}                            & 2019 & CSTS                               & \cmark & \xmark & \xmark & \xmark & \ref{subappendix:annotation-cost} \\
		\hline
		\rowcolor{Gray}
		\citetsupplement{supplementary-Hu2019}                              & 2019 & ALPF                               & \xmark & \cmark & \xmark & \xmark & \ref{subappendix:instance_queries} \\
		\hline
		\citetsupplement{supplementary-Bhattacharya2019}                    & 2019 & N/A                                & \xmark & \cmark & \xmark & \xmark & \ref{subappendix:instance_queries} \\
		\hline
		\rowcolor{Gray}
		\citetsupplement{supplementary-Teso2019}                            & 2019 & N/A                                & \xmark & \cmark & \xmark & \xmark & \ref{subappendix:instance_queries} \\
		\hline
		\citetsupplement{supplementary-Luo2018,supplementary-Hauskrecht2018,supplementary-Luo2019}      & 2019 & HALG \& ($\text{A}^*$)HALR         & \xmark & \cmark & \xmark & \xmark & \ref{subappendix:region_queries} \\
		\hline
		\rowcolor{Gray}
		\citetsupplement{supplementary-Calma2018a}                          & 2018 & N/A                                & \xmark & \cmark & \xmark & \xmark & \ref{subappendix:instance_queries} \\
		\hline
		\citetsupplement{supplementary-Song2018}                            & 2018 & N/A                                & \xmark & \cmark & \xmark & \xmark & \ref{subappendix:instance_queries} \\
		\hline
		\rowcolor{Gray}
		\citetsupplement{supplementary-Yang2018}                            & 2018 & DALC                                & \xmark & \xmark & \cmark & \cmark & \ref{subappendix:query-dependent-annotator-performance}, \ref{subappendix:sequential-selection} \\
		\hline
		\citetsupplement{supplementary-Huang2017}                           & 2017 & CEAL                               & \cmark & \xmark & \cmark & \cmark & \ref{subappendix:annotation-cost}, \ref{subappendix:query-dependent-annotator-performance}, \ref{subappendix:joint-selection} \\
		\hline
		\rowcolor{Gray}
		\citetsupplement{supplementary-Kane2017}                            & 2017 & ACCQ                               & \xmark & \cmark & \xmark & \xmark & \ref{subappendix:comparison_queries} \\
		\hline
		\citetsupplement{supplementary-Xu2017}                              & 2017 & ADGAC                              & \xmark & \cmark & \xmark & \xmark & \ref{subappendix:comparison_queries} \\
		\hline
		\rowcolor{Gray}
		\citetsupplement{supplementary-Huang2016a}                          & 2016 & N/A                                & \cmark & \xmark & \xmark & \xmark & \ref{subappendix:misclassification-cost} \\
		\hline
		\citetsupplement{supplementary-Long2013,supplementary-Long2016}                   & 2016 & JGPC-ASAL                                & \xmark & \xmark & \cmark & \cmark & \ref{subappendix:uniform-annotator-performance}, \ref{subappendix:sequential-selection} \\
		\hline
		\rowcolor{Gray}
		\citetsupplement{supplementary-Long2015}                   & 2015 & MARMGPC-ASAA                               & \xmark & \xmark & \cmark & \cmark & \ref{subappendix:uniform-annotator-performance}, \ref{subappendix:sequential-selection} \\
		\hline
		\citetsupplement{supplementary-Krempl2015}                          & 2015 & OPAL                               & \cmark & \xmark & \xmark & \xmark & \ref{subappendix:misclassification-cost} \\
		\hline
		\rowcolor{Gray}
		\citetsupplement{supplementary-Nguyen2015}                          & 2015 & N/A                                & \cmark & \xmark & \cmark & \xmark & \ref{subappendix:misclassification-cost}, \ref{subappendix:annotation-cost}, \ref{subappendix:annotation-dependent-annotator-performance} \\
		\hline
		\citetsupplement{supplementary-Kaeding2015}                         & 2015 & $\text{GP-EMOC}_{\text{PDE+R}}$    & \cmark & \cmark & \cmark & \xmark & \ref{subappendix:misclassification-cost}, \ref{subappendix:instance_queries}, \ref{subappendix:query-dependent-annotator-performance} \\
		\hline
		\rowcolor{Gray}
		\citetsupplement{supplementary-Zhong2015}                           & 2015 & ALCU-SVM                           & \xmark & \cmark & \cmark & \xmark &  \ref{subappendix:instance_queries}, \ref{subappendix:query-dependent-annotator-performance} \\
		\hline
		\citetsupplement{supplementary-Xiong2015}                           & 2015 & N/A                                & \xmark & \cmark & \xmark & \xmark & \ref{subappendix:comparison_queries} \\
		\hline
		\rowcolor{Gray}
		\citetsupplement{supplementary-Qian2015}                            & 2015 & ARP                                & \xmark & \cmark & \xmark & \xmark & \ref{subappendix:comparison_queries} \\
		\hline
		\citetsupplement{supplementary-Moon2014}                            & 2014 & N/A                                & \cmark & \xmark & \cmark & \cmark & \ref{subappendix:annotation-cost}, \ref{subappendix:annotation-dependent-annotator-performance}, \ref{subappendix:joint-selection} \\
		\hline
		\rowcolor{Gray}
		\citetsupplement{supplementary-Fu2011,supplementary-Fu2014}                       & 2014 & QHAL \& PHAL                       & \xmark & \cmark & \xmark & \xmark & \ref{subappendix:comparison_queries} \\
		\hline
		\citetsupplement{supplementary-Rodrigues2014}                       & 2014 & GPC-MA                             & \xmark & \xmark & \cmark & \cmark & \ref{subappendix:annotation-dependent-annotator-performance}, \ref{subappendix:sequential-selection} \\
		\hline
		\rowcolor{Gray}
		\citetsupplement{supplementary-Fang2013e}                           & 2014 & EIAL                               & \xmark & \xmark & \cmark & \xmark & \ref{subappendix:instance_queries}, \ref{subappendix:query-dependent-annotator-performance} \\
		\hline
		\citetsupplement{supplementary-Fang2013,supplementary-Fang2014}                   & 2014 & AL+kTrM \& ALM+TrU                 & \xmark & \xmark & \cmark & \cmark & \ref{subappendix:sequential-selection}, \ref{subappendix:query-dependent-annotator-performance} \\
		\hline
		\rowcolor{Gray}
		\citetsupplement{supplementary-Zhao2014}                            & 2014 & N/A                                & \xmark & \xmark & \cmark & \cmark & \ref{subappendix:query-dependent-annotator-performance}, \ref{subappendix:sequential-selection} \\
		\hline
		\citetsupplement{supplementary-Chen2013}                            & 2013 & MEC \& CWMM                        & \cmark & \xmark & \xmark & \xmark & \ref{subappendix:misclassification-cost} \\
		\hline
		\rowcolor{Gray}
		\citetsupplement{supplementary-Biswas2013}                          & 2013 & N/A                                & \xmark & \cmark & \xmark & \xmark & \ref{subappendix:instance_queries} \\
		\hline
		\citetsupplement{supplementary-Wu2013b}                             & 2013 & PMActive                           & \xmark & \xmark & \cmark & \cmark & \ref{subappendix:annotation-dependent-annotator-performance}, \ref{subappendix:sequential-selection} \\
		\hline
		\rowcolor{Gray}
		\citetsupplement{supplementary-Haque2013}                           & 2013 & GQAL                               & \xmark & \cmark & \xmark & \xmark & \ref{subappendix:region_queries} \\
		\hline
		\citetsupplement{supplementary-Joshi2010,supplementary-Joshi2012}                 & 2012 & N/A                                & \cmark & \cmark & \xmark & \xmark & \ref{subappendix:misclassification-cost}, \ref{subappendix:annotation-cost}, \ref{subappendix:comparison_queries} \\
		\hline
		\rowcolor{Gray}
		\citetsupplement{supplementary-Cebron2012}                          & 2012 & N/A                                & \xmark & \cmark & \xmark & \xmark & \ref{subappendix:instance_queries} \\
		\hline
		\citetsupplement{supplementary-Ni2012}                              & 2012 & BMO                                & \xmark & \cmark & \cmark & \cmark & \ref{subappendix:instance_queries}, \ref{subappendix:query-dependent-annotator-performance}, \ref{subappendix:sequential-selection} \\
		\hline
		\rowcolor{Gray}
		\citetsupplement{supplementary-Fang2012}                            & 2012 & STAL                               & \xmark & \xmark & \cmark & \cmark & \ref{subappendix:query-dependent-annotator-performance}, \ref{subappendix:sequential-selection} \\
		\hline
		\citetsupplement{supplementary-Yan2012b}                             & 2012 & N/A                                & \xmark & \xmark & \cmark & \cmark & \ref{subappendix:query-dependent-annotator-performance}, \ref{subappendix:joint-selection} \\
		\hline
		\rowcolor{Gray}
		\citetsupplement{supplementary-Yan2011}                             & 2011 & N/A                                & \xmark & \xmark & \cmark & \cmark & \ref{subappendix:query-dependent-annotator-performance}, \ref{subappendix:joint-selection} \\
		\hline
		\citetsupplement{supplementary-Wallace2011}                         & 2011 & MEAL                               & \cmark & \cmark & \cmark & \cmark & \ref{subappendix:annotation-cost},  \ref{subappendix:instance_queries}, \ref{subappendix:query-dependent-annotator-performance}, \ref{subappendix:sequential-selection} \\
		\hline
		\rowcolor{Gray}
		\citetsupplement{supplementary-Settles2011a}                        & 2011 & DUALIST                            & \xmark & \cmark & \xmark & \xmark & \ref{subappendix:region_queries} \\
		\hline
		\citetsupplement{supplementary-Rashidi2011}                         & 2011 & RIQY                               & \xmark & \cmark & \xmark & \xmark & \ref{subappendix:region_queries} \\
		\hline
		\rowcolor{Gray}
		\citetsupplement{supplementary-Zheng2010}                           & 2010 & IEAdjCost                          & \cmark & \xmark & \cmark & \cmark & \ref{subappendix:annotation-cost}, \ref{subappendix:uniform-annotator-performance}, \ref{subappendix:sequential-selection} \\
		\hline
		\citetsupplement{supplementary-Donmez2008b,supplementary-Donmez2010b}             & 2010 & N/A                                & \cmark & \cmark & \cmark & \cmark & \ref{subappendix:annotation-cost},  \ref{subappendix:instance_queries}, \ref{subappendix:query-dependent-annotator-performance}, \ref{subappendix:joint-selection} \\
		\hline
		\rowcolor{Gray}
		\citetsupplement{supplementary-Tomanek2010}                         & 2010 & N/A                                & \cmark & \xmark & \xmark & \xmark & \ref{subappendix:annotation-cost} \\
		\hline
		\citetsupplement{supplementary-Wallace2010}                         & 2010 & N/A                                & \cmark & \xmark & \xmark & \xmark & \ref{subappendix:annotation-cost} \\
		\hline
		\rowcolor{Gray}
		\citetsupplement{supplementary-Du2010}                              & 2010 & N/A                                & \xmark & \xmark & \cmark & \xmark & \ref{subappendix:query-dependent-annotator-performance} \\
		\hline
		\citetsupplement{supplementary-Donmez2010}                          & 2010 & SFilter                            & \xmark & \xmark & \cmark & \xmark & \ref{subappendix:uniform-annotator-performance} \\
		\hline
		\multicolumn{8}{|r|}{{\textsc{Continued on the next page.}}} \\ 
		\hline
		\bottomrule
	\end{tabularx}
\end{table*}
\begin{table*}[!h]
	\ContinuedFloat
	\caption{Part II: List of \acronym strategies reviewed in this survey.}
	\footnotesize
	\begin{tabularx}{\textwidth}{|L{.225\textwidth}|c|L{.165\textwidth}|C{.04\textwidth}|C{.075\textwidth}|C{.069\textwidth}|C{.069\textwidth}|Y|}
		\toprule
		\multicolumn{1}{|c|}{\textbf{Strategy}} & \textbf{Year} & \multicolumn{1}{c|}{\textbf{Acronym}} & \textbf{Cost Types} & \textbf{Interaction Scheme} & \textbf{Annotator Model} & \textbf{Selection Algorithm} & \multicolumn{1}{c|}{\textbf{Appendix}} \\
		\midrule
		\hline
		\rowcolor{Gray}
		\citetsupplement{supplementary-Du2009,supplementary-Du2010a}                      & 2009 & AGQ \& $\text{AGQ}^{+}$                              & \xmark & \cmark & \xmark & \xmark & \ref{subappendix:region_queries} \\
		\hline
		\citetsupplement{supplementary-Liu2009}                             & 2009 & CS USST                            & \cmark & \xmark & \xmark & \xmark & \ref{subappendix:misclassification-cost} \\
		\hline
		\rowcolor{Gray}
		\citetsupplement{supplementary-Arora2009}                           & 2009 & N/A                                & \cmark & \xmark & \xmark & \xmark & \ref{subappendix:annotation-cost} \\
		\hline
		\citetsupplement{supplementary-Druck2009}                           & 2009 & GE WU                             & \xmark & \cmark & \xmark & \xmark & \ref{subappendix:region_queries} \\
		\hline
		\rowcolor{Gray}
		\citetsupplement{supplementary-Donmez2009}                          & 2009 & IEThresh                           & \xmark & \xmark & \cmark & \cmark & \ref{subappendix:uniform-annotator-performance}, \ref{subappendix:sequential-selection} \\
		\hline
		\citetsupplement{supplementary-Settles2008a}                        & 2008 & N/A                                & \cmark & \xmark & \xmark & \xmark & \ref{subappendix:annotation-cost} \\
		\hline
		\rowcolor{Gray}
		\citetsupplement{supplementary-Haertel2008}                         & 2008 & N/A                                & \cmark & \xmark & \xmark & \xmark & \ref{subappendix:annotation-cost} \\
		\hline
		\citetsupplement{supplementary-Margineantu2005}                     & 2005 & ACTIVE-CSL                         & \cmark & \xmark & \xmark & \xmark & \ref{subappendix:misclassification-cost} \\
		\hline
		\bottomrule
	\end{tabularx}
\end{table*}

\section{Cost Types}
\label{appendix:cost-types}
In this appendix, we analyze concrete \acronym strategies regarding their handling of costs.
We structure this analysis according to the cost types, i.e., AC and MC, including their underlying cost schemes identified in Section~\ref{sec:types_of_cost} in the associated survey.

\subsection{Misclassification Cost}
\label{subappendix:misclassification-cost}
\textbf{Class-dependent MC:}
\citetsupplement{supplementary-Margineantu2005} proposed \textit{active cost-sensitive learning} (ACTIVE-CSL) as one of the first \acronym strategies taking class-dependent MC into account. 
It expects a cost matrix as input. Concerning the already annotated instances, ACTIVE-CSL computes the expected MC after annotating an instance and adding it to the classification model's training set. 
Since the annotation of an instance is not known in advance, ACTIVE-CSL takes the expectation over all possible annotations.
This query utility measure is similar to EER. 
As a result, it involves a lot of retraining and is thus computationally intensive.
Moreover, taking only the set of already annotated instances into account biases the estimation of the expected MC. 
This is, in particular, true in the early stage of the AL process, where only a few instances have been annotated.

\citetsupplement{supplementary-Joshi2010,supplementary-Joshi2012} proposed a similar cost-sensitive EER variant. 
In contrast to ACTIVE-CSL, this measure also exploits the non-annotated set of instances to evaluate the expected MC of the classification model.
Since the true class labels of these instances are unknown, it relies on the estimated class membership probabilities of the classification model after each retraining.
However, the high computational complexity of this strategy remains, as for ACTIVE-CSL, a limitation to train classification models with computation-intensive training procedures.

The more recent strategy \textit{optimized probabilistic active learning} (OPAL), proposed by \citetsupplement{supplementary-Krempl2015}, partially overcomes the issue of high computational complexity.
It computes the density-weighted reduction in the MC when annotating an instance.
Different from ACTIVE-CSL, the expected MC is computed regarding the candidate instance.
Therefor, it relies on so-called kernel frequency estimates.
We can interpret them as the number of annotations per class in the neighborhood of an instance.
They are often estimated through a kernel function quantifying similarities between instances.
The kernel frequency estimates allow for a closed-form solution for computing the expected MC.
Additionally, they can be easily updated when adding additional annotations.
Accordingly, OPAL is non-myopic by considering more than one annotation acquisition at once.
The major disadvantage of OPAL is its need for an appropriate kernel frequency estimation, which is difficult for domains such as images.
Another disadvantage is OPAL's restriction on binary classification problems.

Instead of computing the MC reduction, \citetsupplement{supplementary-Kaeding2015} proposed the \textit{expected model output change (EMOC)} as a utility measure. It quantifies how the classification model's predictions change by simulating the annotation of an instance. 
For this, it compares the updated and old classification model's predictions through a cost (loss) function and assigns high utilities to instances leading to significant differences between the prediction pairs.
Compared to an EER-based approach, EMOC does not need highly reliable estimates of the class membership probabilities to compute meaningful utilities.
Nevertheless, many retraining procedures of the classification model are required and hence lead to high computational complexity.

In favor of computational efficiency, \citetsupplement{supplementary-Liu2009} proposed the strategy \textit{cost-sensitive uncertainty sampling with self-training} (CS USST). 
Its idea is to use US to select instances for finding the decision boundary minimizing the misclassification rate. 
Subsequently, a classification model is trained on the update annotated set $\mathcal{L}$ and used to obtain predictions for the instances of the non-annotated set $\mathcal{U}$.
Finally, a cost-sensitive classification model is trained on the union of both sets, including the previously obtained predictions.
Using this semi-supervised learning approach to determine the parameters of a cost-sensitive classification model, CS USST aims to overcome the selection bias caused by taking only the instances selected by US into account.
Although this strategy resolves some limitations of US, the missing exploration issue of US remains.

\citetsupplement{supplementary-Chen2013} proposed two further uncertainty-based \acronym strategies, namely \textit{maximum expected cost} (MEC) and \textit{cost-weighted minimum margin} (CWMM).
MEC generalizes the minimum confidence variant of US, and its utility measure is defined through 
\begin{equation}
\label{eq:mec}
\begin{gathered}
\phi_{\text{MEC}}(\mathbf{x} \mid \boldsymbol{\theta}_{\mathcal{D}}) \define \\\sum_{y \in \Omega_Y} \Pr(Y=y \mid X=\mathbf{x},  \boldsymbol{\theta}_{\mathcal{D}}) \mathbf{C}[y, \hat{y}(\mathbf{x} \mid \boldsymbol{\theta}_{\mathcal{D}})],
\end{gathered}
\end{equation}
where $\mathbf{C} \in \mathbb{R}_{\geq 0}^{C \times C}$ is a user-defined cost matrix.
In contrast, CWMM is a generalization of the minimum margin US variant.
It computes the MC difference between the prediction with the lowest~$\hat{y}^{(1)} \in \Omega_Y$ and second lowest cost~$\hat{y}^{(2)} \in \Omega_Y$:
\begin{equation}
\label{eq:cwmm}
\begin{gathered}
\phi_{\text{CWMM}}(\mathbf{x} \mid \boldsymbol{\theta}_{\mathcal{D}}) \define \sum_{y \in \Omega_Y} \Pr(Y=y \mid X=\mathbf{x}, \boldsymbol{\theta}_{\mathcal{D}}) \\(\mathbf{C}[y, \hat{y}^{(1)}(\mathbf{x} \mid \boldsymbol{\theta}_{\mathcal{D}})]-\mathbf{C}[y, \hat{y}^{(2)}(\mathbf{x} \mid \boldsymbol{\theta}_{\mathcal{D}})]),
\end{gathered}
\end{equation}
Both utility measures are easy to compute and can be used in combination with any probabilistic classification model.
However, they share similar disadvantages as US-based utility measures, e.g., no consideration of representativeness and missing exploration.

\citetsupplement{supplementary-Huang2016a} introduced a different way of calculating uncertainty. 
The strategy \textit{active learning with cost embedding} (ALCE) is based on a cost embedding approach which transfers the cost information into a distance measure of an $(O \in \mathbb{N})$-dimensional latent space.
Therefor, the class labels in $\Omega_Y$ are encoded as vectors $\mathbf{u_1}, \dots, \mathbf{u}_C \in \mathbb{R}^O$ such that $||\mathbf{u}_i - \mathbf{u}_j|| < ||\mathbf{u}_k - \mathbf{u}_l||$ holds for the Euclidean distance if and only if $\mathbf{C}[i, j] < \mathbf{C}[k, l]$.
A cost-sensitive classifier is implemented through a multi-target regression model predicting latent vectors ${\hat{\mathbf{u}}(\mathbf{x} \mid \boldsymbol{\theta}_{\mathcal{D}}) \in \mathbb{R}^O}$. The final class label prediction 
\begin{equation}
\hat{y}(\mathbf{x} \mid \boldsymbol{\theta}_{\mathcal{D}}) \define \argmin_{y \in \Omega_Y} \left(||\mathbf{u}_y - \mathbf{\hat{u}}(\mathbf{x} \mid \boldsymbol{\theta}_{\mathcal{D}})||\right)
\end{equation}
is obtained by finding the nearest vector of the transformed class labels.
This embedding allows ALCE to define the cost-sensitive uncertainty measure
\begin{equation}
\label{eq:alce}
\phi_{\text{ALCE}}(\mathbf{x} \mid \boldsymbol{\theta}_{\mathcal{D}}) \define \min_{y \in \Omega_Y} \left(||\mathbf{u}_y - \mathbf{\hat{u}}(\mathbf{x} \mid \boldsymbol{\theta}_{\mathcal{D}})||\right).
\end{equation}
It assigns high utility to instances whose expected costs are high.
On the one hand, ALCE overcomes the requirement of a probabilistic classifier.
On the other hand, a suitable multi-target regression model is required instead.

\citetsupplement{supplementary-Nguyen2015} proposed a strategy whose utility measure quantifies the expected MC reduction for querying an instance's annotation from cheap crowd workers or expensive experts.
It separates candidate instances into non-annotated instances and those annotated through the majority vote annotation obtained from multiple crowd workers.
A non-annotated instance can be selected for annotation through the crowd workers, whereas an annotated instance can be chosen to obtain an annotation from an expert.
The latter case can be important for instances whose crowd worker annotations are likely wrong.
The computation of the expected MC follows the idea of a cost-sensitive EER variant and extends it by modeling the  crowd workers' error-proneness.
Moreover, the strategy pre-selects a set of candidate instances (according to the selection criterion of US) to reduce its computation complexity.
Nevertheless, it involves multiple retraining iterations of a probabilistic classification model.

Following a methodology of divide and conquer~\citesupplement{supplementary-Yao2009}, \citetsupplement{supplementary-Min2019} proposed the strategy \textit{tri-partition active learning through $k$-NN}~(TALK).
Based on a cost-sensitive $k \in \mathbb{N}$-\textit{nearest neighbor} ($k$-NN) model, TALK divides the non-annotated instances into three regions. 
For this, the MC of each non-annotated instance is estimated as instance utility through
\begin{equation}
\label{eq:talk}
\phi_{\text{TALK}}(\mathbf{x} \mid \boldsymbol{\theta}_{\mathcal{D}}) \define \frac{1}{k} \sum_{\mathbf{x}_n \in \mathcal{N}^k_\mathbf{x}} \mathbf{C}[\hat{y}(\mathbf{x} \mid \boldsymbol{\theta}_{\mathcal{D}}), y_n].
\end{equation}
The set $\mathcal{N}^k_\mathbf{x} \subset \mathcal{X}$ contains the $k$ annotated NN of instance $\mathbf{x}$. 
If the estimated MC of an instance is lower than its AC, the instance is assigned to the first region and annotated according to the classification model's prediction.
The remaining non-annotated instances are sorted according to their estimated MCs.
A predefined number of the instances with the highest MCs are assigned to the second region and annotated by a human annotator.
The other non-annotated instances form the third region and will be processed in the next cycle iteration.
The major issues of TALK are its requirement for an appropriate $k$-NN model and its missing consideration of an instance's representativeness.

Two similar strategies to TALK are \textit{cost-sensitive active learning through density clustering under a label uniform distribution} (CADU), proposed by~\citetsupplement{supplementary-Wu2019}, and \textit{cost-sensitive active learning through statistical methods} (CATS), presented by ~\citetsupplement{supplementary-Wang2019}. They additionally consider an instance's density and aim to query the annotations of an MC-optimal number of instances per region/cluster.

\textbf{Instance-dependent MC:}
Each of the previously discussed strategies assumes class-dependent MC, e.g., defined by a cost matrix.
In contrast, \citetsupplement{supplementary-Krishnamurthy2017,supplementary-Krishnamurthy2019} presented \textit{cost overlapped active learning} (COAL) as an approach for instance-dependent MC.
Correspondingly, the cost of interchanging two classes may differ from instance to instance.
COAL requires access to a set of regression functions, which provide the range of possible costs that a predicted class label for an instance may cause. 
The idea of COAL is to actively query the cost of predicting a specific class label for an instance instead of querying class labels.
Therefore, COAL aims to query only the cost information of class labels with high uncertainty regarding their possible cost values.
This uncertainty is quantified through a cost range that we can interpret as COAL's utility measure.

\subsection{Annotation Cost}
\label{subappendix:annotation-cost}

\textbf{Annotator-dependent AC:} In the following, we denote the annotators' individual ACs as $\boldsymbol{\nu} \define \left(\nu_1, \dots, \nu_M\right)^\mathrm{T} \in \mathbb{R}_{>0}^{M}$ with $\nu_m$ as the AC for querying annotator $a_m \in \mathcal{A}$.

For this setting, \citetsupplement{supplementary-Zheng2010} proposed the strategy IEAdjCost. It solves an optimization problem to determine a subset of annotators whose majority vote annotations achieve in average a user-defined level of accuracy $\rho \in [0, 1]$ for the minimum sum of the annotators' ACs.
Mathematically, this annotator set is defined through
\begin{equation}
\label{eq:ieadjcost}
\argmin_{\mathcal{A}^\prime \subseteq \mathcal{A}}\left(\sum_{a_m \in \mathcal{A}^\prime} \nu_m\right) \text{ subject to } \mathcal{A}^\prime_\text{acc} \geq \rho,
\end{equation}
where $\mathcal{A}^\prime_\text{acc}$ denotes the estimated probability that the majority vote annotation of $\mathcal{A}^\prime$ for an arbitrary instance will be correct.

The strategy of \citetsupplement{supplementary-Chakraborty2020} also solves an optimization problem to determine in each iteration cycle a set of annotators with high performances and low ACs to annotate a batch of queries.
Therefor, \citeauthor{supplementary-Chakraborty2020} combines query utility, annotator performance, and AC into one final score, i.e., the annotation error-weighted AC normalized by the query utility.

Using the annotators' ACs as a normalization factor to assess the utility of a query or the annotators' performances is another method to consider annotator-dependent AC explicitly.
The strategies, proposed in \citesupplement{supplementary-Huang2017,supplementary-Moon2014,supplementary-Donmez2008b,supplementary-Donmez2010b}, compute a kind of AC-effective performance:
\begin{equation}
\frac{\psi(q_l, a_m \mid \boldsymbol{\omega}_{\mathcal{D}})}{\nu_m}.
\end{equation}
One disadvantage of such a normalization approach may be the non-linear relation between annotator performance and AC since they are computed on different scales.
In contrast, the strategy of \citetsupplement{supplementary-Nguyen2015} uses the AC to normalize the expected reduction in MC.
Both cost types are estimated on the same scale.
As a result, this strategy may find a better trade-off between both cost types.

\textbf{Query-dependent AC:} In the following, we denote the queries' individual ACs as $\boldsymbol{\nu} \define \left(\nu_1, \dots, \nu_L\right)^\mathrm{T} \in \mathbb{R}_{>0}^{L}$ with $\nu_l$ as the AC for obtaining an annotation for query $q_l \in \mathcal{Q}_{\mathcal{X}}$.
Many strategies~\citesupplement{supplementary-Joshi2010,supplementary-Joshi2012,supplementary-Donmez2008b,supplementary-Donmez2010b,supplementary-Margineantu2005} take query-dependent AC into account by subtracting the AC of a query from its utility:
\begin{equation}
\label{eq:utility-minus-AC-unit}
\phi(q_l \mid \boldsymbol{\theta}_{\mathcal{D}}) - \nu_l.
\end{equation}
Computing query utility per AC unit is another common approach to make a strategy cost-sensitive~\citesupplement{supplementary-Settles2008a,supplementary-Haertel2008,supplementary-Tomanek2010,supplementary-Tsou2019,supplementary-Wallace2010}:
\begin{equation}
\label{eq:utility-per-AC-unit}
\frac{\phi(q_l \mid \boldsymbol{\theta}_{\mathcal{D}})}{\nu_l}.
\end{equation}
The approaches in Eq.~\ref{eq:utility-minus-AC-unit} and Eq.~\ref{eq:utility-per-AC-unit} implicitly assume that the utilities and ACs can be expressed on the same scale~\citesupplement{supplementary-Tomanek2010}.
If this is assumption is violated, one may re-scale the AC, e.g., through a linear transformation:
\begin{equation}
\label{eq:utility-per-transformed-AC-unit}
\frac{\phi(q_l \mid \boldsymbol{\theta}_{\mathcal{D}})}{\alpha_0 \cdot \nu_l + \alpha_1},
\end{equation}
where $\alpha_0,\alpha_1 \in \mathbb{R}$ are corresponding coefficients to be determined.
However, these coefficients are often unknown in real-world settings, or even non-linear transformations of the ACs are required~\citesupplement{supplementary-Haertel2008}.
For this reason, \citetsupplement{supplementary-Haertel2008} proposed two other approaches to consider query-dependent AC explicitly.
On the one hand, the selection of queries can be constrained to a user-defined maximum AC $\nu_{\text{max}} \in \mathbb{R}_{>0}$ such that all queries with $\nu_l > \nu_{\text{max}}$ are excluded from the annotation process.
On the other hand, a query's utility and its AC can be combined through a linear combination of their ranks.
In this context, a high query utility and a low AC leads to high ranks:
\begin{equation}
\label{eq:utility-AC-rank}
\rho\cdot\mathrm{rank}(\phi(q_l \mid \boldsymbol{\theta}_{\mathcal{D}})) + (1-\rho)\cdot\mathrm{rank}(-\nu_l)
\end{equation}
with $\rho \in [0, 1]$ as a user-defined weighting term.
Both approaches have the disadvantage of finding appropriate values for $\rho$ and $\nu_{\text{max}}$.

Knowledge about the AC of each query is a prerequisite for employing the above approaches as part of a \acronym strategy.
Therefore, \citetsupplement{supplementary-Margineantu2005} assumes that the AC per query is known in advance.
In contrast, \citetsupplement{supplementary-Donmez2008b,supplementary-Donmez2010b} and \citetsupplement{supplementary-Joshi2010,supplementary-Joshi2012} expect that the AC follows a fixed cost model where the AC of a query is correlated to the probability of obtaining its optimal annotation.

More advanced strategies~\citesupplement{supplementary-Tsou2019,supplementary-Tomanek2010,supplementary-Wallace2010,supplementary-Settles2008a,supplementary-Haertel2008} estimate the individual ACs at run-time during the annotation process.
In particular, when we define AC through the annotation time, such an estimation is crucial.
\citetsupplement{supplementary-Settles2008a} provided a detailed analysis of annotation time used as a proxy for the AC.
For this purpose, four data sets with four different tasks (i.e., extracting entities from news articles, classifying biomedical abstracts, extracting contact details from e-mail signature lines, and classifying segments in images) were annotated.
The corresponding annotation times were logged. 
The results indicate a high variability of the annotation time per query on all four data sets.
Moreover, the annotation time is found to be substantially annotator-dependent.
Another investigated issue concerns the development of the annotation time during the annotation process.
It has been observed that, in general, the annotation time decreases because the annotators can adapt quickly to the annotation task. 
The observations regarding the annotation time made by~\citetsupplement{supplementary-Settles2008a} have been mostly confirmed by~\citetsupplement{supplementary-Arora2009} in another empirical investigation.
In a further case study, \citetsupplement{supplementary-Raghavan2006} found that the annotation time strongly depends on the type of query, i.e., annotating the importance of a feature regarding a document classification problem took about one-fifth of the time required for annotating a document.

Following the above studies, there are several methods for estimating times for annotating documents~\citesupplement{supplementary-Wallace2010,supplementary-Settles2008a,supplementary-Haertel2008}. 
They estimate the annotation time using a regression model, e.g., support vector regression~\citesupplement{supplementary-Smola2004} or ordinary least squares~\citesupplement{supplementary-Nievergelt2000}, as a function of simple numerical features such as the number of words in a document.
Experiments with these models demonstrated that annotation times are fairly learnable.

\citetsupplement{supplementary-Tsou2019} proposed a more general approach for estimating query-dependent AC. Their \acronym strategy \textit{cost-sensitive tree sampling} (CSTS) constructs a decision tree dividing the observed instances into disjoint leaves during the AL process. The AC of querying an instance's annotation is then estimated through the average AC of the already annotated instances in the corresponding leaf of this tree.

\textbf{Query- and annotator-dependent AC:} In the following, we denote the ACs for each combination of query and annotator as $\boldsymbol{\nu} \define \left(\nu_{11}, \dots, \nu_{LM}\right)^\mathrm{T} \in \mathbb{R}_{>0}^{L}$ with $\nu_{lm}$ as the AC for obtaining an annotation for query $q_l \in \mathcal{Q}_{\mathcal{X}}$ from annotator~$a_m \in \mathcal{A}$.

\citetsupplement{supplementary-Wallace2010} proposed a straightforward approach to compute the AC per query-annotator pair in the domain of document classification.
Therefor, they assume that the salary per time unit of each annotator is known.
As a result, the AC $\nu_{lm}$ is estimated by multiplying the expected time to annotate query $q_l$ by the salary per time unit of annotator~$a_m$.
To estimate a query's annotation time, \citeauthor{supplementary-Wallace2010} simply expect that all annotators read a predefined number of words per minute and transform the length of a document to an annotation time under this model.

The assumption that each annotator requires similar annotation times is often violated in real-world applications~\citesupplement{supplementary-Arora2009,supplementary-Settles2008a}.
For this reason, \citetsupplement{supplementary-Arora2009} proposed an approach to estimate the annotation time as a function of query and annotator characteristics.
Since this approach also focuses on documents, they use character length and percentage of stop words to describe a document's query.
For the annotators, an ordinal scale is used to assess whether an annotator is a native speaker of English.
Combined with respective logged annotation times, these characteristics are used as inputs to a linear or support vector regression model~\citesupplement{supplementary-Smola2004} to estimate the annotation time for new query-annotator pairs.
A significant advantage of such an approach is its inductive learning toward annotators for which no annotation times have been collected yet.
However, the approach was only tested on a relatively small data set.

\section{Interaction Schemes}
\label{appendix:interaction-schemes}
In this appendix, we analyze concrete \acronym strategies regarding their interaction schemes.
We structure this analysis according to the query types, i.e., instance, region, and comparison query, identified in Section~\ref{sec:types_of_queries_and_annotations} in the associated survey.

\subsection{Instance Queries}
\label{subappendix:instance_queries}
Instance queries are the core of research in traditional AL and have already been reviewed in several other surveys~\citesupplement{supplementary-Kumar2020,supplementary-Aggarwal2014,supplementary-Fu2013,supplementary-Settles2010}.
Therefore, we focus on the following strategies using queries going beyond the standard formulation: ``To which class does instance~$\mathbf{x}_n \in \mathcal{X}$ belong?''.

\begin{figure}[h!]
	\centering
	\includegraphics[width=\columnwidth]{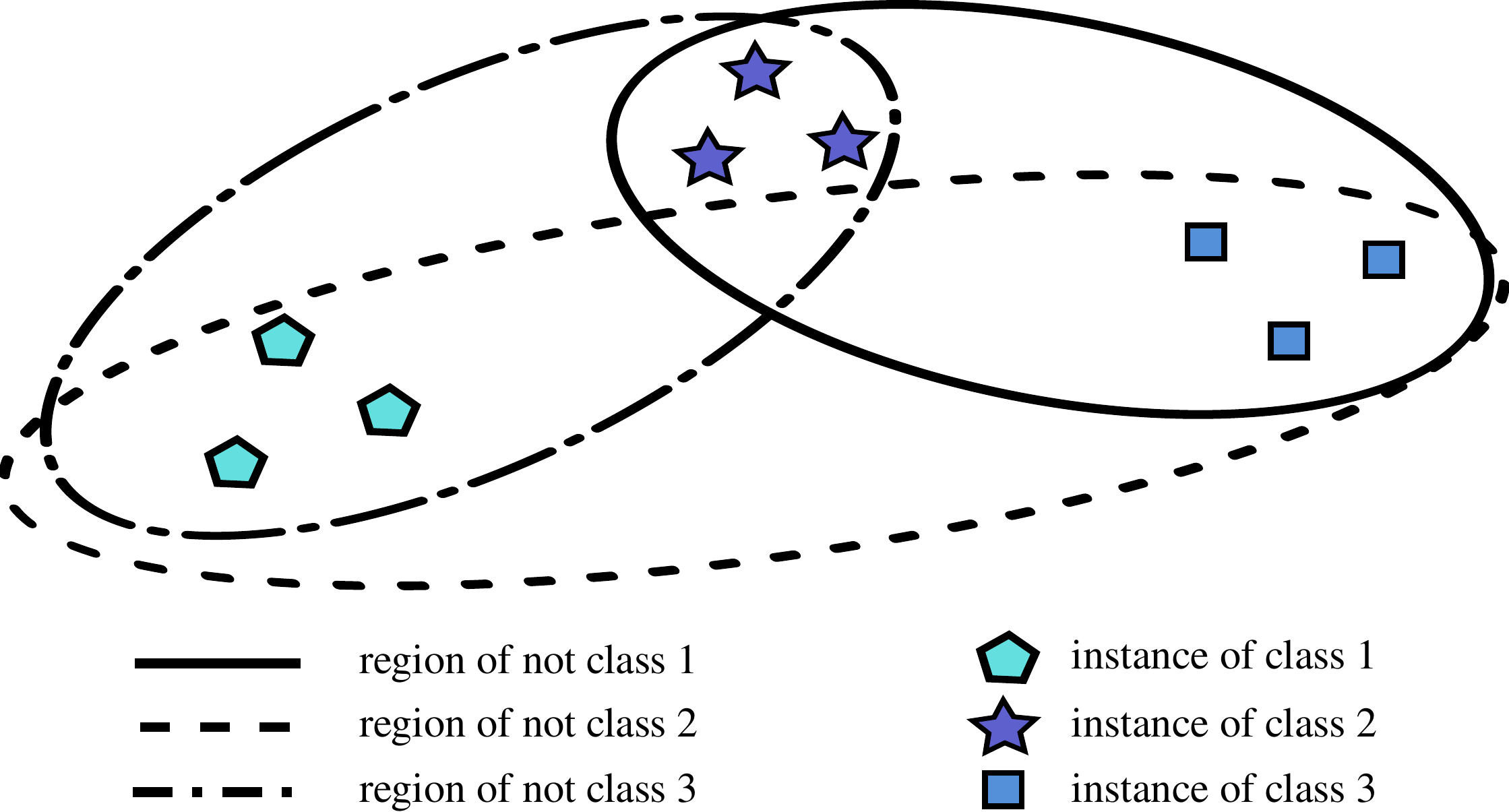}
	\caption{Illustration of the strategy of \protect\citetsupplement{supplementary-Cebron2012}: For each class, a one-class SVM is fitted to predict whether an instance does not belong to the respective class.}
	\label{fig:not-class}
\end{figure}

\textbf{Instance queries with partial label information:}
Classification problems with many classes often challenge human annotators because they have to pick one out of many classes as an instance's true class. 
This is, in particular, difficult if multiple classes may be in question for an instance.
Therefore, \citetsupplement{supplementary-Cebron2012} proposed a strategy overcoming such issues by facilitating instance queries.
Instead of asking for an instance's class label, it refers to partial label information by querying to which classes an instance does not belong.
Correspondingly, annotators are allowed to annotate an instance with a subset of class labels, i.e., $\Omega_Z \define \mathcal{P}({\Omega_Y})$.
This partial label information is used to create a set of instances for each class.
Such a set consists of instances that do not belong to the respective class.
As illustrated by Fig.~\ref{fig:not-class}, a one-class \textit{support vector machine} (SVM)~\citesupplement{supplementary-Tax2004} is trained for each set to define a region of instances not belonging to the respective class.
An instance's prediction is obtained by computing the distance to each of the $C$ regions and assigning the instance to the class with the highest distance.
The utility of an instance is estimated through the expected error of the classification model, i.e., the committee of the $C$ one-class SVMs.
Instances with high distances to all regions have high utilities because the classification model cannot exclude specific classes for an instance.
The strategy works for different numbers of excluded classes as annotation, e.g., an annotator can only exclude one or two classes in case of a classification problem with many classes.
In this context, \citeauthor{supplementary-Cebron2012} made the limiting assumption that each class has an equal probability of being excluded.
However, real human annotators could have certain preferences to exclude a class for an instance.

\textit{Active learning with partial feedback} (ALPF) proposed by \citetsupplement{supplementary-Hu2019} is another strategy querying partial label information about instances. 
ALPF assumes that the class labels can be organized into composite classes $\mathcal{K} \define \{\mathcal{K}_1, \dots, \mathcal{K}_O\}$, where each composite class is a subset of one or multiple classes, i.e., $\mathcal{K}_1, \dots, \mathcal{K}_O \subset \Omega_Y$.
These composite classes can be generated through an existing hierarchy of the class labels~$\Omega_Y$, e.g., when classifying animals, a composite class for dogs would contain all dog breeds in $\Omega_Y$.
These composite classes are then used in combination with an instance to formulate queries of the type: ``Does instance~$\mathbf{x}_n \in \mathcal{X}$ belong to the composite class $\mathcal{K}_o \in \mathcal{K}$?''.
Accordingly, the set of queries is defined through $\mathcal{Q}_\mathcal{X} \define \mathcal{X} \times \mathcal{K}$.
An annotator can either answer $\texttt{yes}$ or $\texttt{no}$ such that the annotation set is $\Omega_Z \define \{\texttt{yes}, \texttt{no}\}$.
% Given the query $(\mathbf{x}_n, \mathcal{K}_o)$ and an annotation $z \in \Omega_Z$, the set of possible labels for instance $\mathbf{x}_n$ is redefined as:
% \begin{equation}
% \end{equation}
ALPF's main idea is to use these \texttt{yes}/\texttt{no} queries to gradually reduce the set of potential classes to which an instance could belong.
A neural network~\citesupplement{supplementary-Jain1996} with an extended variant of the cross-entropy loss function is trained with this partial label information.
\citeauthor{supplementary-Hu2019} introduced three different utility measures to select queries, namely expected reduction in entropy, expected remaining classes, and expected decrease in classes.
The first measure can be seen as a variant of US for partial labels.
The second and third measures estimate how much an annotation would affect the set of potential classes an instance can belong to.
ALPF showed promising performance results on large-scale classification benchmark data sets.
Thus, it made a step toward annotating real-world data sets with a considerable number of classes, which cannot be surveyed in their entirety by a human annotator.

\textbf{Instance queries with self-assessments:}
For various classification problems, studies have shown that annotators can provide meaningful self-assessments in addition to a class label as an annotation for an instance~\citesupplement{supplementary-Calma2018a,supplementary-Wallace2011}. 

Allowing annotators to express their missing knowledge regarding an instance's annotation is a common approach to incorporate annotators' self-assessments~\citesupplement{supplementary-Kaeding2015,supplementary-Zhong2015,supplementary-Fang2013e,supplementary-Wallace2011,supplementary-Donmez2008b,supplementary-Donmez2010b}.
In this case, we can define the annotation set as ${\Omega_Z \define \Omega_Y \cup \{\texttt{unconfident}\}}$, where $\texttt{unconfident}$ represents that an annotator is not able or rejects to provide a class label as annotation.
\citetsupplement{supplementary-Wallace2011} proposed the strategy \textit{multiple expert active learning} (MEAL) differing between novices and experts as annotators.
MEAL queries experts only for instances that could not previously be assigned to any class with certainty by the novices.
In contrast, the strategies~\citesupplement{supplementary-Kaeding2015,supplementary-Zhong2015,supplementary-Fang2013e,supplementary-Donmez2008b,supplementary-Donmez2010b} use the instances annotated as \texttt{uncertain} to train their annotator models estimating the annotators' performances.
We provide a more detailed discussion on how they obtain these performance estimates in Appendix~\ref{appendix:annotator-performance-models}.

\citetsupplement{supplementary-Song2018} proposed a \acronym strategy with confidence-based answers for crowdsourcing annotation tasks.
It aims at aggregating the answers of the annotators of the crowd for creating highly accurate data sets at minimum AC.
For this purpose, an annotator is required to provide a confidence score $z \in \Omega_Z \define [-1, 1]$ as an instance's annotation.
Only binary classification tasks are included, such that a confidence score is transformed to the probability $(z + 1)/2$ for the positive class.
Relying on these probability estimates provided by multiple annotators for a single instance, a class label is aggregated by determining the maximum likelihood solution of a Beta distribution given the observed confidence scores. 
The inferred mean of this Beta distribution is an estimate of the true class posterior probability.
For the instance selection, US is applied in combination with the confidence intervals obtained by the fitted Beta distribution.
Hence, the strategy of \citeauthor{supplementary-Song2018} selects instances for which the decision of the classification model and the aggregated label are uncertain.

\citetsupplement{supplementary-Ni2012} proposed a similar strategy named \textit{best multiple oracles} (BMO). 
It aims at guaranteeing a user-defined threshold~$c_{\text{min}} \in \Omega_C \define [0.5, 1]$ for the correctness of the class labels used for training a classification model.
The main idea is to re-annotate an instance until the instance's class label certainty reaches the certainty threshold~$c_{\text{min}}$.
BMO selects instances according to a user-defined selection strategy, e.g., US or EER.
For annotating an instance, an annotator provides an estimated class label $y \in \Omega_Y \define \{1, 2\}$ and a confidence score~$c \in \Omega_C$ as additional feedback.
Such an annotation $(y, c) \in \Omega_Z \define \Omega_Y \times \Omega_C$ is assumed to describe that $y$ is an instance's correct class label with probability $c$.

The strategies of \citetsupplement{supplementary-Song2018} and \citetsupplement{supplementary-Ni2012} are limited to binary classification problems.
In contrast, \citetsupplement{supplementary-Calma2018a} proposed a strategy copying with multi-class problems.
Therefor, it expects a class label $y \in \Omega_Y$ and a corresponding confidence score $c \in \Omega_C \define [0, 1]$ as annotation.
Hence, the annotation set is given by ${\Omega_Z \define \Omega_Y \times \Omega_C}$.
Each annotation $(y, c)$ is transformed to a vector ${\mathbf{p} = (p_1, \dots, p_C)^\mathrm{T} \in [0, 1]^C}$ of class probabilities.
An element of this vector is computed according to
\begin{equation}
p_i \define \begin{cases} c + \frac{1 - c}{C} \text{ if } i \doteq y, \\ \frac{1 - c}{C} \text{ otherwise.} \end{cases}
\end{equation}
The obtained probability vectors are then used to train a generative classification model in combination with the 4DS strategy~\citesupplement{supplementary-Reitmaier2013} as a basis for the estimation of instances' utilities.

On the one hand, strategies asking for numerical confidence scores can improve the classification model's performance if these scores are well-calibrated.
On the other hand, confidence scores increase the annotation effort and can be in multi-annotator settings strongly biased.

\textbf{Instance queries with model predictions:}
Using the classification model predictions during the formulation of queries may ease and speed up the annotation process.
In this context, \citetsupplement{supplementary-Bhattacharya2019} proposed a strategy reducing the annotation effort per query.
Its idea is to preselect the set of possible class labels to which an instance may belong.
This set is defined through a user-defined number $O \in \{2, \dots, C-1\}$ of the most probable class labels predicted by the classification model.
Accordingly, a query is formulated as: ``To which class in $\{y^{(n_1)}, \dots, y^{(n_O)}\} \subset \Omega_Y$ does instance $\mathbf{x}_n \in \mathcal{X}$ belong?''.
%Accordingly, the set of potential queries is $\mathcal{Q}_\mathcal{X} \define \mathcal{X} \times \{\mathcal{Y}_H \mid \mathcal{Y}_H \subset \Omega_Y \wedge |\mathcal{Y}_H| = H\}$.
The utility of such a query considers only the probabilities of the $O$ most likely classes for instance $\mathbf{x}_n$.
For this purpose, a probability difference matrix $\mathbf{D}_n \in [0, 1]^{O \times O}$ with 
\begin{gather}
\mathbf{D}_n[i, j] \define \\ 
|\Pr(Y=y^{(n_i)} \mid X=\mathbf{x}_n, \boldsymbol{\theta}_{\mathcal{D}}) - \Pr(Y=y^{(n_j)} \mid X=\mathbf{x}_n, \boldsymbol{\theta}_{\mathcal{D}})| \nonumber
\end{gather}
is computed. 
Subsequently, the maximum eigenvalue of the inverse matrix $\mathbf{D}_n^{-1}$ represents the instance's utility.
This eigenvalue is inversely proportional to the average absolute difference between the estimated probabilities~\citesupplement{supplementary-Balasubramanian2009}.
As a result, instances whose top $O$ estimated class-membership probabilities are close to each other have high utilities.
During the entire annotation process, \citeauthor{supplementary-Bhattacharya2019} assume that an annotator always provides the true class label of an instance.
This assumption includes cases where an instance's true class label is not part of the preselected top $O$ likely classes.
However, in real-world applications, an annotator may be confused in such a case, in particular, if the annotator trusts the classification model's predictions.

\citetsupplement{supplementary-Biswas2013} also proposed a strategy directly incorporating the classification model's predictions into queries.
Therefor, it employs queries of the form:`` Does instance ${\mathbf{x}_n \in \mathcal{X}}$ belong to class ${y \in \Omega_Y}$? If this is not the case, can you explain the reason?''.
Accordingly, a query consists of a pair of instance and class label: $(\mathbf{x}_n, y) \in \Omega_X \define \mathcal{X} \times \Omega_Y$.
As annotation, a \texttt{yes} is expected if the instance $\mathbf{x}_n$ actually belongs to the class $y$.
Otherwise, the annotator is required to explain why this is not the case.
For example, an image of a city is given, and the \acronym strategy queries whether this image shows a forest.
The answer is not $\texttt{yes}$.
As a result, the annotator explains that this image does not belong to the class forest because it is not natural enough.
The term natural is a so-called relative attribute in this context.
We can interpret a relative attribute as a high-level feature to compare instances among each other.
Assuming that there are $r_1, \dots, r_O$ relative attributes, the set of possible explanations is defined as $\Omega_E \define \{r_1^\uparrow, \dots, r_O^\uparrow, r_1^\downarrow, \dots ,r_O^\downarrow\}$ with
$r_v^\uparrow/ r_v^\downarrow \in \Omega_E$ representing a  too high/low value for the relative attribute~$r_o$.
Together with the answer $\texttt{yes}$ the explanations form the annotation set: $\Omega_Z \define \{\texttt{yes}\} \cup \Omega_E$.
Given this interaction scheme, \citeauthor{supplementary-Biswas2013} answered three main questions.
\begin{enumerate}
	\item[(1)] How to learn from attribute-based explanations? If an annotator says that ``$\mathbf{x}_n$ is too $r_o$ to belong to class $y$``, i.e. $r_o^{\uparrow}$ as annotation, the AL strategy computes the strength of the attribute $r_o$ in the queried instance $\mathbf{x}_n$ as $r_o(\mathbf{x}_n)$. Then, the strategy identifies all non-annotated instances with an attribute strength about $r_o(\mathbf{x}_n)$. These instances can also not belong to class $y$ if $\mathbf{x}_n$ is too $r_o$ to belong to class $y$. Hence, the training data of the classification model is updated by adding these instances as counterexamples for class~$y$. This kind of label propagation works analogously for the case that the annotator provides $r_o^{\uparrow}$ as an annotation.
	\item[(2)] How to learn relative attribute models during the annotation process? A prerequisite for learning from relative attributes is the specification of the attribute predictors $r_1, \dots, r_O$.
	Such an attribute predictor is a function with $r_o(\mathbf{x}) \define \mathbf{w}_o^\mathrm{T}\mathbf{x}$. Given a ranking of instances regarding the $o$-th attribute, the weights $\mathbf{w}_o \in \mathbb{R}^D$ ideally satisfy $r_o(\mathbf{x}_n) > r_o(\mathbf{x}_m)$ if instance $\mathbf{x}_n$ has a stronger presence of attribute $r_o$ than instance $\mathbf{x}_m$.
	The attribute predictors can be either pre-trained~\citesupplement{supplementary-Parkash2012} or learned during the annotation process.
	In the latter case, the ranking of instances is intelligently built upon the relative feedback of the annotators.
	\item[(3)] How to consider relative feedback annotations during the query selection? Traditional US considers only class labels as annotations for an instance. 
	Therefore, \citeauthor{supplementary-Biswas2013} extended traditional US by accounting for the potential attribute-based feedback. The corresponding utility measure estimates the possible reduction of the entropy when annotating a query.
	For this purpose, it computes the expected entropy reduction over the different potential annotations in $\Omega_Z$. Finally, the query leading to the maximum entropy reduction is selected.
\end{enumerate}
The strategy of \citetsupplement{supplementary-Biswas2013} expects the annotators to provide explanations if an instance does not belong to a specific class.
In contrast, \citetsupplement{supplementary-Teso2019} proposed the strategy CAIPI where a query includes an instance's predictions with an explanation for an annotator.
As a result, a query takes the form: ``Does instance ${\mathbf{x}_n \in \mathcal{X}}$ belong to class ${y^{n_1} \in \Omega_Y}$ because of explanation $e_n \in \Omega_E$?''.
The set of possible queries can be formalized as $\mathcal{Q}_\mathcal{X} \define \mathcal{X} \times \Omega_Y \times \Omega_E$.
The instance to be queried is selected through a user-defined utility measure, e.g., US.
The class label is simply the selected instance's most probable class outputted by the classification model. 
The explanation is generated through the implementation of \textit{local interpretable model-agnostic explanations} (LIME)~\citesupplement{supplementary-Ribeiro2016}. 
Such an explanation is a collection of relevant components, e.g., words in a document or objects in an image, that mainly lead to the classification model's prediction.
The set of explanations can be represented through weights $(w_1, \dots, w_O)^\mathrm{T} \in \Omega_E \define \mathbb{R}^{O}$ for each of the $O \in \mathbb{N}$ component.
The magnitude $|w_o|$ indicates the overall contribution of the $o$-th component to the classification model's prediction. In contrast, the weight's sign indicates whether the component is a positive or negative indicator for the prediction.
Once a query has been selected, an annotator can interact in three different ways:
\begin{enumerate}
	\item[(1)] The annotator confirms the query if the prediction and explanation are correct.
	\item[(2)] The annotator contradicts the query if the prediction is wrong and provides the instance's true class label.
	\item[(3)] The annotator corrects the explanation if the prediction is correct while the explanation is wrong.
\end{enumerate}
These three different interaction possibilities can be summarized through the annotation set $\Omega_Z \define \{\texttt{yes}\} \cup \Omega_Y \cup \Omega_E$.
The third case is novel in AL. 
CAIPI incorporates the corrected explanation by generating synthetic instances that teach the classification model to identify irrelevant components.
\citeauthor{supplementary-Teso2019} empirically showed  that CAIPI could increase the annotators' trust in the classification model and that corrected explanations can enormously improve the classification model's performance.
A disadvantage of the CAIPI is its need for appropriate component definitions regarding the classification problem at hand.

\subsection{Region Queries}
\label{subappendix:region_queries}
Real-world AL strategies may use region queries to capture high-level information about a classification task.
The main challenge concerns the generation of human-understandable region queries whose annotations enhance the performance of a classification model.
This challenge is similar to membership query synthesis~\citesupplement{supplementary-Angluin1988}, where the class label of any instance in the feature space $\Omega_X$ can be queried.
In this setting, Baum and Lang~\citesupplement{supplementary-Baum1992} encountered the problem that many synthetic instances had no natural semantic meaning and were thus difficult to annotate by human annotators.   
A further challenge of AL with region queries is the number of regions in a feature space.
In the case of numerical features, there are infinitely many regions.
Hence, existing region query utility measures take only a finite subset of possible regions into account.
We discuss the main approaches for limiting the number of queries and measuring their utilities in the following.

\textbf{Region queries for natural language processing:}
In natural language processing~\citesupplement{supplementary-Nadkarni2011} tasks, documents as instances are often described by many features.
Therefore, the sole use of instance queries may result in poor classification performance~\citesupplement{supplementary-Druck2009}. As an alternative, specific features can be queried~\citesupplement{supplementary-Cakmak2012,supplementary-Druck2008}, such as the correlation between a feature and a class. For example, in the classification task distinguishing \texttt{hockey} from \texttt{baseball} related text documents, the presence of the word \texttt{puck} is a strong indicator for the class \texttt{hockey}~\citesupplement{supplementary-Druck2009}.
Assuming the feature value $x_{nd} \geq 0$ is given by the frequency of the word \texttt{puck} indexed by $d$ in the document instance $\mathbf{x}_n$, the exemplary region query is formalized by $q_{d} \define X_\texttt{d} > 0$ with its annotation~$z_d \define \texttt{hockey}$. As a result, all documents containing the word \texttt{puck} are annotated with the class $\texttt{hockey}$. 
A feature can also be an indicator for multiple classes~\citesupplement{supplementary-Druck2009,supplementary-Druck2008}, e.g., the presence of the word \texttt{player} may indicate a \texttt{baseball} or \texttt{hockey} related document.
\citetsupplement{supplementary-Druck2009} proposed the strategy GE WU which selects a query $X_{d} > 0 \in \mathcal{Q}_\mathcal{X} \define \{X_1 > 0, \dots, X_D > 0\}$ for annotation through a weighted US variant. The weighting is to balance the trade-off between very frequent and infrequent features (i.e., words).

The strategy DUALIST~\citesupplement{supplementary-Settles2011a} uses the same form of region queries.
However, it additionally combines them with instance queries.
In each learning cycle, DUALIST selects a fixed number of the most uncertain instances (i.e., documents) employing the entropy-based US (cf. Eq.~\ref{eq:us} in the survey).
Moreover, the top informative region queries are presented to an annotator.
Their utilities are defined as information gains that their annotations would provide:
\begin{equation}
\begin{gathered}
\phi_{\text{DUALIST}}(X_d > 0 \mid \boldsymbol{\theta}_{\mathcal{D}}) \define \sum_{\delta_d = 0}^1 \sum_{y \in \Omega_Y} \Pr(\delta_d, Y=y \mid \boldsymbol{\theta}_{\mathcal{D}}) \\ \log\left(\frac{\Pr(\delta_d, Y=y \mid \boldsymbol{\theta}_{\mathcal{D}})}{\Pr(\delta_d \mid \boldsymbol{\theta}_{\mathcal{D}})\Pr(Y=y \mid \boldsymbol{\theta}_{\mathcal{D}})}\right)
\end{gathered}
\end{equation}
with $\delta_d$ indicating the presence or absence of a feature (i.e., word) in an instance (i.e., document).
This measure is inspired by a common feature selection method specifying the most salient features in text classification~\citesupplement{supplementary-Sebastiani2002}.
To speed up the annotation process, DUALIST organizes the selected region queries into classes.
Therefor, the query $X_d > 0$ is associated with the class with which it occurs most frequently and any other class with which it appears at least 75\% often.

\textbf{Asking generalized questions:}
Region queries of the form $X_d > 0$  consider only a single feature and may be overly general.
As a result, an annotator may not provide a meaningful annotation.
Although the AL strategies GE WU and DUALIST allow an annotator to ignore a region query, the AL strategy \textit{asking generalized queries} (AGQ)~\citesupplement{supplementary-Du2009} goes further. It constructs queries with an adaptive degree of specificity.
For this purpose, the most uncertain instance $\mathbf{x}_{n^*}$ is selected according to US in the first step (cf. Eqs.~\ref{eq:selection},~\ref{eq:us} in the survey).
Instead of asking for the annotation of the specific instance~$\mathbf{x}_{n^*}$, irrelevant features are identified and removed from its feature vector in the second step. 
The resulting query is formalized by $q_{n^*}~\define~\bigwedge_{d \in \mathcal{R}_{n^*}} (X_d \doteq x_{n^*d})$, where $\mathcal{R}_{n^*} \subseteq \{1, \dots, D\}$ denotes the index set of estimated relevant features regarding instance~$\mathbf{x}_{n^*}$.
To determine those features, the set $\mathcal{R}_{n^*}$ (initially containing all feature indices $\mathcal{R}_{n^*} \define \{1, \dots, D\}$) is gradually reduced by removing the indices of irrelevant features.
It is assumed that the values of irrelevant features have a negligible impact on the class membership probabilities estimated by the classification model.
Based on this assumption and given the current index set $\mathcal{R}_{n^*}$ of assumed relevant features, for each feature $X_d$ with $d \in \mathcal{R}_{n^*}$, synthetic instances are generated to test how the probabilistic estimates of the classification model are affected when the feature values for the feature $X_d$ are varied.
The feature $d^*$ leading to the smallest change in the probabilistic estimates is selected as irrelevant: $\mathcal{R}_{n^*} \leftarrow \mathcal{R}_{n^*} \setminus \{d^*\}$.
This procedure is repeated until the change in the probabilistic estimates for each feature~$X_d$ with $d \in \mathcal{R}_n$ is below a user-defined threshold. 

Region queries generated by that AL strategy AGQ are still restricted in their form, since a feature $X_d$ is either entirely irrelevant or set to a specific value ${X_d \doteq x_{{n^*}d}}$.
In particular, these queries are overly specific when dealing with numerical features.
The AL strategy $\text{AGQ}^{+}$~\citesupplement{supplementary-Du2010a}, being an extension of AGQ, resolves this problem.
After the indices $\mathcal{R}_{n^*}$ of the relevant features have been identified according to AGQ, the robustness of the estimated class membership probabilities regarding the value ranges of each of the relevant features $X_d$ with $d \in \mathcal{R}_{n^*}$ is tested.
Therefor, a set of values $\mathcal{X}_{{n^*}d}$ is assigned to each relevant nominal feature $X_d$ with $d\in \mathcal{R}_{n^*}^{\text{nom}} \subseteq \mathcal{R}_n$, whereas an interval $[x_{{n^*}d}^{\text{min}}, x_{{n^*}d}^{\text{max}}]$ is assigned to each numerical feature $X_d$ with $d \in \mathcal{R}_{n^*}^{\text{num}} \subseteq \mathcal{R}_{n^*}$.
The corresponding query takes the form 
\begin{equation}
\label{eq:disciminative_feature_rule}
q_{n^*} \define \bigwedge_{d \in \mathcal{R}_{n^*}^{\text{nom}}} (X_d \in \mathcal{X}_{{n^*}d})  \bigwedge_{d \in \mathcal{R}_{n^*}^{\text{num}}} (X_d \in [x_{{n^*}d}^{\text{min}}, x_{{n^*}d}^{\text{max}}]).
\end{equation}
Since a region query generated by $\text{AGQ}$ or $\text{AGQ}^+$ may cover regions with instances of varying classes, so-called proportion labels are expected as annotations.
In a binary classification problem, the proportion label of a region is a number between zero and one: $\Omega_{Z} \define [0, 1]$.
It informs about the proportion of positive instances in the region.
There are a variety of methods to train classification models using proportion labels.
The corresponding research area is called learning with label proportions~\citesupplement{supplementary-Stolpe2011,supplementary-Yu2013}.

\citetsupplement{supplementary-Haque2013} proposed \textit{generalized query-based active learning}~(GQAL) as another strategy for region queries. It is closely related to $\text{ACQ}^+$.
The main difference is that GQAL supports multi-class classification problems whereas $\text{AGQ}^+$ is restricted to a binary classification setting.

\textbf{Rule induced active learning query:} 
As pointed out in~\citesupplement{supplementary-Rashidi2011}, a drawback of AGQ and $\text{AGQ}^{+}$ is the use of synthetic instances drawn from an estimated distribution.
If the sampled instances do not reflect the actual distribution, the inferred query fails at defining a critical region within the feature space.
For this reason, the AL strategy \textit{rule-induced active learning query} (RIQY)~\citesupplement{supplementary-Rashidi2011} takes only the observed instances into account.
Following the AL strategy $\text{AGQ}^{+}$, the instance~$\mathbf{x}_{n^*}$ with maximum utility is selected.
However, instead of applying exclusively $\phi_{\text{US}}$ (cf. Eq.~\ref{eq:us} in the survey) as a utility measure, RIQY combines the uncertainty regarding an instance's class membership  with its density and its dissimilarity to previously selected instances.
Thus, this utility measure selects less redundant and more representative instances.
Subsequently, a set of similar non-annotated instances (neighbors) $\mathcal{N}_{\mathbf{x}_{n^*}} \subset \mathcal{U}$ and a set of dissimilar instances (enemies) $\mathcal{E}_{\mathbf{x}_{n^*}} \subset \mathcal{U}$ are specified with reference to the selected instance $\mathbf{x}_{n^*}$.
They form the training set of a rule induction classifier such as C4.5 decision tree~\citesupplement{supplementary-Quinlan2014}, whose task is to generate classification rules separating the instances in the neighbor set~$\mathcal{N}_{\mathbf{x}_{n^*}}$ from instances in the enemy set $\mathcal{E}_{\mathbf{x}_{n^*}}$.
The learned rules define regions around instance $\mathbf{x}_{n^*}$ and represent the possible queries having the same form like the one shown in Eq.~\ref{eq:disciminative_feature_rule}. 
% Such a query can be formalized by
% \begin{align}
% 	q_n = &\bigwedge_{d \in \mathcal{R}_n^{\text{nom}}} (X_d \in \mathcal{X}_{nd})  \bigwedge_{d \in \mathcal{R}_n^{\text{num,min}}} (X_d \geq x_{nd}^{\text{min}})\nonumber\\ &\bigwedge_{d \in \mathcal{R}_n^{\text{num,max}}} (X_d \leq x_{nd}^{\text{max}}),
% \end{align}
% where the sets $\mathcal{R}_n^{\text{nom}}, \mathcal{R}_n^{\text{num,min}}, \mathcal{R}_n^{\text{num,max}} \subseteq \{1\dots,D\}$ are determined by the decision tree.
Since a rule-based classifier generates multiple rules, a rule selection is performed based on the rule's accuracy and coverage.
The rule's accuracy is a proxy for its discriminative power, and its coverage describes how many observed instances are located in the region defined by the rule.
The rules with accuracy above a minimum threshold are ranked according to their coverage scores, and the top ones are selected as queries for annotation with proportion labels.

\textbf{Hierachical region queries:}
Since RIQY and $\text{AGQ}^+$ extend the standard US by defining a region around the most uncertain instance, both AL strategies reduce the problem of utility estimation for a region query on utility estimation for an instance query.
This way, the quality of the resulting region is not directly considered during the estimation of the query utility. 
Hence, the constructed region may not be meaningful.
In this context, the term meaningful describes two points of view~\citesupplement{supplementary-Luo2019}. 
\begin{enumerate}
	\item[(1)] From a human annotator's view, the region represents a reasonable and interpretable population of possible instances.
	\item[(2)] Whereas from the classification model's viewpoint, the region covers a large part of the feature space while being pure in the sense that the vast majority of instances in this region belong to the same class.
\end{enumerate}
Several AL strategies aim at defining such regions in a hierarchical manner, namely \textit{hierarchical active learning with group proportion feedback} (HALG)~\citesupplement{supplementary-Hauskrecht2018}, \textit{hierarchical active learning with proportion feedback on regions} (HALR)~\citesupplement{supplementary-Luo2018}, and \textit{adaptive hierarchical active learning with proportion feedback on regions} ($\text{A}^*$HALR)~\citesupplement{supplementary-Luo2019}. 
The AL strategy $\text{A}^*$HALR is an advancement of the other two AL strategies, and we summarize its main idea in the following.

\begin{figure}[h!]
	\centering
	\includegraphics[width=\columnwidth]{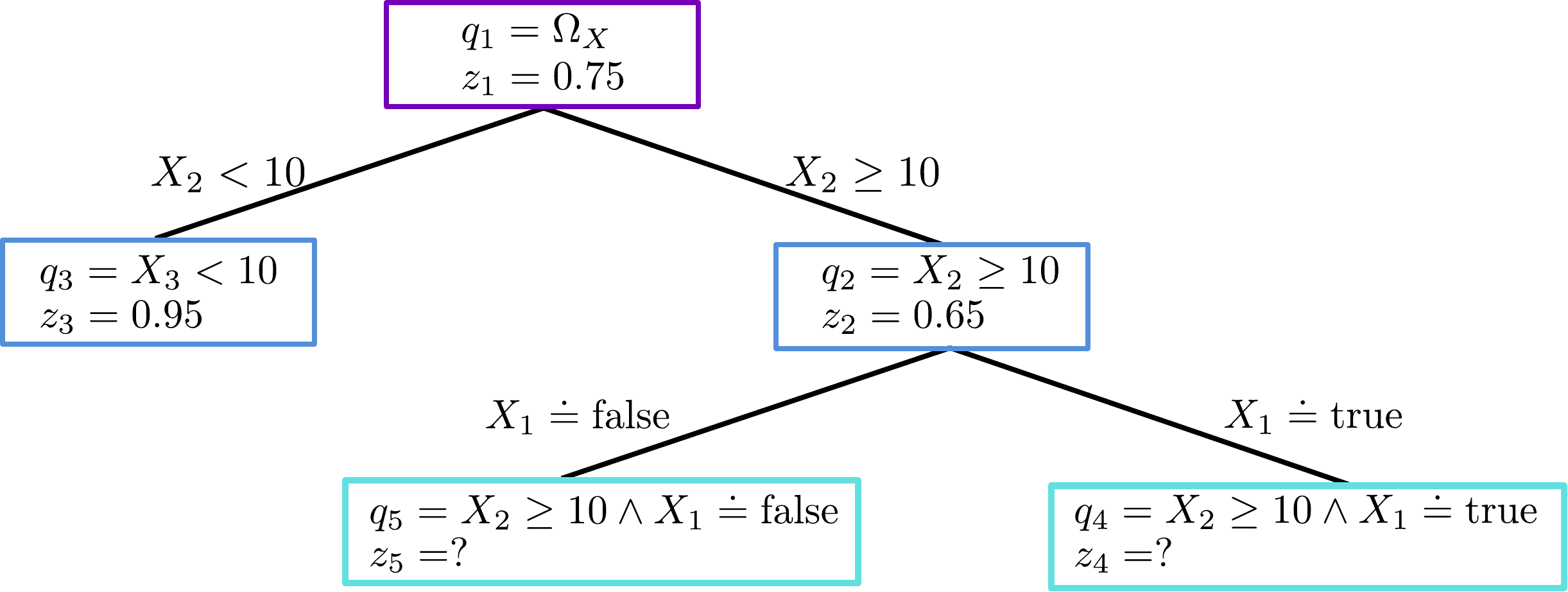}
	\caption{Illustration of a binary hierarchical tree of region queries: The nodes represent region query and annotation pairs. The edges are value constraints defining the division into sub-regions.}
	\label{fig:hierachical_tree}
\end{figure}

It constructs a binary hierarchical tree. 
Each node of this tree represents one region for which an annotator can be queried to provide a proportion label. 
An exemplary tree is illustrated in Fig~\ref{fig:hierachical_tree}.
The region of the root node covers the entire feature space $\Omega_{X}$, and the first query $q_1 \define \Omega_{X}$ asks for the proportion label of this feature space.
The obtained proportion label $z_1 \in \Omega_Z \define [0, 1]$ can be interpreted as the prior probability for the positive class. In the example of Fig.~\ref{fig:hierachical_tree}, the proportion label $z_1 \define 0.75$ indicates $75\%$ positive instances and $25\%$ negative instances.
Subsequently, the root node is divided into two sub-regions through value constraints on one of the features $X_1, \dots, X_D$.
For example, in Fig.~\ref{fig:hierachical_tree}, two sub-regions are defined through the value constraints $X_2 < 10$ and $X_2 \geq 10$.
Such a split for the region of a query $q_i$ is defined according to the standard decision tree splitting based on information gain.
However, $\text{A}^*$HALR has no access to the class label of each instance in $\mathcal{G}_i \subseteq \mathcal{X}$ denoting the groups of observed instances contained by the region of query~$q_i$.
For this reason, it relies on two heuristics generating instance-level annotations as proxies of the class labels.
On the one hand, an unsupervised heuristic can be applied.
In this case, a \textit{Gaussian mixture model} (GMM)~\citesupplement{supplementary-Bishop2006} divides the instances~$\mathcal{G}_i$ into two clusters. 
The computed cluster membership probabilities for one of the two clusters, also known as responsibilities, are used as annotations.
On the other hand, a supervised heuristic can be used.
For this purpose, the positive class membership probabilities predicted by the current classifier for all instances in the group $\mathcal{G}_i$ are used as annotations to compute a split. 
As only one of both heuristics can be applied as the splitting criterion, $\text{A}^*$HALR formulates the selection problem as a two-arm bandit problem~\citesupplement{supplementary-Besbes2014}. 
It adapts its policy for the heuristic selection during the entire annotation process.

After each split, the proportion label for one of the two resulting region queries is requested.
The example in Fig.~\ref{fig:hierachical_tree} shows that an annotator provided the proportion label $z_2 \define 0.65$ for the query $q_2$.
Based on this information and in combination with the proportion label $z_1$ as annotation of the query $q_1$, the proportion label $z_3$ of the query $q_3$ can be inferred.
Generally speaking, let $\mathcal{G}_i \define \mathcal{G}_j \cupdot \mathcal{G}_k \subseteq \mathcal{X}$ define the partitioning of the instance group of query $q_i$ into the instance groups of query $q_j$ and $q_k$. Moreover, the annotations $z_i, z_j$ are known.
Then the proportion label for query $q_k$ is inferred through
\begin{equation}
z_k \define \frac{|\mathcal{G}_{i}| \cdot z_{i} - |\mathcal{G}_{j}| \cdot z_{j}}{|\mathcal{G}_{k}|}.
\end{equation}
Next to the definition of a split, the selection of the region to be split is essential.
The potential regions that can be split are given by the leaves of the current hierarchical tree of region queries.
After the initial split of query $q_1$ and the annotation of query $q_2$ and $q_3$, the region represented by query $q_2$ was selected for splitting in Fig.~\ref{fig:hierachical_tree}.
But, one could have also split the region represented by query $q_3$.
To decide which region is to be split, $\text{A}^*$HALR computes a region's uncertainty as a proxy of its utility measurement.
The uncertainty of a region considers two factors: the label impurity and the number of observed instances enclosed by a region.
These factors are combined into a kind of Gini-index being a product of the region's proportions of positive and negative instances and its number of instances:
\begin{equation}
\phi_\text{A*HALR}(q_i \mid \boldsymbol{\theta}_{\mathcal{D}}) \define z_i (1 - z_i) |\mathcal{G}_i|. 
\end{equation}
The steps of region selection, splitting, and annotation are executed consecutively in each AL cycle.

\subsection{Comparison Queries}
\label{subappendix:comparison_queries}
Comparison queries ask for relative information between multiple instances.
The main task concerns the selection of the instances to be compared.
For this purpose, a utility measure assesses possible comparison queries.
In the following, we discuss the main approaches for measuring the utility of this type of query.

\textbf{Class-based comparisons of instances:}
In classification problems with many classes, annotators require much time to assign an instance to a class. 
In contrast, binary annotations, i.e., ${\Omega_Z \define \{\texttt{yes}, \texttt{no}\}}$ answers, are less time-consuming~\citesupplement{supplementary-Joshi2010,supplementary-Joshi2012} and less error-prone~\citesupplement{supplementary-Fu2014}.
Therefore, several AL strategies aim at reducing the annotation effort and the number of mistakes by querying whether two instances belong to the same class.

\citetsupplement{supplementary-Joshi2010,supplementary-Joshi2012} proposed one of these AL strategies.
Its main idea is to select a non-annotated instance and compare it with an annotated instance whose class label is already known.
If an annotator confirms that both instances belong to the same class, the non-annotated instance is added to the set of annotated instances.
Otherwise, another annotated instance is selected for comparison with the non-annotated instance.
This process is iterated until a match is found.
Accordingly, the set of potential queries is given through $\mathcal{Q}_{\mathcal{X}} \define \{\{\mathbf{x}, \mathbf{x}^\prime\} \mid \mathbf{x} \in \mathcal{U} \wedge \exists y \in \Omega_y: \, (\mathbf{x}^\prime, y) \in \mathcal{L}\}$.
Their utilities are measured separately to select a pair of instances.
The utility of a non-annotated instance is computed as the difference between the negative expected MC (estimated by a cost-sensitive variant of EER) and the expected AC (estimated by the expected number of comparisons to obtain a match). 
Once the non-annotated instance $\mathbf{x}_{n^*}$ with maximum utility has been selected, its class membership probability estimates are sorted.
In the first comparison, the annotator has to decide whether the instance $\mathbf{x}_{n^{*}}$ and another randomly picked instance of the class with the highest probability belong to the same class. 
If they do not match, $\mathbf{x}_{n^{*}}$ is compared to another randomly picked instance of the class with the second-highest probability and so on.
Hence, we can interpret the class membership probability estimates of $\mathbf{x}_{n^{*}}$ as utilities for selecting one of the already annotated instances.

\begin{figure}[!h]
	\centering
	\includegraphics[width=\columnwidth]{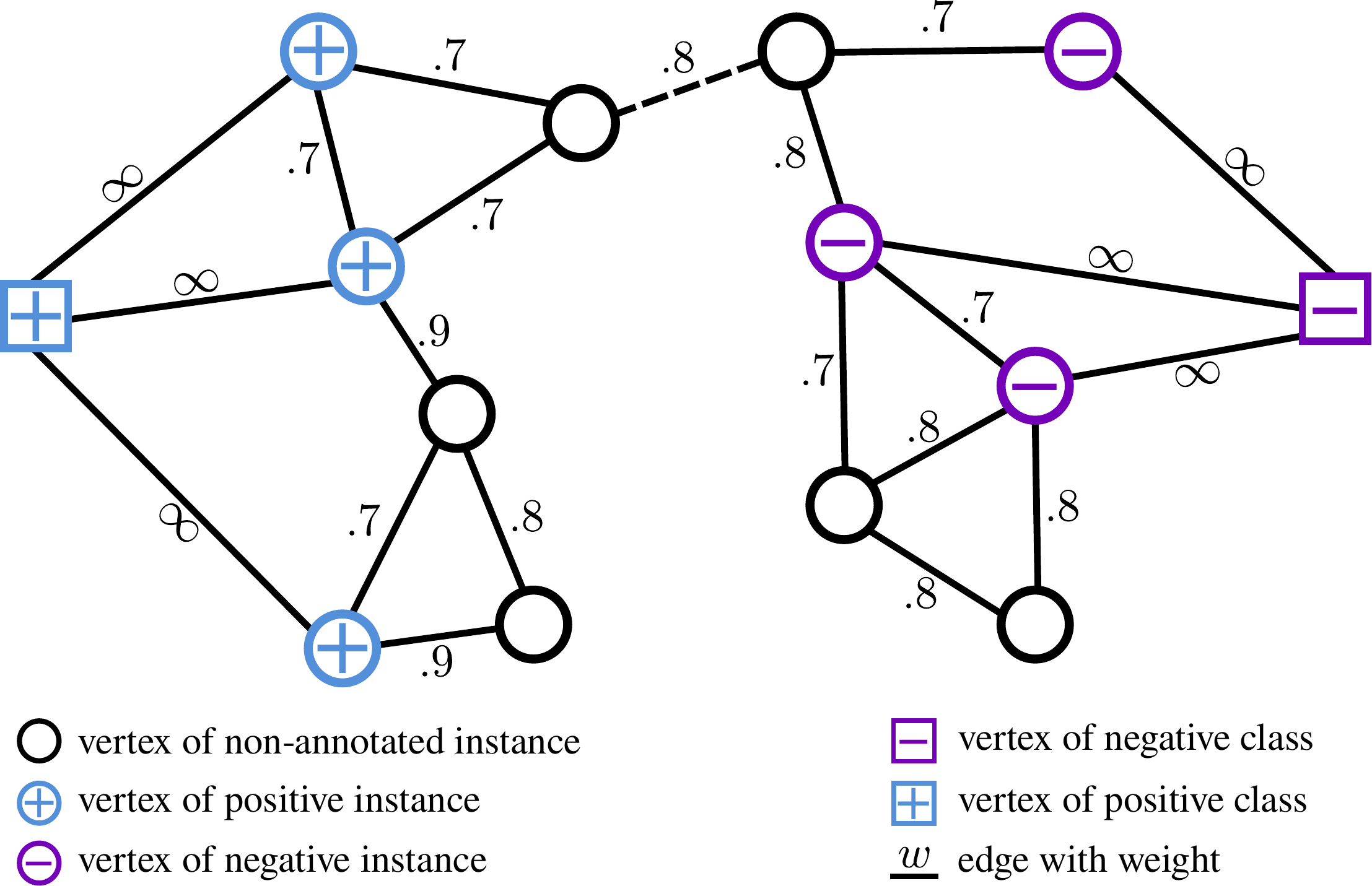}
	\caption{Illustration of the graph min-cut learning algorithm based on a ${(k=3)}$-NN: In such a graph, the vertices are defined through the observed instances and one extra vertex for each class. An edge exists between the vertices of two instances if one instance belongs to the other instance's top $k$-NN and vice versa. The weight of an edge corresponds to the similarity between instances. Depending on the assigned class label, the vertex of an annotated instance is additionally connected to one of the classification vertices with an infinite large weight.  The dashed edge marks the minimum cut to assign instances either to the positive class (left side of the dashed edge) or the negative class (right side of the dashed edge).}
	\label{fig:phal}
\end{figure}

In summary, the strategy of \citetsupplement{supplementary-Joshi2010,supplementary-Joshi2012} obtains class labels by comparing non-annotated to annotated instances.
A single comparison reveals only new class information regarding the non-annotated instance.
In contrast, the AL strategies \textit{querying pairwise label homogeneity active learning} (QHAL)~\citesupplement{supplementary-Fu2011} and its advancement \textit{pairwise homogeneity based active learning} (PHAL)~\citesupplement{supplementary-Fu2014} aim at increasing the information content by asking whether two non-annotated instances belong to the same class.
Thus, the query set is given by $\mathcal{Q}_{\mathcal{X}} \define \{\{\mathbf{x}, \mathbf{x}^\prime\} \mid \mathbf{x}, \mathbf{x}^\prime \in \mathcal{U} \wedge \mathbf{x} \neq \mathbf{x}^\prime\}$.
At the start of the annotation process, an ensemble with the semi-supervised graph min-cut learning algorithm~\citesupplement{supplementary-Blum2001} as the base learner is constructed. 
The ensemble's classification models create so-called $k$-NN graphs from the observed instances for different values of $k$ (cf. Fig.~\ref{fig:phal}).
Each ensemble member determines a classification of the instances by partitioning its graph. 
It minimizes the number of similar pairs of instances belonging to different classes, i.e., partitions.
The idea of PHAL is to repeatedly refine the edge weights of these graphs by querying the label homogeneity information regarding two non-annotated instances. 
If both instances belong to the same class, the edge weight is increased. 
Otherwise, it is decreased.  
PHAL selects only pairs of non-annotated instances lying on the max-flow path of a graph in the ensemble. 
These pairs are more effective (i.e., have a higher utility) in reducing the upper bound of the graph min-cut learning algorithm's misclassification error than a random selection.
At the end of each learning cycle, a predefined number of non-annotated instances with the ensemble's highest prediction confidence are added to the set of annotated instances.

PHAL has two  drawbacks on the one hand.
First, it considers only binary classification problems. 
Second, it has high computational complexity due to the costly execution of graph min-cut learning algorithm. 
On the other hand, PHAL is less sensitive to incorrect annotations than traditional AL strategies because erroneous changes of edge weights do not significantly affect the overall model.

\citetsupplement{supplementary-Kane2017} developed a rather theoretical strategy.  
Next to instance queries asking for class labels, it employs comparison queries of the form: ``Is instance $\mathbf{x}_n$ more likely to belong to the positive/negative class than instance $\mathbf{x}_n$?''.
The corresponding query set is $\mathcal{Q}_\mathcal{X} \define \mathcal{X} \cup (\mathcal{X} \times \mathcal{X})$.
The strategy focuses on the learning of halfspaces~\citesupplement{supplementary-Guruswami2009}.
For this class of classification problems, including certain assumptions, the authors prove that the additional use of comparison queries leads to an exponential improvement over traditional AL strategies. 
Hence, approximately $\mathcal{O}(\log N)$ queries are sufficient to reveal all class labels of the $N$~observed instances in~$\mathcal{X}$.
A similar setting is targeted by \citetsupplement{supplementary-Xu2017} and \citetsupplement{supplementary-Hopkins2020}.
For annotating comparison and instance queries, they also allow certain noise types, e.g., Tsybakov~\citesupplement{supplementary-Tsybakov2004} or Massart~\citesupplement{supplementary-Massart2006} noise.
The theoretical analyses of these strategies point out the benefit of comparison queries since they can substantially reduce the query complexity.

\textbf{Similarity-based comparisons of instances:}
In complex tasks, such as predicting a patient's risk for suffering from a particular disease, it is difficult to confidently provide a class label, even for medical experts~\citesupplement{supplementary-Bridge2016}. Therefore, the strategy \textit{active patient risk prediction} (ARP)~\citesupplement{supplementary-Qian2015} resorts to relative comparisons between patients. For example, given an image created with \textit{magnetic resonance imaging} (MRI), ARP does not ask whether the corresponding patient will get a particular disease but requests a similarity-based comparison to MRI images of other patients. An exemplary query would be:
``Is patient A more similar to patient B than patient C?''. ARP aims to use the resulting similarity information for updating a similarity matrix $\mathbf{S} \in \mathbb{R}_{\geq0}^{N \times N}$ between the patient's medical records. These records represent the observed instances $\mathcal{X}$ from an ML perspective. Since the number of theoretical comparisons would grow exponentially with the number of instances, ARP restricts the query set to $\mathcal{Q}_\mathcal{X} \define \{\mathcal{N}_1, \dots, \mathcal{N}_N\}$ where $\mathcal{N}_n \subset \mathcal{X}$ is the neighborhood set of the $n$-th instance. These neighborhood sets can be defined according to the $k$-NN algorithm, for example. An annotation of a selected neighborhood set $\mathcal{N}_{n^*}$ induces an ordering of the contained instances regarding their similarities to instance $\mathbf{x}_{n^*}$. Such annotation is used to update the similarity matrix~$\mathbf{S}$. Therefor, the linear neighborhood propagation framework~\citesupplement{supplementary-Wang2008} is employed. Its idea is to learn the similarity matrix $\mathbf{S}$ minimizing the reconstruction error:
\begin{equation*}
\sum_{n=1}^N \biggl|\biggl|\mathbf{x}_n - \sum_{\mathbf{x}_{m} \in \mathcal{G}_n} \mathbf{S}[n,m] \mathbf{x}_m \biggr|\biggr|^2.
\end{equation*}
When solving this optimization problem, the obtained annotations are incorporated as relative constraints, e.g., $\mathbf{S}[n,m] \geq \mathbf{S}[n,l]$ is required, if instance $\mathbf{x}_m$ is more similar to instance $\mathbf{x}_n$ than instance $\mathbf{x}_l$. The annotation costs are reduced by selecting the neighborhood set $\mathcal{G}_{n^*}$ with the highest utility. This utility is estimated by solving a counting set cover problem~\citesupplement{supplementary-Gionis2012} greedily. By doing so, ARP aims at determining a minimum set of neighborhood sets whose union covers all observed instances. Based on an initial subset of instances (patients) with already assigned risk predictions, the learned similarity matrix can be used to estimate the risk for the remaining instances (patients).
\citesupplement{supplementary-Qian2015} experimentally showed that ARP performs better than randomly selecting neighborhood sets for annotation and is even competitive to instance queries asking for class labels (e.g., absolute risk estimates). However, it is questionable how the neighborhood sets' initial definition affects ARP's performance.

\citetsupplement{supplementary-Xiong2015} proposed a related strategy to ARP. It queries relative similarities among samples. However, instead of neighborhood sets, it assumes triples of instances as queries: ${q_{lmn} \define (\mathbf{x}_l, \mathbf{x}_m, \mathbf{x}_n) \in \mathcal{Q}_\mathcal{X} \define \mathcal{X}^3}$. We can interpret such a triplet as the question ``Is $\mathbf{x}_l$ more similar to $\mathbf{x}_m$ than $\mathbf{x}_n$?''. An annotator returns an answer ${z_{lmn} \in \Omega_Z = \{\texttt{yes}, \texttt{no}, \texttt{uncertain}\}}$ as annotation. Generally, it is assumed that the annotation depends only on the true class labels of the triplet's instances:
\begin{equation}
z_{lmn} \define \begin{cases}\texttt{yes} \text{ if } y_l \doteq y_m \not\doteq y_n, \\ \texttt{no}  \text{ if } y_l \not\doteq y_m \doteq y_n, \\ \texttt{uncertain} \text{ otherwise.} \end{cases} 
\end{equation}
The utility of an instance triplet is estimated in a probabilistic manner.
More specifically, the strategy employs the mutual information criterion~\citesupplement{supplementary-Cover1991}.
% :
% \begin{equation}
%     \phi_{\text{AL-RC}}(q_{lmn}) = H[\mathbf{Y}_{lmn} \mid \mathcal{D}(t)] - H[\mathbf{Y}_{lmn} \mid \mathcal{D}(t), Z_{lmn}],
% \end{equation}
% where $\mathbf{Y}_{lmn} = (Y_l, Y_m, Y_n)$ denotes the random variables of the three instance's true class labels and $Z_{lmn}$ the random variable of the annotation.
Accordingly, the utility of a query $q_{lmn}$, i.e., instance triplet, is the degree upon which the query's annotation reduces the uncertainty of the three instance's unknown class labels $y_l, y_m, y_n$.
Since there are $|\mathcal{Q}_\mathcal{X}| = |\mathcal{X}^3| = N^3$ possible queries, the utility computation is infeasible for a large number $N$ of observed instances.
In this case, the strategy randomly samples a subset of queries and selects the query with the highest utility in this set.
The annotated queries in $\mathcal{D}$ are used to learn a distance metric~\citesupplement{supplementary-Schultz2004} for the observed instances $\mathcal{X}$.
Subsequently, the instances are partitioned into clusters, e.g., using the $k$-means clustering algorithm~\citesupplement{supplementary-Xu2005}.
The obtained cluster assignments are interpreted as proxies of the class labels and serve as training data for an arbitrary classification model.
One major issue of this strategy is the information loss if an annotator provides \texttt{uncertain} as an annotation. 
Then, the annotation is not used by the distance learning algorithm, and the annotation effort is wasted.

\section{Annotator Performance Models}
\label{appendix:annotator-performance-models}
In this appendix, we analyze concrete \acronym strategies regarding their annotator performance models.
We structure this analysis according to the annotator performance  types, i.e., uniform, annotation-dependent, and query-dependent performance, identified in Section~\ref{sec:types_of_queries_and_annotations} in the associated survey.
To the best of our knowledge, each of these models assumes instance queries such that we define $\mathcal{Q}_\mathcal{X} \define \mathcal{X}$ as the query set for this appendix.
The interpretation of the annotators' performance random variables $P_1, \dots, P_M$ depends on the concrete models.
Without further specification we assume that these variables are binary where $P_m = 1$ indicates an optimal (correct) and $P_m = 0$ a non-optimal (false) annotation.

\subsection{Uniform Annotator Performance}
\label{subappendix:uniform-annotator-performance}
Uniform annotator performance means that the quality of the annotations depends only on an annotator's characteristics. 
As a result, the goal of the following models is to define an annotator performance function $\psi$ as a proxy for the distributions $\Pr(P_m)$ in case of persistent or $\Pr(P_m \mid t)$ in case of time-varying performances for each annotator $a_m \in \mathcal{A}$.

\textbf{Annotator models based on interval estimation learning:}
\citetsupplement{supplementary-Donmez2009} proposed IEThresh as one of the first \acronym strategies considering error-prone annotators.
It employs an annotator model based on \textit{interval estimation} (IE) learning.
Originally, IE was developed to address the exploration-exploitation trade-off regarding action selection in reinforcement learning~\citesupplement{supplementary-Kaelbling1993}.
In our \acronym setting, we face a similar issue.
There, finding an action yielding the maximum expected reward corresponds to selecting the annotator with the maximum expected performance.
For this purpose, IEThresh introduces the reward function $r: \mathcal{A} \rightarrow \{0, 1\}$ as mapping from the set of annotators to binary values. Given the majority vote annotation $\hat{z}_n  \in \Omega_{Y}$ for instance $\mathbf{x}_n$, the reward is one, if the annotator~$a_m$ agrees with the majority vote annotation and zero otherwise:
\begin{align}
\label{eq:reward}
&r(a_m \mid \hat{z}_n) \define \delta(z_{nm} \doteq \hat{z}_n) \text{ with }\\
\label{eq:majority-vote}
&\hat{z}_n \define \argmax\limits_{y \in \Omega_{Y}}\left(\sum_{(\mathbf{x}, a, z) \in \mathcal{D}} \delta(z \doteq y) \cdot \delta(\mathbf{x} \doteq \mathbf{x}_n)\right).
\end{align}
This reward estimate requires selecting multiple annotators per instance to take the majority vote of their annotations.
Based on the reward function, IEThresh defines the quality of an annotator $a_m$ as the upper confidence bound of the expected reward. 
It is computed according to:
\begin{equation}
\label{eq:ucb}
\psi(\mathbf{x}_n, a_m \mid \boldsymbol{\omega}_\mathcal{D}) \define \overline{r}_m + t_{\gamma}^{(N_m - 1)} \frac{s_m}{\sqrt{N_m}},
\end{equation}
where  $\overline{r}_m \in [0, 1]$ is the mean reward of $a_m$, $s_m^2 \in [0, 0.25]$ is the reward's variance of $a_m$, and $t_{\gamma}^{(N_m - 1)} \in (0, \infty)$ is the $\gamma \in (0.5, 1)$ quantile of the Student's t-distribution whose degrees of freedom is defined as $N_m \in \mathbb{N}$, i.e., the number of instances annotated by $a_m$ (cf. Eq.~\ref{eq:num-annotations} in the survey).
%A common value of $\gamma$ is $0.975$.
As annotator performance, the reward's upper confidence bound prefers annotators with a high expected reward (first summand in Eq.~\ref{eq:ucb}) and/or a high uncertainty in the estimated reward (second summand in Eq.~\ref{eq:ucb}). 
In total, the model requires a data set~$\mathcal{D}$ and a value for the hyperparameter~$\gamma$ to compute performance values such that parameters can be defined as $\boldsymbol{\omega}_{\mathcal{D}} \define (\mathcal{D}, \gamma)$.
The annotator model of IEThresh can also start with a data set $\mathcal{D} \define \emptyset$ containing no annotations. 
For this purpose, it assumes that each annotator has initially provided one correct and one false annotation.
Concerning the hyperparameter~$\gamma$, the annotator model is quite robust.
However, the quality of the majority vote annotations as estimates of the true class labels strongly affects the accuracy of the annotator performance estimates.

The AL strategy IEAdjCost~\citesupplement{supplementary-Zheng2010} employs a similar annotator model, which considers annotators receiving different payments per annotation.
For this purpose, it defines two phases during the annotation process. 
In the first phase, it explores the annotation qualities, which are estimated according to Eq.~\ref{eq:ucb}.
This exploration phase can start with zero annotations.
At the same time, the annotator model monitors the accuracy of these estimates by computing the confidence interval length
\begin{equation}
l_m \define 2 \cdot t_{\gamma}^{(N_m - 1)} \frac{s_m}{\sqrt{N_m}}
\end{equation}
for each annotator $a_m \in \mathcal{A}$.
The performance value of an annotator $a_m$ is estimated well, if the confidence interval length is below or equal to a threshold: $l_m \leq \delta \in [0.2, 0.4]$. 
Whenever a user-defined fraction $\lambda \in (0, 1)$ of the $M$ annotators have well estimated performance values, the annotator model tries to determine an AC-optimal set of annotators $\mathcal{A}^* \subseteq \mathcal{A}$ with $|\mathcal{A}^{*}| = \lceil \lambda M \rceil$ (cf. Eq.~\ref{eq:ieadjcost}).
If the annotators~$\mathcal{A}^*$ have a combined performance value above or equal to a threshold~$\rho \in [0, 1]$, IEAdjCost enters the second phase by exploiting the performance of the annotators $\mathcal{A}^*$.
We can summarize this two-phase annotator performance measure as
\begin{equation}
\label{eq:ie-adj-cost}
\psi(\mathbf{x}_n, a_m \mid \boldsymbol{\omega}_{\mathcal{D}}) = \begin{cases}
\overline{r}_m + t_{\gamma}^{(N_m - 1)} \frac{s_m}{\sqrt{N_m}} \text{ in phase 1,}\\
\delta(a_m \in \mathcal{A}^*) \text{ in phase 2.}
\end{cases}
\end{equation}
The many hyperparameters in $\boldsymbol{\omega}_{\mathcal{D}} \define (\mathcal{D}, \gamma, \delta, \lambda, R)$ are a major disadvantage of this annotator model.
%A common parametrization is $\gamma \define 0.975$, $\delta \define 0.4$, $\lambda \in [0.25, 0.5]$, and  $R \define 0.95$. 

\textbf{Probabilistic annotator models:}
\citetsupplement{supplementary-Long2013,supplementary-Long2016} and \citetsupplement{supplementary-Long2015} proposed probabilistic annotator models based on Gaussian processes.
They directly model a global noise term in the annotations.
This way, they ensure a certain level of robustness against wrongly annotated instances far from the decision boundary.
Furthermore, they define and estimate the performance of each annotator $a_m \in \mathcal{A}$ as the correctness probability across all instances:
\begin{equation}
\psi(\mathbf{x}_n, a_m \mid \boldsymbol{\omega}_{\mathcal{D}}) = \Pr(P_m = 1 \mid \boldsymbol{\omega}_{\mathcal{D}}).
\end{equation}
The parameters $\boldsymbol{\omega}_{\mathcal{D}}$ are learned using the EP algorithm.

The previously described annotator models assume persistent annotator performances.
In contrast, the annotator model of the AL strategy SFilter~\citesupplement{supplementary-Donmez2010} drops this assumption.
It models the quality of each annotator as a time-varying latent state sequence.
For this purpose, it assumes that the change in the annotator performance from one to the next step follows a Gaussian distribution with a zero-mean and a known variance, which is shared among all annotators.
The use of a zero-mean Gaussian distribution avoids any directional bias.
As a result, the annotator model can detect increases or decreases in the performance of an annotator.
Due to a lack of real-world data with timestamps for each annotation, it is challenging to validate the assumptions of SFilter's annotator model.

\subsection{Annotation-dependent Annotator Performance}
\label{subappendix:annotation-dependent-annotator-performance}
Annotation-dependent annotator performance means that the quality of the annotations depends on an annotator's characteristics and the optimal annotation for a query.
The following strategies consider class labels as annotations, i.e., $\Omega_Z = \Omega_Y$.
As a result, their goal is to define an annotator performance function $\psi$ as a proxy for the distributions $\Pr(P_m \mid Y=y)$ in case of persistent or $\Pr(P_m \mid Y=y, t)$ in case of time-varying performances for each class $y \in \Omega_Y$ and annotator $a_m \in \mathcal{A}$.

\textbf{Probabilistic annotator models:}
Many strategies~\citesupplement{supplementary-Moon2014,supplementary-Nguyen2015,supplementary-Rodrigues2014,supplementary-Wu2013b} employ an annotator model estimating the probability that an annotator $a_m$ provides the correct annotation for an instance of class $y \in \Omega_{Y}$:
\begin{equation}
\Pr(P_m = 1 \mid Y=y) = \Pr(Z_{m} = y \mid Y = y).
\end{equation}
These annotator models mainly differ in their approaches for approximating this probability.

The annotator model of PMActive~\citesupplement{supplementary-Wu2013b} determines its parameters in a maximum likelihood approach using the EM algorithm~\citesupplement{supplementary-Moon1996}.
In the E-step, PMActive estimates the instance's actual class labels in the data set $\mathcal{D}$ and uses them to update the model parameters $\boldsymbol{\mathcal{\omega}}_\mathcal{D}$ in the M-step.
Considering the binary case with $C=2$ classes, the annotator model of PMActive employs a logistic regression model~\citesupplement{supplementary-Ng2002} with the parameters $\mathbf{w} \in \mathbb{R}^{O}, O \in \mathbb{N}$ to estimate an instance's true class membership probability
\begin{align}
\label{eq:logistic-regression-model}
\Pr(Y=1 \mid X=\mathbf{x}, \boldsymbol{\mathcal{\omega}}_\mathcal{D}) &\define \sigma(\mathbf{w}^\mathrm{T} \widetilde{\mathbf{x}}),
\end{align}
where $\sigma: \mathbb{R} \rightarrow (0, 1)$ denotes the logistic function and $\widetilde{\mathbf{x}} \in \mathbb{R}^O$ represents a (non-linear) transformation of instance~$\mathbf{x}$.
The computed class membership probabilities are used to determine the distribution ${\Pr(Z_{m} \mid Y = y, \boldsymbol{\omega}_\mathcal{D})}$, which is modeled as a Bernoulli distribution:
\begin{equation}
\Pr(Z_m = y^\prime \mid Y = y, \boldsymbol{\mathcal{\omega}}_\mathcal{D}) = (\mu_{my})^{\delta(y \doteq y^\prime)} (1 - \mu_{my})^{1-\delta(y \doteq y^\prime)}.
\end{equation}  
The parameter $\mu_{my} \in [0, 1]$ represents the estimated performance value of the annotator~$a_m$ regarding instances of class~$y$.
Altogether, the parameters to be optimized of this model are given by ${\boldsymbol{\omega}_{\mathcal{D}} = (\mathbf{w}, \mu_{11}, \mu_{12}, \dots, \mu_{M1}, \mu_{M2})}$ in the binary case.
The final performance value of an annotator~$a_m$ concerning an instance~$\mathbf{x}_n$ is computed
by taking the maximum of these class-dependent performance values:
\begin{align}
\label{eq:pmactive}
\psi(\mathbf{x}_n, a_m \mid \boldsymbol{\theta}_\mathcal{D}) &= \argmax_{y \in \Omega_{Y}} \left(\Pr(Z_m = y \mid Y=y, \boldsymbol{\omega}_\mathcal{D})\right)\nonumber\\ &= \argmax_{y \in \Omega_{Y}} \left(\mu_{my}\right).
\end{align}
A problem of this performance computation is the missing consideration of the unknown true class label $y_n \in \mathcal{Y}$ of the instance $\mathbf{x}_n \in \mathcal{X}$.
%As a result, it does not matter whether the true class label is $y_n=1$ or $y_n=2$.

The annotator model of the strategy GPC-MA~\citesupplement{supplementary-Rodrigues2014} overcomes this issue by taking an instance's estimated class membership probabilities into account. 
The corresponding performance value of an annotator~$a_m$ regarding an instance~$\mathbf{x}_n$ is given by
\begin{equation}
\begin{gathered}
\label{eq:class-dependent-model}
\psi(\mathbf{x}_n, a_m \mid \boldsymbol{\omega}_\mathcal{D}) = \\\sum\limits_{y \in \Omega_{Y}} \Pr(Y = y \mid X=\mathbf{x}_n, \boldsymbol{\omega}_\mathcal{D}) \Pr(Z_{m} = y \mid Y = y, \boldsymbol{\omega}_\mathcal{D}).
\end{gathered}
\end{equation}
To estimate the required probabilities in the above equation, GPC-MA performs a fully Bayesian treatment based on a Gaussian processes classifier~\citesupplement{supplementary-Rasmussen2003}.
The model's parameters are determined with the EP algorithm~\citesupplement{supplementary-Minka2001}.
The overall estimation procedure is highly computationally intensive.

The annotator model of the strategy Proactive~\citesupplement{supplementary-Moon2014} also defines the annotator performance following Eq.~\ref{eq:class-dependent-model}.
However, it resorts to a simpler and more efficient approach compared to the model of GPC-MA.
For this purpose, it requires a an initial subset of fully annotated instances:
\begin{equation}
\label{eq:fully-annotated-data-set}
\mathcal{D}_\text{init} = \{(\mathbf{x}_n, z_{n1}, \dots, z_{nM}) \mid n \in \mathcal{I} \subset \{1, \dots, N\}\},
\end{equation}
where $\mathcal{I}$ represents the indices of the initially fully annotated instances.
The model computes the majority vote annotation $\hat{z}_n$ (cf. Eq.~\ref{eq:majority-vote}) for each of these instances as an estimator of the true class label. 
Then, the performance value of an annotator~$a_m$ per class~$y$ is computed according to
\begin{gather}
\Pr(Z_m = y \mid Y = y, \mathcal{D}_\text{init}) = \nonumber\\ 
\label{eq:empirical-accuracy}
\frac{\sum\limits_{(\mathbf{x}_n, z_{n1}, \dots, z_{nM}) \in \mathcal{D}_\text{init}} \delta(z_{nm} \doteq \hat{z}_n) \cdot \delta(\hat{z}_n \doteq y)}{\sum\limits_{(\mathbf{x}_n, z_{n1}, \dots, z_{nM}) \in \mathcal{D}_\text{init}} \delta(\hat{z}_n \doteq y)}.
\end{gather}
Despite its efficiency, a  disadvantage of this model is the uncertainty in the specification of the data set $\mathcal{D}_{\text{init}}$.
Especially, its size $|\mathcal{D}_{\text{init}}|$ is a critical issue.
Moreover, once the model has computed the performance values in Eq.~\ref{eq:empirical-accuracy} on the data set~$\mathcal{D}_{\text{init}}$, they are not refined during the following learning cycles.
The class membership probabilities in Eq.~\ref{eq:class-dependent-model} are obtained from any desired probabilistic classification model~$\boldsymbol{\theta}_\mathcal{D}$.
Following our notation, the parameters of this annotator model are defined as $\boldsymbol{\omega}_{\mathcal{D}} = (\mathcal{D}_\text{init}, \boldsymbol{\theta}_\mathcal{D})$.

\citetsupplement{supplementary-Nguyen2015} proposed a similar annotator model differing regarding two aspects.
First, it requires expert annotations as estimates of ground truth labels and compares them with the annotations of crowd workers.
Second, it incorporates prior terms into the nominator and denominator of Eq.~\ref{eq:empirical-accuracy}. 
These prior terms are smoothing hyperparameters.
Taking a Bayesian view, we can interpret them as prior counts.

\subsection{Query-dependent Annotator Performance}
\label{subappendix:query-dependent-annotator-performance}
Query-dependent annotator performance means that the quality of the annotations depends on an annotator's characteristics, a query, and optionally the optimal annotation for this query.
As previously mentioned, the following annotator models consider only instances queries, i.e., $\Omega_X = \mathcal{X}$.
As a result, their goal is to define an annotator performance function~$\psi$ as a proxy for the distributions $\Pr(P_m \mid Z=z, X=\mathbf{x}_n)$ in case of persistent or $\Pr(P_m \mid Z=z, X=\mathbf{x}_n, t)$ in case of time-varying performances for each observed instance $\mathbf{x}_n \in \mathcal{X}$, the optimal annotation $z \in \Omega_Z$, and annotator $a_m \in \mathcal{A}$.

\textbf{Annotator models using self-assessments:}
Many annotator models rely on annotators' self-assessments to estimate their performances.
Such a self-assessment can be a confidence score or a confirmation of a lack of knowledge.

Several \acronym strategies~\citesupplement{supplementary-Kaeding2015,supplementary-Zhong2015,supplementary-Fang2013e,supplementary-Donmez2008b,supplementary-Donmez2010b} allow the annotators to provide \texttt{uncertain} as an alternative annotation to class labels.
Accordingly, these strategies define ${\Omega_Z = \Omega_Y \cup \{\texttt{uncertain}\}}$ as set of possible annotations.
The idea is to query annotators only for instances where an annotator will not provide $\texttt{uncertain}$ as an annotation.
For this purpose, the corresponding annotator models define the annotator performance as:
\begin{equation}
\label{eq:uncertain-performance}
\psi(\mathbf{x}_n, a_m) = 1 - \Pr(Z_m = \texttt{uncertain} \mid \mathbf{x}_n, \boldsymbol{\omega}_{\mathcal{D}}).
\end{equation}
The probability in Eq.~\ref{eq:uncertain-performance} is estimated by defining a binary classification problem where instances annotated with a class label $y \in \Omega_Y$ and the ones annotated with $\texttt{uncertain}$ belong to separate classes.
The annotator models differ in their approaches for solving this classification problem.
\citetsupplement{supplementary-Fang2013e} proposed the strategy EIAL whose annotator model uses the diverse density concept~\citesupplement{supplementary-Maron1998} to transform instances into a new feature space before a user-defined classification model is trained.
\citetsupplement{supplementary-Zhong2015} with their strategy ALCU-SVM employ an SVM and \citetsupplement{supplementary-Kaeding2015} with their strategy $\text{GP-EMOC}_{\text{PDE+R}}$ employ Gaussian processes to solve the binary classification problem.
\citetsupplement{supplementary-Donmez2008b,supplementary-Donmez2010b} proposed an annotator model employing a $k$-means clustering~\citesupplement{supplementary-Xu2005} to estimate the probability in Eq.~\ref{eq:uncertain-performance}. An annotator is queried to annotate the $k$ instances closest to the respective $k$~cluster centroids. If an annotator provides a class label, then the belief of obtaining a class label is propagated to nearby instances within this cluster. Otherwise, the belief of getting $\texttt{uncertain}$ as annotation is propagated analogously. 

In another scenario, \citetsupplement{supplementary-Donmez2008b,supplementary-Donmez2010b} extended their model toward numerical confidence scores by propagating obtained confidence scores within a cluster.
\citetsupplement{supplementary-Ni2012} proposed the strategy BMO.
Its annotator model also expects numerical confidence scores as annotations to estimate the annotators' performances.
It assumes binary class labels $\Omega_Y = \{1, 2\}$ including confidence scores $\Omega_C = [0.5, 1.0]$ as annotations, i.e., $\Omega_Z = \Omega_Y \times \Omega_C$.
Its annotator model estimates the annotators' confidence scores as proxies of their performances.
The model is based on a $k$-NN approach and trained on the confidence scores of each annotator.
It predicts the annotator performance according to:
\begin{equation}
\label{eq:perf-ni2012}
\psi(a_m, \mathbf{x}_n \mid \boldsymbol{\omega}_\mathcal{D}) = \frac{\frac{1}{k} \cdot \sum\limits_{\mathbf{x}_o  \in \mathcal{N}_{\mathbf{x}_n,m}^{k}}  c_{om}}{1+\frac{1}{k}\cdot\sum\limits_{\mathbf{x}_o \in \mathcal{N}_{\mathbf{x}_n,m}^{k}} ||\mathbf{x}_n - \mathbf{x}_o||},
\end{equation}
where $c_{om} \in \Omega_C$ denotes the confidence score of annotator~$a_m$ for instance~$\mathbf{x}_o$ and $\mathcal{N}_{m}^{k}(\mathbf{x}_n) \subset \mathcal{X}$ contains the \mbox{$k$-NN} of instance~$\mathbf{x}_n$ among the instances annotated by the annotator~$a_m$.
Correspondingly, the denominator of Eq.~\ref{eq:perf-ni2012} represents the average confidence of annotator~$a_m$ for the \mbox{$k$-NN} of instance~$\mathbf{x}_n$. 
The nominator is the average distance, where the addition of one prevents dividing by zero.

Annotator models relying on annotators' self-assessments make the central assumption that these assessments are meaningful.
However, there are settings for which humans fail at assessing their capabilities~\citesupplement{supplementary-Kruger1999}.
As a result, the previously discussed annotator models provide unreliable annotator performance estimates.

\textbf{Probabilistic annotator models:}
\citetsupplement{supplementary-Du2010a} proposed a quite simple annotator model as part of their strategy \textit{active
	learning algorithm with human-like noisy oracle} \mbox{(AL-HO)}.
Considering only a single annotator, the goal is to train a classification model behaving similarly to the annotator.
Accordingly, AL-HO expects the classification model to output probabilities that represent the single annotator's performance.
In the case of a linear relationship between annotator and classification model, the annotator performance can be directly expressed through the classification model's most confident prediction:
\begin{equation}
\psi(\mathbf{x}_n, a_m \mid \boldsymbol{\omega}_\mathcal{D}) = \max\limits_{y \in \Omega_Y} \left(\Pr(Y=y \mid X=\mathbf{x}_n, \boldsymbol{\theta}_\mathcal{D})\right).
\end{equation}
Accordingly, the annotator model is equal to classification model, i.e., $\boldsymbol{\omega}_\mathcal{D} = \boldsymbol{\theta}_\mathcal{D}$.
The primary issue of this model is its restriction to the single annotator scenario.

\citetsupplement{supplementary-Zhao2014} developed an annotator model that considering the expertise and difficulty of annotating an instance. 
The expertise of an annotator $a_m$ is represented through a real-valued number $e_m \in \mathbb{R}$. 
As the expertise~$e_m$ approaches~$+\infty$, the performance of annotator $a_m$ increases.
As the expertise~$e_m$ approaches~$-\infty$, the annotator~$a_m$ becomes malicious and provides incorrect annotations on purpose.
In the case $e_m = 0$, the annotator $a_m$ randomly guesses annotations.
The difficulty of annotating an instance $\mathbf{x}_n$ is denoted by $d_n \in (0, +\infty)$.
For simple instances, the difficulty $d_n$ approaches zero, whereas it converges to $+\infty$ for complicated instances.
In the case of a binary classification problem, the model computes the performance according to:
\begin{equation}
\label{eq:eap}
\psi(\mathbf{x}_n, a_m \mid \boldsymbol{\omega}_\mathcal{D}) = \Pr(P_m=1 \mid X=\mathbf{x}_n, \boldsymbol{\omega}_\mathcal{D}) = \sigma\left(\frac{e_m}{d_n}\right),
\end{equation}
where the parameters  $\boldsymbol{\omega}_\mathcal{D}= (e_1, \dots, e_M, d_1, \dots, d_N)$ are estimated in a maximum likelihood approach using the EM algorithm.
In the E-step, the instances' true class labels are estimated.
Subsequently, these estimates are used to update the current estimates of annotators' expertises and instance difficulties.
The application of this annotator model is restricted to problems, where each instance has at least one assigned annotation.
Another issue of this model is that an instance's level of difficulty is not subjective but identical for each annotator. 
\citetsupplement{supplementary-Wallace2010a} also adopt this modeling approach.
However, they do not directly estimate the instances' difficulties and the annotators' expertises but assume that the annotators' salaries are rough proxies of their expertises.

The annotator model proposed by~\citetsupplement{supplementary-Yan2012b,supplementary-Yan2011} drops such assumptions.
It posits that the annotation $z_{nm}$ provided by an annotator $a_m$ depends on the instance $\mathbf{x}_n$ and its actual but unknown class label $y_n$. 
To model this behavior, it uses a Bernoulli distribution for binary classification:
\begin{equation}
\begin{gathered}
\Pr(Z_m=z_{nm} \mid X=\mathbf{x}_n, Y=y_n, \boldsymbol{\omega}_\mathcal{D}) =\\ \sigma(\boldsymbol{w}_m^\mathrm{T}\widetilde{\mathbf{x}}_n)^{\delta(z_{nm} \doteq y_n)}(1-\sigma(\boldsymbol{w}_m^\mathrm{T}\widetilde{\mathbf{x}}_n))^{1-\delta(z_{nm}\doteq y_n)},
\end{gathered}
\end{equation}
where $\mathbf{w}_m \in \mathbb{R}^O$ are the parameters of a logistic regression model for annotator $a_m$.
This model estimates the probability of obtaining the true, i.e., $P_m=1$, or false, i.e., $P_m=0$,  class label as annotation.
As a result, the annotator performance is defined through:
\begin{equation}
\psi(\mathbf{x}_n, a_m \mid \boldsymbol{\omega}_\mathcal{D}) = \Pr(P_m = 1 \mid \mathbf{x}_n, \boldsymbol{\omega}_\mathcal{D}) = \sigma(\mathbf{w}_m^\mathrm{T}\mathbf{x}_n).
\end{equation}
All parameters $\boldsymbol{\omega}_{\mathcal{D}} = (\mathbf{w}_1, \dots, \mathbf{w}_M)$ of this annotator model are determined in a maximum likelihood approach using the EM algorithm, where the true class labels are estimated in the E-step and used to update the model parameters in the M-step.

The annotator models proposed by~\citetsupplement{supplementary-Fang2013,supplementary-Fang2014,supplementary-Fang2012} make similar assumptions to the one of~\citetsupplement{supplementary-Yan2012b,supplementary-Yan2011}.
However, they additionally introduce latent topics, e.g., sports, entertainment, and politics, in case of a text classification problem.
Based on them, they estimate expertise for each pair of annotator and topic.
Since an instance may belong to one or multiple topics, the annotator model estimates the instances' membership degrees regarding the different topics.
The latent topics and associated membership degrees are learned through a GMM~\citesupplement{supplementary-Bishop2006}.
Again, the EM algorithm is employed to learn the annotator model's parameters.
The annotator model of the strategy \textit{self-taught active learning} (STAL)~\citesupplement{supplementary-Fang2012} additionally considers collaborations between annotators.
For this purpose, the estimated best annotator teaches the estimated worst one regarding the annotation of a specific instance.
The learning process of the worst annotator is then modeled by simulating that the worst annotator provided the same annotation as the best one.
In this case, the annotator performance is time-varying because the annotators can get better throughout the annotation process.
Finding an appropriate representation of the aforesaid topics is often difficult.
Therefore, the annotator model~\citetsupplement{supplementary-Fang2013,supplementary-Fang2014} additionally exploits transfer learning.
Its goal is to find common latent topics minimizing the divergence between the target domain and a related domain.

\citetsupplement{supplementary-Yang2018} proposed the \textit{learning-from-targeted crowds} (LFTC) model working with latent topics.
However, it does not use a pre-trained GMM for this purpose but learns them during the training.
The LFTC model models the annotation $z_{nm}$ of an annotator $a_m$ as a function of the instance $\mathbf{x}_n$ and its true (but unknown) class label $y_n$ through a Bernoulli distribution:
\begin{equation}
\begin{gathered}
\Pr(Z_m = z_{nm} \mid X=\mathbf{x}_n, Y=y_n, \boldsymbol{\omega}_\mathcal{D}) = \\\sigma(\boldsymbol{w}_m^\mathrm{T}\mathbf{F}\widetilde{\mathbf{x}}_n)^{\delta(z_{nm} \doteq y_n)}(1-\sigma(\boldsymbol{w}_m^\mathrm{T}\mathbf{F}\widetilde{\mathbf{x}}_n))^{1-\delta(z_{nm}\doteq y_n)},
\end{gathered}
\end{equation}
where $\mathbf{F} \in \mathbb{R}^{L \times O}$ denotes a matrix with $L \in \mathbb{N}, L \ll \text{min}(n, m)$ and $\mathbf{w}_m \in \mathbb{R}^O$ is an annotator-dependent vector.
We can interpret the product $\mathbf{F}\widetilde{\mathbf{x}}_n$ as a representation of instance $\mathbf{x}_n$ through $L$ latent topics and the vector $\mathbf{w}_m$ as an embedding of the expertise of annotator $a_m$ regarding these topics.
The parameters $\boldsymbol{\omega}_\mathcal{D} = (\mathbf{F}, \mathbf{w}_1, \dots, \mathbf{w}_M)$ of the LFTC model are learned in an iterative fashion using the EM algorithm.
\citeauthor{supplementary-Yang2018} have shown that their LFTC model also scales toward deep learning applications.

The previously discussed probabilistic annotator models do not consider the annotator model's epistemic uncertainty regarding its performance estimates.
Therefore, \citetsupplement{supplementary-Herde2021} proposed the \textit{Beta annotator model} (BAM) as part of their \textit{multi-annotator probabilistic active learning} (MaPAL) strategy.
For each annotator, it solves a binary classification problem  with correct and false annotation as possible classes.
These class labels are unknown and have to be estimated.
For this, BAM trains a classifier with the annotations of the annotators $\mathcal{A}\setminus\{a_m\}$ to assess the correctness of the annotations of annotator $a_m$.
This way, any bias toward one annotator is avoided~\citesupplement{supplementary-Rodrigues2014}.
The performance of annotator~$a_m$ is subsequently modeled through a Beta distributions:
\begin{equation}
\Pr(P_m \mid X=\mathbf{x}_n, \boldsymbol{\omega}_\mathcal{D}) = \text{Beta}(P_m \mid \mathbf{f}_{nm} + \boldsymbol{\beta}), 
\end{equation}
where the random variable $P_m \in [0, 1]$ denotes the annotation accuracy, $\mathbf{f}_{nm} = (f_{nm1}, f_{nm2})^\mathrm{T}  \in \mathbb{R}_{\geq 0}^2$ are kernel frequency estimates, and $\boldsymbol{\beta} = (\beta_1, \beta_2)^\mathrm{T} \in \mathbb{R}_{>0}^2$ are hyperparameters.
The kernel frequency estimates $\mathbf{f}_{nm}$ represent the estimated numbers of false and true annotations given by annotator $a_m$ in the neighborhood of instance $\mathbf{x}_n$.
Using a kernel function to quantify neighborhood relations between instances, they are estimated through a Parzen window approach~\citesupplement{supplementary-Chapelle2005}.
Accordingly, we can interpret the vector $\boldsymbol{\beta}$ as prior observations of false and true annotations.
Finally, BAM defines the annotator performance as the Beta distribution's expectation:
\begin{align}
\psi(\mathbf{x}_n, a_m \mid \boldsymbol{\omega}_\mathcal{D}) &= \mathbb{E}[\text{Beta}(P_m \mid \mathbf{f}_{nm} + \boldsymbol{\beta})]\\
&= \frac{f_{nm2} + \beta_2}{f_{nm1} + \beta_1 + f_{nm2} + \beta_2}.
\end{align}
This way, BAM allows the direct incorporation of prior knowledge about the annotators and additionally considers the uncertainty in the performance estimated.
BAM's disadvantages are its high computational complexity and its restriction to the kernel-based Parzen window approach.

\textbf{Static annotator models:} Similar to the annotator model of~\citetsupplement{supplementary-Moon2014}, the model used by the strategy CEAL~\citesupplement{supplementary-Huang2017} requires an initial fully annotated data set~$\mathcal{D}_\text{init}$ according to Eq.~\ref{eq:fully-annotated-data-set}. Additionally, it expects a matrix $\mathbf{S} \in \mathbb{R}_{\geq 0}^{N \times N}$ of similarities between all pairs of observed instances. 
Given these preliminaries, it computes the performance of annotator $a_m$ regarding instance $\mathbf{x}_n$ according to
\begin{equation}
\begin{gathered}
\label{eq:ceal}
\psi(\mathbf{x}_n, a_m \mid \boldsymbol{\omega}_\mathcal{D}) = \frac{1}{k} \sum_{\mathbf{x}_{o} \in \mathcal{N}_{\mathbf{x}_n, \mathcal{D}_{\text{init}}}^{k}}  \mathbf{S}[n,m] \delta(z_{{o}m} = \hat{z}_{o}),
\end{gathered}
\end{equation}
where $\mathcal{N}_{\mathbf{x}_n, \mathcal{D}_{\text{init}}}^{k}$ denotes the $k$-NN of the instance~$\mathbf{x}_n$ in the set~$\mathcal{D}_\text{init}$.
Correspondingly, we can interpret the annotator performance as the similarity-weighted average number of annotations agreeing with the majority vote annotations $\hat{z}_{n^\prime}$ (cf. Eq.~\ref{eq:majority-vote}).
\citetsupplement{supplementary-Chakraborty2020} proposed a similar annotator model.
Instead of a $k$-NN approach, it trains one logistic regression model per annotator based on the initial fully annotated data $\mathcal{D}_\text{init}$.
Each of these models solves a binary classification problem where agreement and disagreement with the majority vote annotation represent the two classes to be distinguished.
Due to the requirement for an initial fully annotated data set $\mathcal{D}_\text{init}$, the disadvantages of both annotators model are related to the discussed ones of the annotator model of~\citetsupplement{supplementary-Moon2014}.
In particular, these models are static such that their performance estimates do not change during the annotation process.

\section{Selection Algorithms}
\label{appendix:selection-algorithms}
In this appendix, we analyze concrete \acronym strategies regarding their selection of query-annotator pairs.
We structure this analysis according to sequential and joint selection algorithms identified in Section~\ref{sec:selection_strategies} in the associated survey.

\subsection{Sequential Selection of Queries and Annotators}
\label{subappendix:sequential-selection}
Sequential selection of queries and annotators is made in two steps. In the first step, one or multiple queries are selected, and corresponding annotators are assigned in the second step. 

\textbf{Single query:} Most selection algorithms select a single query $q_{l^*} \in \mathcal{Q}_\mathcal{X}$ and a single or multiple annotators $\mathcal{A}_{l^*} \subseteq \mathcal{A}$. Mathematically, we express this selection as
\begin{equation}
\mathcal{S} = \{q_{l^*}\} \times \mathcal{A}_{l^*}.
\end{equation}
Many selection algorithms~\citesupplement{supplementary-Long2013,supplementary-Long2016,supplementary-Zhong2015,supplementary-Long2015,supplementary-Rodrigues2014,supplementary-Fang2013,supplementary-Fang2014,supplementary-Zhao2014,supplementary-Wu2013b,supplementary-Ni2012,supplementary-Fang2012,supplementary-Wallace2011,supplementary-Zheng2010,supplementary-Donmez2009} choose the query with maximum utility:
\begin{equation}
q_{l^*} = \argmax\limits_{q_l \in \mathcal{Q}_\mathcal{X}}\left(\phi\left(q_l \mid \boldsymbol{\theta}_{\mathcal{D}}\right)\right).
\end{equation}
As a result, annotating the selected query $q_{l^*}$ is expected to be most beneficial for the training of the classification model.
Subsequently, several AL strategies~\citesupplement{supplementary-Zhong2015,supplementary-Rodrigues2014,supplementary-Fang2013,supplementary-Fang2014,supplementary-Wu2013b,supplementary-Ni2012} present the selected query~$q_{l^*}$ to a single annotator $a^{(l^{*})}$ whose estimated performance regarding this query is maximum compared to the remaining annotators:
\begin{equation}
\label{eq:best-annotator}
\mathcal{A}_{l^*} = \{a^{(l^{*})}\} = \left\{\argmax_{a_m \in \mathcal{A}} \left(\psi\left(q_{l^*}, a_m \mid \boldsymbol{\omega}_{\mathcal{D}}\right)\right)\right\}.
\end{equation}
If the annotator performance estimates are inaccurate, the actual best annotator may be ignored.
Moreover, querying too often the same annotator can bias the annotator performance estimates~\citesupplement{supplementary-Rodrigues2014}.
In particular, performance estimates regarding annotators who have been queried only a few times are often unreliable~\citesupplement{supplementary-Donmez2009}.
To resolve these issues, \citetsupplement{supplementary-Zhao2014} proposed two alternative annotator selection algorithms. 
The first algorithm chooses an annotator with the probability being proportional to the respective performance estimate:
\begin{equation}
\Pr(A=a_m \mid Q=q_{l^*}) = \frac{\psi\left(q_{l^*}, a_m \mid \boldsymbol{\omega}_{\mathcal{D}}\right)}{\sum\limits_{a \in \mathcal{A}}\psi\left(q_{l^*}, a \mid \boldsymbol{\omega}_{\mathcal{D}}\right)},
\end{equation}
where $\psi\left(q_l, a \mid \boldsymbol{\omega}_{\mathcal{D}}\right) \geq 0$ is required and $A$ denotes the random variable defined over the annotator set $\mathcal{A}$.
The second strategy is inspired by the $\epsilon$-greedy algorithm used for the multi-armed bandit problem in reinforcement learning~\citesupplement{supplementary-Kuleshov2014}.
The probability of selecting an annotator $a_m$ is computed as
\begin{equation}
\Pr(A=a_m \mid Q=q_{l^*}) = \begin{cases}1 - \epsilon + \frac{\epsilon}{M} \text{ if } a_m \doteq a^{(l^{*})},\\ \frac{\epsilon}{M}\text{ otherwise.}\end{cases}
\end{equation} 
The hyperparameter $\epsilon \in [0, 1)$ controls the exploration-exploitation trade-off.
The estimated best annotator is selected to exploit the knowledge about the annotators' performances, whereas another random annotator is picked for exploration.

Selecting only a single annotator per iteration cycle, i.e., $|\mathcal{A}_{l^*}|=1$, can be disadvantageous.
In particular, in the initial learning phase, retraining the annotator model and the classification model with partially false annotations leads to non-reliable estimates of annotator performances and query utilities in subsequent iteration cycles.
To overcome this issue, querying a user-defined number of annotators per query represents a more reliable alternative~\citesupplement{supplementary-Long2013,supplementary-Long2016,supplementary-Long2015} at the expense of increased AC per query.
However, this number is a fixed hyperparameter during the entire annotation process. 
Thus, the number of selected annotators is independent of their performance estimates.
To obtain a more adaptive selection of multiple annotators, \citetsupplement{supplementary-Donmez2009} proposed a threshold-based selection as part of their strategy IEThresh.
It specifies the annotator set according to
\begin{equation}
\mathcal{A}_{l^*} = \left\{a \in \mathcal{A} \, \bigl| \, \psi\left(q_{l^*}, a \mid \boldsymbol{\omega}_{\mathcal{D}}\right) \geq \rho \cdot \psi\left(q_{l^*}, a^{(l^{*})}  \, \bigl| \, \boldsymbol{\omega}_{\mathcal{D}}\right)\right\},
\end{equation}
where $\rho \in [0, 1]$ is a hyperparameter.
It specifies the minimum annotator performance to be selected in dependence of the performance of the estimated best annotator $a^{(l^{*})}$.
Although the number of selected annotators is adaptive, an appropriate parametrization of $\rho$ needs to be determined regarding the characteristics of the classification problem and the available annotators.
The strategy IEAdjCost of \citetsupplement{supplementary-Zheng2010} uses different annotator selection algorithms for different stages during the annotation process.
In the initial phase, the annotator selection is performed similarly to IEThresh. 
The idea is to explore the performances of the annotators in this stage.
Once these performance estimates have been sufficiently explored, IEAdjCost switches the annotator selection algorithm to exploit the gained knowledge about the annotators.
In this exploitation phase, the strategy determines a fixed subset of annotators reaching a combined accuracy above a user-defined threshold while minimizing the AC (cf.~Eq.~\ref{eq:ieadjcost}).

None of the above-described selection algorithms considers any forms of collaboration among the annotators.
However, \citetsupplement{supplementary-Chang2017} showed that collaboration could lead to improved annotations.
The strategy STAL of~\citetsupplement{supplementary-Fang2012} selects next to the estimated best annotator also the estimated worst annotator.
Instead of querying the two annotators independently, the best annotator chooses an annotation and explains this decision's worst annotator.
This way, the worst annotator can gain new knowledge to enhance the performance.
While it would be an option to propagate the  best annotator's knowledge to more than one annotator, the pairwise collaboration ensures that error knowledge is not propagated among too many annotators that would eventually deteriorate the annotation process.

Another important aspect when selecting annotators concerns their annotation workloads.
For example, assigning too many queries to the same annotator may decelerate the annotation process since this annotator has to process the queries sequentially.
As a solution, \citetsupplement{supplementary-Wallace2011} explicitly model the workload across multiple annotators as part of their strategy MEAL.
They use a categorical distribution representing either a preference for uniform workloads among the annotators or another objective.
From this distribution, annotators are then drawn and assigned to respective queries.

\textbf{Batch of queries:} Selecting only a single query during each AL iteration cycle is likely to have no significant impact on deep learning models'
performances because of their local optimization methods~\citesupplement{supplementary-Sener2018}.
Therefore, the strategy \textit{deep active learning from targeted crowds} (DALC) proposed by~\citetsupplement{supplementary-Yang2018} selects a batch with a user-defined number of queries having the highest utilities per iteration cycle. 
Subsequently, it assigns each of these queries to the annotator with the respective highest estimated performance.
As this selection algorithm does not consider query diversity, the queries may request redundant learning information leading to worse performance than random sampling~\citesupplement{supplementary-Kirsch2019}.

\subsection{Joint Selection of Queries and Annotators}
\label{subappendix:joint-selection}
Selecting queries without considering the performances of the available annotators can result in low-quality annotations because there is no guarantee that at least one annotator has a sufficient performance regarding the selected query~\citesupplement{supplementary-Zhong2015}.
This problem can be resolved by applying a selection strategy that jointly selects queries and annotators.

\textbf{Single query}: 
For this purpose, the query utility measure~$\phi$ and the annotator performance measure~$\psi$ are combined appropriately.
Taking the product of both represents a simple but effective combination~{\citesupplement{supplementary-Huang2017,supplementary-Moon2014,supplementary-Donmez2008b,supplementary-Donmez2010b}}.
Accordingly, a query-annotator pair is selected through
\begin{equation}
\mathcal{S} = \left\{\argmax\limits_{(q_l, a_m) \in \mathcal{Q}_\mathcal{X} \times \mathcal{A}} \left(\phi(q_l \mid \boldsymbol{\theta}_{\mathcal{D}}) \cdot \psi(q_l, a_m \mid \boldsymbol{\omega}_{\mathcal{D}})\right)\right\}.
\end{equation}
This selection balances the trade-off between query utility and annotator performance by ensuring that both need to be high.

There are more advanced approaches to combine annotator performance and query utility~\citesupplement{supplementary-Herde2021,supplementary-Chakraborty2020,supplementary-Yan2012b,supplementary-Yan2011}.
All of them focus on instance queries and class labels as annotations.

\citetsupplement{supplementary-Yan2011} proposed a selection algorithm picking an instance-annotator pair by solving a linearly constrained, bi-convex optimization problem with the BFGS~\citesupplement{supplementary-Nocedal2006} algorithm.
As a solution, one obtains an instance including annotator importance values.
This instance is not guaranteed to be in the set $\mathcal{X}$ of observed instances.
Therefore, \citeauthor{supplementary-Yan2011} resorts to select the observed instance closest to the optimal one and the annotator with the respective highest performance.

The strategy $\text{ML+CI}^{*}$ of \citetsupplement{supplementary-Yan2012b} estimates the information that an annotator's class label provides regarding an instance's true but unknown class membership.
For this, mutual information~\citesupplement{supplementary-Cover1991} is employed as a criterion for selecting an instance-annotator pair.

The previous two selection algorithms employed US as instance utility criterion.
In contrast, \citetsupplement{supplementary-Nguyen2015} and \citetsupplement{supplementary-Herde2021} compute the classification model's expected performance gain when obtaining an instance's class label from a certain annotator.
In this context, \citetsupplement{supplementary-Nguyen2015} differs only between two groups of annotators: error-prone crowd workers and omniscient experts.
For crowd workers, the expected performance gain is computed on the set of non-annotated instances.
This performance gain directly considers the estimated accuracy of the crowd workers' majority vote annotation for an instance.
The expected performance gain is computed for the experts on the set of instances already annotated by the crowd workers.
The idea is to query only experts for re-annotating instances that the crowd workers likely assigned to the wrong class.
\citetsupplement{supplementary-Herde2021} with their strategy MaPAL advance the performance gain computation by explicitly differing between all individual annotators instead of groups.
Based on the probabilistic active learning framework~\citesupplement{supplementary-Kottke2021}, MaPAL selects the instance-annotator pair maximizing the classification model's probabilistic performance gain.
This selection is non-myopic by simulating an instance's annotations from multiple annotators.
However, such a kind of look-ahead increases the computational complexity of the instance-annotator pair selection.

\textbf{Batch of queries:} All of these joint selection algorithms select only a single query-annotator pair during each iteration, whereas \citetsupplement{supplementary-Chakraborty2020} allows for a batch selection, i.e., $|\mathcal{S}|>1$.
The corresponding selection algorithm solves an optimization problem finding a trade-off between annotator performances, instance query utilities, and redundancies between selected instance queries.
The redundancies are considered by incorporating cosine similarity measurements between instances into the objective function.
This way, a batch of instances with low similarities to each other will be selected.
\citeauthor{supplementary-Chakraborty2020} has shown that the optimization problem can be formulated as an equivalent linear programming problem.

\bibliographystylesupplement{IEEEtranN}
\bibliographysupplement{supplementary}

\end{document}